\documentclass[journal=jcisd8,manuscript=article,layout=onecolumn]{achemso} %layout=twocolumn

\usepackage{chemformula} % Formula subscripts using \ch{}
\usepackage[T1]{fontenc} % Use modern font encodings

\usepackage{url}
\usepackage{float}
\usepackage{caption}
\usepackage{amsmath}
\usepackage{graphicx}
\usepackage{listings}
\usepackage{float}
\usepackage{tabularx}
\usepackage{amsfonts}
\usepackage{hyperref}
\usepackage{listings}
\usepackage{verbatim}
\usepackage{xcolor}
\usepackage{soul}
\usepackage{fancyvrb}
\usepackage{mdframed}
\usepackage{adjustbox}
\usepackage{makecell}
\usepackage{comment}
\usepackage{enumitem}
\usepackage{booktabs}
\usepackage{arydshln}
\usepackage{subcaption}     
\usepackage{chngcntr}
\usepackage[left]{lineno} % Line numbers
\usepackage[final,markup=underlined,commandnameprefix=always]{changes} % For final version

\definecolor{codebg}{RGB}{240,240,240}
\definecolor{codegreen}{RGB}{0,128,0}
\definecolor{codeblue}{RGB}{0,0,128}
\definecolor{codepurple}{RGB}{128,0,128}
\definecolor{codegray}{RGB}{128,128,128}

\lstdefinestyle{mystyle}{
    backgroundcolor=\color{codebg},
    commentstyle=\color{codegreen},
    keywordstyle=\color{codeblue},
    numberstyle=\tiny\color{codegray},
    stringstyle=\color{codepurple},
    basicstyle=\ttfamily\small, % Default font size for code listings
    breakatwhitespace=false,
    breaklines=true,
    captionpos=b,
    keepspaces=true,
    numbers=left,
    numbersep=5pt,
    showspaces=false,
    showstringspaces=false,
    showtabs=false,
    tabsize=2
}

\newcommand{\myverbatim}[1]{%
  \colorbox{codebg}{%
    \begin{minipage}{\dimexpr\linewidth-2\fboxsep}%
      \begin{verbatim}
      #1
      \end{verbatim}
    \end{minipage}%
  }%
}

\usepackage{caption}

\author{Morgan Thomas}
\affiliation[upf]
{Computational Science Laboratory, Universitat Pompeu Fabra, Barcelona Biomedical Research Park (PRBB), C Dr. Aiguader 88, 08003 Barcelona, Spain}
\alsoaffiliation[khalifa]
{Department of Medical Sciences, Khalifa University of Science and Technology, 127788 Abu Dhabi, UAE}
\email{morganthomas263@gmail.com}

\author{Albert Bou}
\affiliation[upf]
{Computational Science Laboratory, Universitat Pompeu Fabra, Barcelona Biomedical Research Park (PRBB), C Dr. Aiguader 88, 08003 Barcelona, Spain}

\author{Gianni De Fabritiis}
\affiliation[icrea]{Instituci\'o Catalana de Recerca i Estudis Avan\c{c}ats (ICREA), Passeig Lluis Companys 23, 08010 Barcelona, Spain}
\alsoaffiliation[upf]
{Computational Science Laboratory, Universitat Pompeu Fabra, Barcelona Biomedical Research Park (PRBB), C Dr. Aiguader 88, 08003 Barcelona, Spain}
\alsoaffiliation[acellera]{Acellera Labs, C Dr Trueta 183, 08005, Barcelona, Spain}
\email{g.defabritiis@gmail.com}

\title[TTT for Chemical Exploration]{Test-Time Training Scaling Laws for Chemical Exploration in Drug Design}

\keywords{De Novo Molecule Generation, Test-Time Training, Reinforcement Learning, Drug Discovery, Chemical Exploration, Molecule Exploration, Chemical Language Models}

\begin{document}

\begin{tocentry}
\centering
\includegraphics{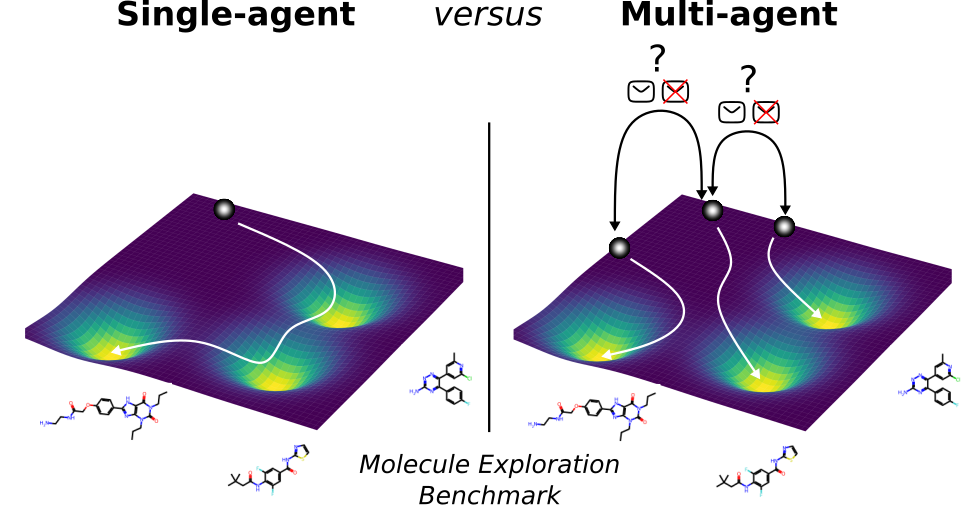}
\end{tocentry}


\begin{abstract}
Chemical Language Models (CLMs) leveraging reinforcement learning (RL) have shown promise in de novo molecular design, yet often suffer from mode collapse, limiting their exploration capabilities. Inspired by Test-Time Training (TTT) in large language models, we propose scaling TTT for CLMs to enhance chemical space exploration. We introduce MolExp, a novel benchmark emphasizing the discovery of structurally diverse molecules with similar bioactivity, simulating real-world drug design challenges. Our results demonstrate that scaling TTT by increasing the number of independent RL agents follows a log-linear scaling law, significantly improving exploration efficiency as measured by MolExp. In contrast, increasing TTT training time yields diminishing returns, even with exploration bonuses. We further evaluate cooperative RL strategies to enhance exploration efficiency. These findings provide a scalable framework for generative molecular design, offering insights into optimizing AI-driven drug discovery.
\end{abstract}

%%%%%%%%%%%%%%%%%%%%%%%%%%%%%%%%%%%%%%%%%%%%%%%%%%
% Introduction
%%%%%%%%%%%%%%%%%%%%%%%%%%%%%%%%%%%%%%%%%%%%%%%%%%
\section{Introduction}\label{introduction}
% Exploring chemical space for drug discovery
The vast chemical space, estimated to contain over $10^{60}$ drug-like molecules \cite{polishchuk2013estimation}, presents a formidable challenge in drug discovery. Efficiently identifying diverse, safe, and efficacious drug candidates can accelerate development timelines and reduce costs by hundreds of millions of dollars \cite{bender2021artificial}. Traditional virtual screening of limited chemical libraries captures only a fraction of possible compounds, restricting exploration. In contrast, goal-directed generative AI, particularly Chemical Language Models (CLMs), enables \textit{de novo} design of novel molecules with optimized properties, implicitly navigating expansive chemical spaces \cite{arus2019exploring}.

% Importance of diverse chemical exploration
Thorough exploration of chemical space is critical to maximize the potential of generative AI in drug design. Identifying structurally diverse molecules with similar bioactivity increases the likelihood of discovering backup series with distinct chemical and biological profiles, mitigating risks in drug development \cite{brown2018recent}. Moreover, exploring novel chemical regions can circumvent patented compounds \cite{shimizu2023aipatent}. 
However, many generative models, including CLMs, are susceptible to mode collapse when fine-tuned and depending on hyperparameters \cite{amabilino2020guidelines},particularly those reliant on machine learning models with limited applicability domains due to epistemic uncertainty \cite{preuer2018frechet, renz2019failure}.

% Role of CLMs and RL in drug design
CLMs, autoregressive models trained on SMILES strings \cite{weininger1988smiles}, have become the leading approach for generative drug design \cite{segler2018generating, grisoni2023chemical, skinnider2021chemical}, excelling in benchmarks like GuacaMol \cite{brown2019guacamol, polykovskiy2020molecular}. By integrating reinforcement learning (RL), seminal works \cite{olivecrona2017molecular, popova2018deep} have enabled CLMs to generate molecules with tailored properties. Advances in RL algorithms \cite{bjerrum2023faster, thomas2022augmented, guo2024augmented, thomas2025reinforcing} and practical applications \cite{fialkova2021libinvent, thomas2024promptsmiles} have further enhanced their utility, with REINFORCE-based methods remaining superior due to robust pre-training \cite{sutton2018reinforcement, bou2024acegen, ahmadian2024back}.

% Test-Time Training and scaling
Inspired by Test-Time Training (TTT) in large language models, where model parameters are temporarily updated for specific tasks \cite{akyurek2024surprising}, we frame RL-based optimization of pre-trained CLMs as a form of TTT. In LLMs, scaling inference-time sampling has shown log-linear performance improvements \cite{brown2024large}. Similarly, scaling TTT for CLMs—by increasing the number of RL agents or training time—offers potential to enhance chemical space exploration. Cooperative RL strategies further amplify exploration through population-based learning \cite{parker2020effective}, though their application in molecular design remains underexplored.

% Contributions of this work
In this work, we make two key contributions. First, we demonstrate that scaling TTT by increasing the number of independent RL agents follows a log-linear scaling law, significantly enhancing exploration efficiency. Conversely, extending TTT training time yields limited benefits, even with exploration bonuses or diversity filters. We also propose and evaluate cooperative RL strategies to further improve exploration efficiency. Second, we introduce MolExp, a novel benchmark to measure \textbf{Mol}ecule \textbf{Exp}loration. MolExp requires the rediscovery of structurally diverse molecules with comparable bioactivity. Unlike existing benchmarks, MolExp requires exploration of \textit{all} high-reward regions of chemical space, extending the boundaries of real-world drug design.

This work provides a robust framework for enhancing chemical space exploration, addressing critical challenges in AI-driven drug discovery. By leveraging TTT and cooperative RL, we offer practical tools through MolExp and scalable strategies that can accelerate the discovery of diverse drug candidates. 

%%%%%%%%%%%%%%%%%%%%%%%%%%%%%%%%%%%%%%%%%%%%%%%%%%
% Related work
%%%%%%%%%%%%%%%%%%%%%%%%%%%%%%%%%%%%%%%%%%%%%%%%%%
%\section{Related Work}

%\textbf{Exploration in reinforcement learning}
Balancing the exploration-exploitation trade-off is a constant challenge in RL, including RL for drug design \cite{langevin2024balancing}. Exploitation initially maximizes the reward quicker but may lead to a lower overall reward, while exploration may lead to a higher overall reward and increased diversity of solutions but takes longer. Diversity has consistently been an important topic in generative molecular design and is increasingly included in the evaluation of performance \cite{thomas2022re, renz2024diverse}. Several approaches have been proposed to increase diversity, for example, using dual networks to balance exploitation-exploration during sampling \cite{liu2019exploration}, penalizing the reward of molecules similar to previously generated ones \cite{blaschke2020memory}, or increasing the reward of molecules that are dissimilar to those previously generated \cite{thiede2022curiosity}. Recent work shows that random network distillation (RND) \cite{burda2018exploration} is particularly effective at increasing molecular diversity with low overhead due to not requiring an explicit memory of previously generated molecules \cite{svensson2024diversity, thomas2025reinforcing}. 

%\textbf{Multi-agent reinforcement learning}
Previous work has applied population-based RL to \textit{de novo} drug design. \citet{hu2024novo} deployed up to 4 GPT agents as CLMs that use the REINVENT loss formulation with an additional penalty to discourage the $k$-th agent from being similar to previous agents. This multi-agent setup outperformed baseline algorithms on the GuacaMol benchmark \cite{brown2019guacamol}. However, this benchmark consists of mostly simple tasks (as also stated by the authors). As demonstrated by their experiment comparing 1, 2, 4 and 8 agents all having near maximum performance and all within standard deviation. Moreover, independent agents and other cooperative strategies were not investigated. Therefore it remains to be seen whether cooperative strategies improve performance significantly, by how much, and which cooperative strategies work best.

Generative molecular design benchmarks must reflect realistic scenarios in drug design for practical application, as gold standard evaluation via synthesis and experimental validation of molecules is too costly at scale. The GuacaMol benchmark \cite{brown2019guacamol} proposed a suite of 20 tasks, the authors highlighted that $\sim 15/20$ tasks are easily solved by the generative algorithms. The MolOpt benchmark limited these tasks to a budget of 10,000 molecules during optimization, shifting the focus to efficiency, however, this likely over-rewards exploitative algorithms as opposed to explorative ones. New criteria have been defined to additionally measure the quality of chemistry generated which has changed rank performance \cite{thomas2022re, renz2024diverse}, moreover, rank performance on this MolOpt benchmark didn't reflect performance on more realistic optimization objectives such as predicted binding affinity via docking \cite{bou2024acegen}. Hence, there remains significant scope for new exploration-focused objectives that better reflect realistic optimization objectives in drug design.

%%%%%%%%%%%%%%%%%%%%%%%%%%%%%%%%%%%%%%%%%%%%%%%%%%
% Methods
%%%%%%%%%%%%%%%%%%%%%%%%%%%%%%%%%%%%%%%%%%%%%%%%%%
\section{Methods}
\subsection{Problem definition}\label{sec:problem_definition}
%%%%%%%%%%%%%%%%%%%%%%%%%%%%%%%%%%%%%%%%%%%%%%%%%%
% Problem definition
%%%%%%%%%%%%%%%%%%%%%%%%%%%%%%%%%%%%%%%%%%%%%%%%%%
% The main goal of generative drug design is optimize multiple properties etc.
The generative molecular design problem is the generation of molecules that are optimal in one or more properties defined by the drug discovery context, for example, drug-likeness and predicting binding affinity against a protein implicated in a disease. The scoring function (also referred to as an oracle) $s(x)\in[0,1]$ may combine one or more property predictors $p_{i}(x)$ that score a given molecule $x$ returning a real value $p_{i}: X \rightarrow \mathbb{R}$. The final score reflects the desirability of the molecule for the given context, where the goal is to generate a molecule with the best score.

\vskip -0.1in
\begin{equation}
    \arg\max_{x \in X} s(x)
\end{equation}
 
However, chemical space is vast, property predictors are inherently inaccurate with high uncertainty, and Pareto optimal solutions have different property profiles.
Hence, multiple molecules exist across chemical space that are considered hits if the score is above a given threshold $T$ (or sometimes more simply the top-$k$ molecules). 

\vskip -0.1in
\begin{equation}
\exists x_i, ..., x_N \in X, \, x_i \neq x_j \text{ such that } s(x_i) \ge T
\end{equation}

We propose that the problem of generative molecular design should extend to finding all possible hits across chemical space $\mathcal{H}_T$, and postulate that this better aligns with practical requirements in drug discovery.

\vskip -0.1in
\begin{equation}
\mathcal{H}_T = \{ x \in X \mid s(x) \ge T \}
\end{equation}

\begin{figure*}[ht]
%\vskip 0.2in
\begin{center}
\centerline{\includegraphics[width=\textwidth]{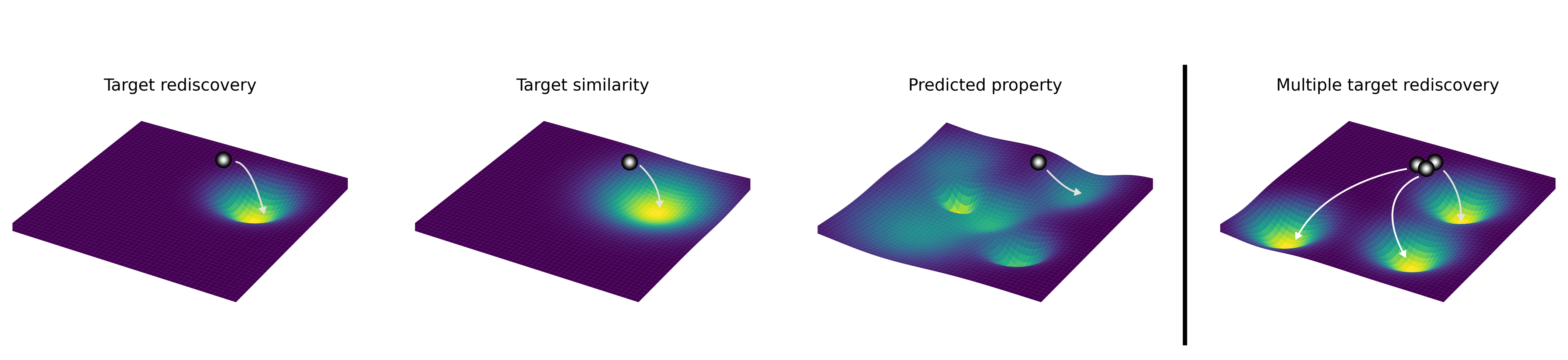}}
\caption{Schematic representation of chemical space manifold according to molecular desirability, where energy wells are desirable sub-spaces of 'high-reward' areas according to a given objective. (Left) Typical objectives in the literature. (Right) Our proposal objective. The sphere represents a generative model learning to optimize desirability or maximize reward. Target rediscovery where the reward reflects maximal similarity to target molecule in chemical space. Target similarity where the reward reflects similarity above a certain threshold to a target molecule in chemical space, perhaps with additional multi-parameter objectives. Predicted property of molecules in chemical space, where multiple regions of reward exist. Lastly, our proposal, multiple target rediscovery where there exists different and distinct regions of high reward chemical space, and the goal is to rediscover all target molecules.}
\label{fig:chem_space}
\end{center}
\vskip -0.2in
\end{figure*}

We propose that a generative model benchmark should adequately mimic this concept of identifying multiple, even structurally unrelated molecular targets in chemical space (\autoref{fig:chem_space}, right). 
Previously published benchmarks don't test this adequately, with objectives focused on a single narrow target, or a single broad target (\autoref{fig:chem_space}, left) and performance measured only by maximum $s(x)$ achieved. Alternative evaluations have been proposed that also measure the number of diverse hits \cite{thomas2022re, renz2024diverse}. Although they are an improvement in benchmark evaluation, there are still some limitations to this approach. 
Firstly, rewarding diverse hits are not suitable for many tasks converging to single solutions, hence why Renz et al. \citeyear{renz2024diverse} focus only on 3/23 tasks from MolOpt.
Secondly, tasks where diverse hits are suitable, such as property prediction where multiple modes exist, are typically trivial. In some cases, maximum performance is achieved and hence no statistical differences can be observed \cite{hu2024novo}. Lastly, when measuring the number of diverse hits it is not known how many possible diverse hits exist in solution space i.e., there is no ground truth. We propose to add a ground truth via a limited number of known modes in solution space. Moreover, ensuring the modes are distinct and structurally unrelated in a chemical sense, increases the difficulty of the benchmark and inherent exploration required, enabling the measurement of statistically significant improvements.

\subsection{Reinforcement learning with chemical language models}
CLMs utilize a language representation, commonly SMILES \cite{weininger1988smiles}, to leverage natural language models such as recurrent neural networks (RNNs) or GPT-style models. For example, SMILES represent the molecular structure of paracetamol with the string ``CC(=O)Nc1ccc(O)cc1". These models operate autoregressively, predicting the next token $x_t$ in a sequence $X=(x_1,x_2,\ldots,x_T)$ based on the context provided by previous tokens. The process can be formalized as maximizing the conditional probability of each token given the preceding sequence $P(X)=\prod_{t=1}^{T} P(x_t | x_{< t})$, where $x_{< t}$ denotes the tokens up to $t - 1$. The model is trained using the cross-entropy loss function which is equivalent to the negative log-likelihood \autoref{eq:NLLloss}. Where $T_i$ is the length of the $i$-th sequence \chadded{in the time domain $t$}, and $\theta$ are the model parameters.

\vskip -0.1in
\begin{equation}\label{eq:NLLloss}
\mathcal{L} = - \mathbb{E}_{x \sim \mathcal{D}} \left[\sum_{t=1}^{T_i} \log \left( P_\theta(x_t | x_{<t}) \right)\right]
\end{equation}

Reinforcement learning (RL) offers a structured approach to molecular generation by formalizing the problem as a Markov Decision Process (MDP) defined by $\langle S, A, R, P, \rho_0 \rangle$. Here, $S$ represents the states (i.e., partial molecular structures), $A$ denotes available actions (i.e., adding the next token that could describe an atom), $R$ defines a scalar reward function (i.e., evaluating molecular properties), $P$ is the state transition probability, and $\rho_0$ is the initial state distribution. Here, the pre-trained CLM acts as an already parameterized policy $\pi_\theta$ that defines an initial transition probability of selecting the next action at a given timestep $P(s_{t+1} | a_t, s_t)$. 

The specific RL algorithm used is a REINFORCE-based policy-gradient algorithm. REINFORCE learns a policy $\pi_\theta$ that maximizes the expected cumulative reward $J(\theta)=\mathbb{E}_{\tau \sim \pi_\theta}[R(\tau)]$, where $\tau$ represents a trajectory of states and actions. Its core objective is to adjust the policy parameters $\theta$ such that the likelihood of actions leading to higher rewards increases. This is achieved by updating $\theta$ proportionally to the gradient of the reward-weighted log-probability of action sequences, expressed as $\nabla J(\theta) = \mathbb{E}_{\tau \sim \pi_{\theta}} \left[ \sum_{t=0}^{T} \nabla_\theta \log \pi_\theta(a_t|s_t) \cdot R(\tau) \right]$. REINFORCE is particularly suited for problems with sparse or delayed rewards, as it treats entire trajectories as units of optimization. This works well when building molecules which only receive a reward when the building is complete. 

We deviate to vanilla REINFORCE by reshaping the reward to regularize the agent policy to the trajectory probability by prior policy denoted as $\log\pi_{prior}(\tau)= \sum_{t=1}^{T} \log \pi_{\text{prior}}(x_t \mid x_{<t})$ with a coefficient $\sigma$ and by adding an exponent $\alpha$, as proposed by \citet{thomas2025reinforcing} shown in \autoref{eq:acegen_reward}. \chadded{Note that clipping is used to ensure the reshaped reward is always positive}. These components improve the quality of molecules proposed by the agent and improve performance by steepening the gradients in the reshaped reward landscape. In addition to reshaping the reward, we implement experience replay to augment on-policy data with off-policy data at each iteration. A replay buffer is kept of high-reward molecules which is sampled with prioritization proportional to their reward. This additionally improves learning efficiency \cite{gao2022sample, thomas2025reinforcing}.

\vskip -0.1in
\begin{equation}\label{eq:acegen_reward}
R(\tau)_{reshaped} = clip(R(\tau) + \sigma \cdot \log\pi_{prior}(\tau))^{\alpha}
\end{equation}

\subsection{Cooperative reinforcement learning}
Cooperative RL extends the framework of single-agent RL to systems with multiple independent or interacting agents, each learning to optimize its own policy in an environment. The problem is the same as defined before, but now with $\mathcal{N}$ agents. 

In the independent case, each agent still aims to maximize its cumulative reward using its own $\langle \mathcal{S}^i, \mathcal{A}^i, R^i, P^i \rangle$ for $i \in \mathcal{N}$ with an independent replay buffer for each agent. Note that independent agents will still deviate due to the stochastic nature of policy $\pi_{i}$ sampling from the transition probabilities $P(s^i_{t+1}|a^i_t,s^i_t)$ leading to the sampling of different actions $a^i_t$. 

In the interaction case, we focused on methods of cooperation within a population-learning setting which typically results in more diverse solutions \cite{oroojlooy2023review}. In this case, using $\langle \mathcal{S}, \mathcal{A}^i, R, P^i \rangle$ for $i \in \mathcal{N}$ where the states $S$ and $R$ may be shared across agents in a cooperative manner. We investigated the following methods of cooperation (for additional detail and equations see \chreplaced{Section S6}{Appendix F}):

\begin{description}[topsep=0.01in, itemsep=0.01in]

\item[Purge] Ensure agent replay buffers are mutually exclusive with respect to states contained within them.
    
\item[Shared] Share a single replay buffer for all agents. This should result in policy convergence and act as a type of negative control.

\item[Shared with bonus] Same as \textbf{Shared} but with a reward bonus for novel states not contained within the shared replay buffer.
    
\item[Noise] Apply Gaussian noise to each agent's policy parameters at the beginning of training. 

\item[RND] Apply RND universally to provide a reward bonus for novel states. 

\item[ENT$_\mathcal{S}$] Minimize the entropy of policy uncertainty across population states to encourage specialized behavior.

\item[CE$_{\mathcal{S}}$] Minimize the cross-entropy loss of policy uncertainty across population states to encourage explicit specialization towards its own states. 
 
\item[DIFF$_\mathcal{S}$] Maximize the difference between policy uncertainty in the agents own states relative to population states.

\item[DIFF$_\mathcal{N}$] Re-implementation of the cooperative strategy applied by \citet{hu2024novo} which seeks to maximize the log-probability difference of on-policy states between the current agent and other agents.

\item[DvD] Diversity via Determinant (DvD) inspired by \citet{parker2020effective}. Calculate a policy behavior embedding and maximize the divergence in policy behavior as measured by the determinant of the behavioral embedding transformed by a kernel function. 

\item[POPNORM] Normalize the agent return by subtracting the average return of the population for on-policy states, encouraging divergent behavior. 
    
\end{description}

%%%%%%%%%%%%%%%%%%%%%%%%%%%%%%%%%%%%%%%%%%%%%%%%%%
% Experiments
%%%%%%%%%%%%%%%%%%%%%%%%%%%%%%%%%%%%%%%%%%%%%%%%%%
\section{Results}
First, we describe our proposed benchmark MolExp, and show performance of various baseline RL algorithms with a comparison to an established benchmark. Then, we conducted three experiments to assess chemical exploration. 

\begin{enumerate}[itemsep=0.01in, topsep=0.01in]
    \item We investigated the effect of scaling TTT by the number of RL agents, or scaling TTT by the training time of a single RL agent by its molecule budget. 
    \item We proposed and tested different cooperative RL strategies to enhance exploration efficiency. 
    \item We translated our findings to a practically relevant task in optimizing the predicted bioactivity of generated compounds.
\end{enumerate}

\subsection{Molecule exploration benchmark}
To address our proposed problem formulation to identify \textit{all} hits in chemical space, we designed a new molecular exploration (MolExp) benchmark. Crucially, a ground truth is required to know that \textit{all} solutions are identified \footnote{Oracles that are more practically relevant, such as bioactivity prediction or docking, don't contain a ground truth set of all possible high scoring solutions.}. For this reason, we use the toy-task of multiple molecule rediscovery via optimization of a similarity scoring function. To ensure relevance to real-world discovery, each set of molecules are at minimum pre-clinical drug candidates bioactive against the same target. Moreover, to make the objective more challenging, the set of molecules are selected to be structurally dissimilar albeit possessing similar bioactivity. This phenomena is the converse of the similarity principle \footnote{In drug discovery, the similarity principle states that similar molecules $Sim(m_{1},m_{2})\ge\epsilon$ possess similar bioactivity $|f(m_{1})-f(m_{2})|\le\delta$. However, this does not prove the converse and many exceptions exist that are more difficult to model.\label{fn:sim_principle}}. Note that activity cliffs are another example converse to the similarity principle that have been used to demonstrate more challenging scenarios for predictive models \cite{kwapien2022implications, van2022exposing}. To test translation of the benchmark to more practically applicable objectives, we also test chemical exploration when optimizing the predicted A2A bioactvity which we call MolExpBio.

Each run is replicated on the benchmark 5 times.

\paragraph{Oracles}

The benchmark consists of four tasks. Each task is based on a set of drug candidates and their shared biological target, including anti-psychotic drugs (AP), Adenosine A$_{2A}$ (A2A) receptor drug candidates, beta-secretase 1 (BACE1) drug candidates, and epidermal growth factor drug candidates (EGFR). These tasks contain a set of 2, 3, 3, \& 4 target molecules respectively, where the objective is to rediscover the full set of target molecules.
This is similar in prospective drug discovery projects where 1 drug candidate will be progressed, but ideally with a small number of backup candidates.
To mimic a chemical space where distinct regions of reward signal exist, the oracle returns the similarity to the nearest target molecule as the reward $R(m)=max(sim(m_{i},m_{t_{1}}), ..., sim(m_{i},m_{t_{N}}))$. As CLMs act in string space, we chose a string-edit similarity function based on Levenshtein distance (MolExpL). However, we additionally report the results when using fingerprint similarity (MolExp) in the \chreplaced{Supporting Information}{Appendix}. A more detailed description of the molecular targets chosen and the similarity function can be found in \chreplaced{Section S2}{Appendix B}. For the MolExpBio experiment, the oracle is the predicted probability of A2A bioactivity, predicted by a Random Forest classification model trained on ECFP4 fingerprints of actives and inactives extracted from ChEMBL \cite{thomas2023pidginv5} made available in MolScore \cite{thomas2024molscore}. 

\paragraph{Metric}

Performance is measured by calculating the product of the maximum similarity achieved to each target molecule $\prod_{t \in T}max(sim(m_{i},m_{t}))$ . This metric focuses on the ability to generate similar molecules to all objectives.
Molecular diversity is also measured by sphere exclusion diversity (SEDiv) at a sample size of 1,000 \cite{thomas2021comparison} which provides a more accurate representation of chemical space coverage than internal diversity \cite{xie2021much, renz2024diverse}. Note this is equivalent to the later published \#Circles \cite{xie2021much} metric at a Tanimoto threshold of 0.65 and normalized by the sample size.

\paragraph{Baseline performance}

\begin{figure}[ht!]
%\vskip 0.2in
\begin{center}
\centerline{\includegraphics[width=\columnwidth]{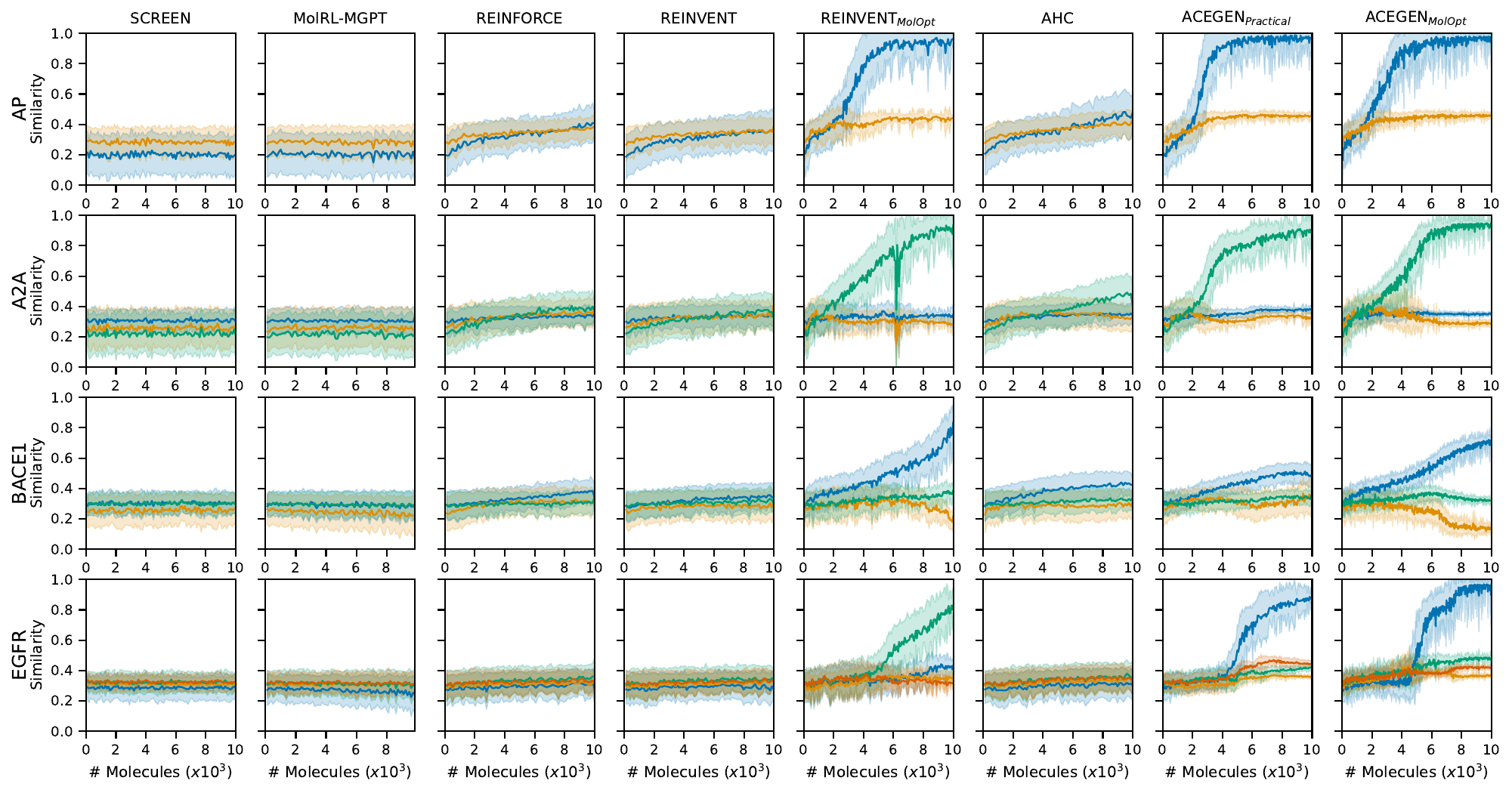}}
\caption{Baseline performance for each MolExpL task during RL training, single replicate. Each line color represents similarity to a target molecule. Note that ACEGEN and REINVENT$_{MolOpt}$ methods outperform due to their enhanced ability to optimize similarity to at-least one target molecule. Interestingly REINVENT$_{MolOpt}$ optimizes similarity to a different target molecule than ACEGEN for the EGFR task, despite identical initial policy parameterization.}
\label{fig:MolExpL_baseline_sim}
\end{center}
%\vskip -0.2in
\end{figure}

We tested virtual screening, MolRL-MGPT \cite{hu2024novo}, and multiple ACEGEN RL algorithms \cite{bou2024acegen, thomas2025reinforcing} on their inherent ability for chemical exploration on the MolExpL benchmark. Note that MolRL-GPT was pre-trained on the same dataset for standardized comparison (see \chreplaced{Section S1}{Appendix A} for pre-training details). \autoref{tab:MolExpL_baseline_performance} shows that with a budget of 10,000 the ACEGEN$_{MolOpt}$ algorithm achieved the highest score, therefore, we use this RL configuration for all further experiments. However, \autoref{fig:MolExpL_baseline_sim} illustrates that none of the tested algorithms optimize all molecular targets within the default budget (for a single replicate), and that the best scores are due to improved efficiency in maximizing similarity to a single target molecule. Therefore, there is still considerable room for improvement. These results are also observed when using fingerprint similarity oracles (see \chreplaced{Section S4.2}{Appendix D.2}). 

\begin{table*}
\centering
\small
\caption{Baseline performance on the MolExpL benchmark. The ACEGEN$_{MolOpt}$ RL configuration performs best in 3/4 tasks which we carry forward for scaling experiments.}
\label{tab:MolExpL_baseline_performance}
\resizebox{\textwidth}{!}{%
\begin{tabular}{l|cccccccc}
\hline
Task & SCREEN & MolRL-MGPT & REINFORCE & REINVENT & REINVENT$_{MolOpt}$ & AHC & ACEGEN$_{Practical}$ & ACEGEN$_{MolOpt}$ \\
\hline
AP & 0.53 ± 0.04 & 0.48 ± 0.03 & 0.56 ± 0.05 & 0.57 ± 0.04 & 0.63 ± 0.02 & 0.62 ± 0.06 & \textbf{0.64 ± 0.02} & \textbf{0.64 ± 0.04} \\
A2A & 0.38 ± 0.03 & 0.37 ± 0.03 & 0.36 ± 0.02 & 0.39 ± 0.08 & 0.38 ± 0.03 & 0.40 ± 0.02 & 0.40 ± 0.07 & \textbf{0.44 ± 0.06} \\
BACE1 & \textbf{0.29 ± 0.08} & 0.20 ± 0.02 & 0.23 ± 0.02 & 0.21 ± 0.01 & 0.27 ± 0.03 & 0.23 ± 0.01 & 0.23 ± 0.03 & 0.28 ± 0.03 \\
EGFR & 0.17 ± 0.04 & 0.17 ± 0.01 & 0.15 ± 0.02 & 0.16 ± 0.02 & 0.22 ± 0.03 & 0.17 ± 0.03 & 0.17 ± 0.02 & \textbf{0.25 ± 0.05} \\
\hline
Sum & 1.36 ± 0.10 & 1.21 ± 0.05 & 1.30 ± 0.06 & 1.33 ± 0.09 & 1.50 ± 0.06 & 1.41 ± 0.07 & 1.44 ± 0.08 & \textbf{1.62 ± 0.09} \\
\hline
\end{tabular}%
}
\end{table*}

\paragraph{Estimated max}

An estimated maximum performance can be calculated for ACEGEN$_{MolOpt}$ RL configuration by training the agent to maximize only the similarity to each individual molecular target with the reward $R(m)=sim(m_i, m_t)$, (see \chreplaced{Section S4.1}{Appendix D.1}). This tests the agents ability to maximize similarity to each individual molecular target only, an expected prerequisite for maximizing similarity to multiple target molecules in the benchmark. Calculating the equivalent benchmark score results in an estimated maximum MolExpL score of 2.91. 

\paragraph{Comparison to GuacaMol}

For reference, baseline algorithms were tested on the GuacaMol benchmark also with a fixed budget of 10,000 molecules. This shows ACEGEN$_{MolOpt}$ achieving the highest score (see Supporting Infomation). However, a key difference is that 5/6 GuacaMol similarity and rediscovery tasks are already solved by ACEGEN$_{MolOpt}$ with a score $\ge 0.9$, whilst the MolExpL benchmark results in lower scores reflecting a higher difficulty of benchmark tasks.

\subsection{Test-time training scaling for chemical exploration}
\begin{figure*}[ht]
     \centering
     \begin{subfigure}[t]{0.32\textwidth}
         \centering
         \includegraphics[width=\textwidth]{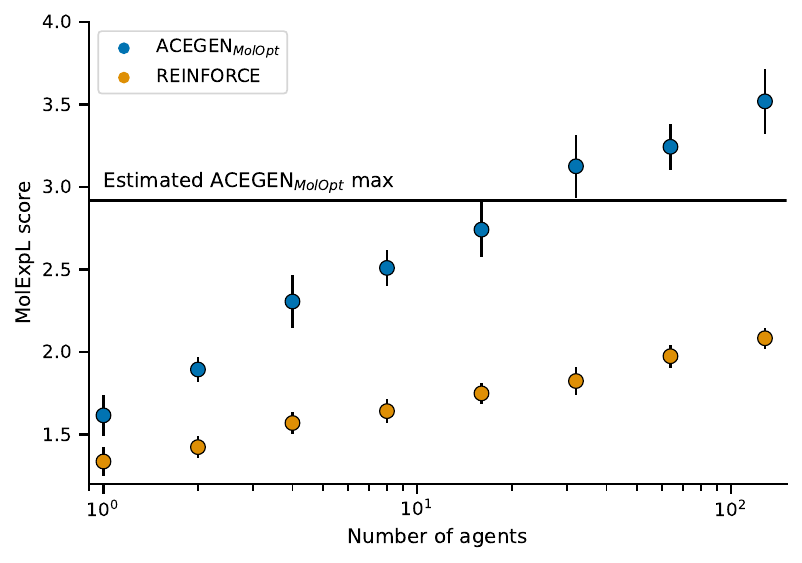}
         \caption{Scaling RL agents}
         \label{fig:MolExpL_scaling_agents}
    \end{subfigure}
    \begin{subfigure}[t]{0.32\textwidth}
         \centering
         \includegraphics[width=\textwidth]{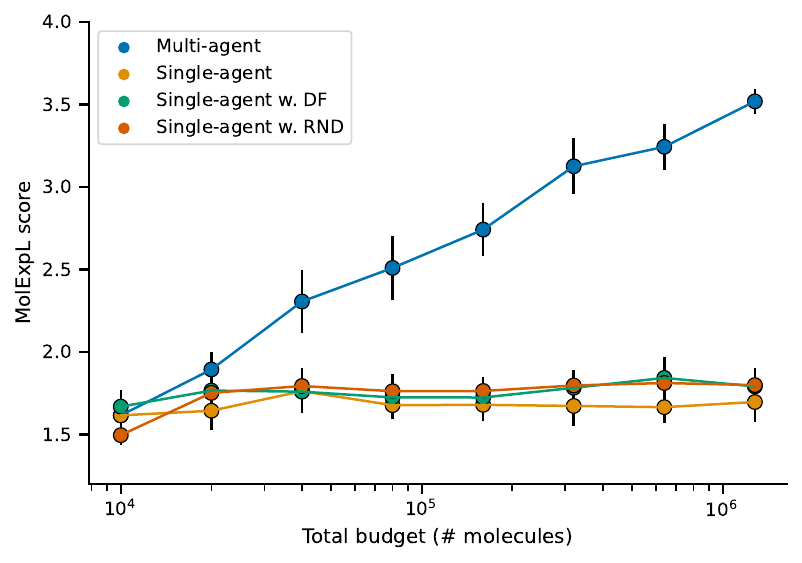}
         \caption{Scaling budget}
         \label{fig:MolExpL_scaling_budget}
    \end{subfigure}
     \begin{subfigure}[t]{0.32\textwidth}
         \centering
         \includegraphics[width=\textwidth]{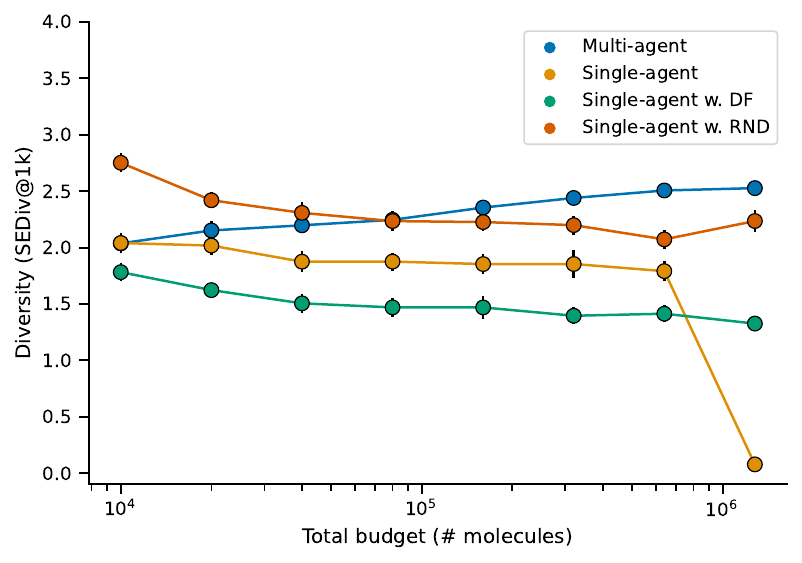}
         \caption{Molecular diversity}
         \label{fig:MolExpL_scaling_diversity}
     \end{subfigure}
     \caption{Performance on the MolExpL benchmark with scaling. (a) Scaling the number of independent ACEGEN$_{MolOpt}$ and REINFORCE RL agents, each with a budget of 10,000. (b) Scaling the total budget allocated to a single agent, a single agent with a RND exploration bonus, and a single agent with a DF. (c) The diversity of sampled compounds as measured by sphere exclusion diversity.}
    
\end{figure*}

We tested if scaling TTT with larger population sizes of independent RL agents would increase chemical exploration and improve benchmark performance. In this case, we chose the best-performing RL configuration-ACEGEN$_{MolOpt}$-and assigned a budget of 10,000 molecules to each agent in the population. \autoref{fig:MolExpL_scaling_agents} shows that MolExpL performance increases log-linearly from $\sim1.6$ with 1 agent to $\sim3.5$ with 128 agents, almost reaching a maximum score of 4. Log-linear scaling was also confirmed using the vanilla REINFORCE RL configuration, albeit with lower benchmark scores. Interestingly, when the population size reached 32 the estimated maximum score for this RL configuration was exceeded, indicating that the additional variance between independent agents overcomes the additionally reward signal ambiguity introduced from multiple molecular targets. \chreplaced{Section S5.1}{Appendix E.1} shows that by increasing agents, there is more variance across the set of molecule targets. The AP task is an exception, where all 128 independent agents all maximize similarity to the same target molecule in the set. Similar results are also observed when using fingerprint similarity oracles as shown in \chreplaced{Section S5.2}{Appendix E.2}. Additionally, similar scaling was achieved with up to 32 agents on the GuacaMol benchmark (\chreplaced{Section S5.3}{Appendix E.3}), at which point performance on many tasks becomes saturated.

\begin{figure}[ht]
%\vskip 0.2in
\begin{center}
\centerline{\includegraphics[width=0.9\columnwidth]{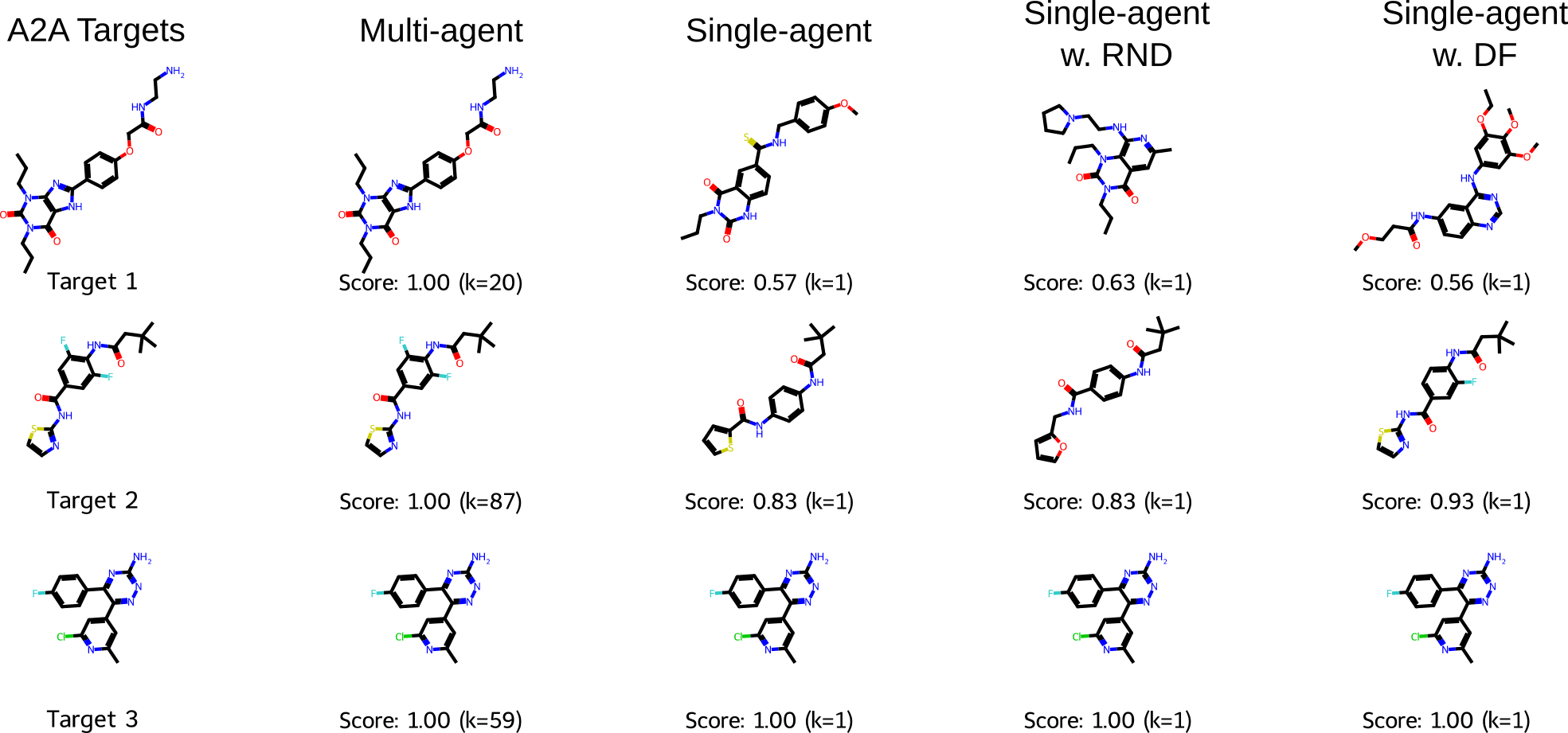}}
\caption{Example molecules generated compared to the set of targets in the MolExpL A2A task. The maximum similarity score and corresponding agent $k$ is labeled. Note that only with 87 agents is the task solved.}
\label{fig:A2A_example}
\end{center}
%\vskip -0.2in
\end{figure}

Increasing the number of agents, each with a budget of 10,000, increases the total budget. Therefore, we compared the alternative approach of scaling TTT via increasing the budget for a single RL agent. However, this did not result in similar MolExpL performance gain as shown in \autoref{fig:MolExpL_scaling_budget}. Single RL agent performance saturates at a budget of 40,000 as the agent expectedly fails to continue exploring once one of the target molecules has been maximized (Figure E.2). Moreover, \autoref{fig:MolExpL_scaling_diversity} shows that molecular diversity decreases while scaling budget. Employing RND as a state-of-the-art exploration strategy, the most performant exploration strategy \cite{svensson2024diversity, thomas2025reinforcing}, marginally improves performance and molecular diversity. Where Figure E.3 shows that RND only results in fluctuations in similarity to one molecular target, as opposed to the desired behavior of switching from one molecular target in the set to another. We additionally tested a simple diversity filter (DF) that assigns a reward of 0 to non-unique canonical SMILES. However, the DF also failed to rescue TTT scaling in this dimension. \autoref{fig:A2A_example} exemplifies how 87 agents are required to solve the A2A task and recover all target molecules, while scaling one agent results in a focus only on the third target molecule.

\subsection{Cooperative multi-agent RL for targeted chemical exploration}
Cooperation within a population of RL agents can theoretically improve performance by leveraging useful information shared between agents, such as, states already visited and their corresponding reward \cite{oroojlooy2023review}. We explored several strategies to leverage cooperative learning to improve the efficiency of chemical exploration. Considering the maximum number of molecular targets in a task set was 4, we investigated strategies utilizing 4 RL agents. Where, the best-case scenario is each agent maximizing similarity to a different molecular target in the set. 

\begin{figure*}[ht]
     \centering
     \begin{subfigure}[l]{\textwidth}
         \centering
         \includegraphics[width=\textwidth]{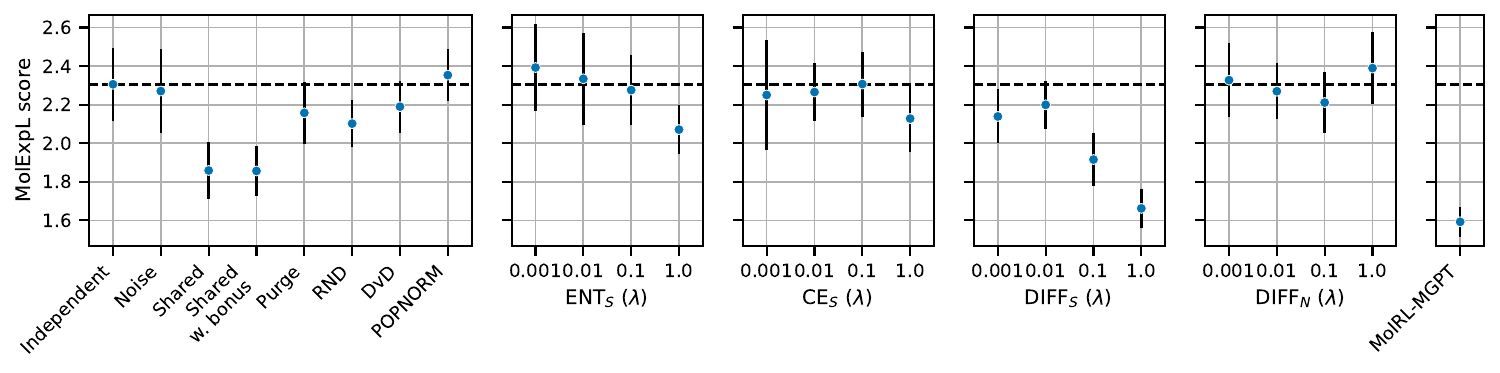}
         \caption{Cooperative RL MolExpL benchmark performance}
         \label{fig:MolExpL_coop_score}
     \end{subfigure}
     \begin{subfigure}[l]{\textwidth}
         \centering
         \includegraphics[width=\textwidth]{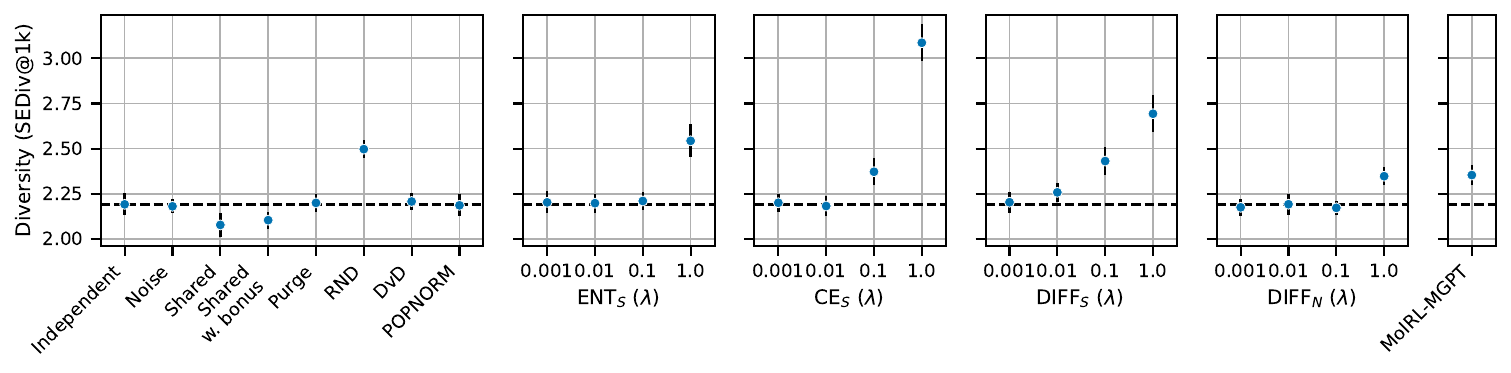}
         \caption{Cooperative RL molecular diversity}
         \label{fig:MolExpL_coop_diversity}
     \end{subfigure}
    \caption{Performance comparison of \chadded{(a) MolExpL Score and (b) Molecular diversity of} different 4-agent cooperative strategies on the MolExpL benchmark, each with a budget of 10,000. The dashed line represents the average of 4 independent agents as baseline.
    }
\end{figure*}

We investigated an extensive range of the different cooperative strategies on the MolExpL benchmark, including the MolRL-MGPT baseline with the same budget and number of agents, shown in \autoref{fig:MolExpL_coop_score}. All of the strategies tested were similar or worse in performance, with the best being POPNORM, ENT$_{\mathcal{S}}$, and DIFF$_{\mathcal{N}}$. However, no strategy significantly outperformed independent agents as measured by a Bonferroni-corrected, one-tailed t-test. On the other hand, \autoref{fig:MolExpL_coop_diversity} shows that many cooperative strategies successfully resulted in corresponding increases in molecular diversity, namely RND, ENT$_{\mathcal{S}}$, CE$_{\mathcal{S}}$, and DIFF$_\mathcal{S}$. This is usually anti-correlated with MolExpL score. This highlights the difficulty and complexity in achieving targeted exploration, where additional diversity appears to hinder learning, or is naive rather than targeted on favorable regions of chemical space. These trends were even more prominent when using fingerprint similarity oracles (see \chreplaced{Section S}{Appendix S7.2}).

Assessing the example behavior of different agents shows that some cooperative strategies do result in divergent behavior, but to the detriment of learning efficiency by slowing or inhibiting the learning of subsequent agents in a population (see \chreplaced{Section S7}{Appendix G}). \chadded{It is hypothesized that this is due to the intra-agent ``repellent'' nature of many of the cooperative strategies tested--especially during the early stages of optimization when no high reward region has been found yet.}

\subsection{Maximizing predicted A2A bioactivity}
To represent a practically applicable objective function, we additionally tested TTT scaling behavior when maximizing the predicted probability of A2A bioactivity as measured by the QSAR classification model from MolScore \cite{thomas2024molscore}. Increasing chemical exploration during optimization \textit{should} increase the probability of recovering the set of known A2A drug candidates used in the MolExp A2A task, two of which are present in the training dataset positive class. Therefore, we use the same metric to measure the maximimum similarity achieved to the A2A target molecule set, we refer to this additional experiment as MolExpBio. However, this now depends on the properties of the QSAR model and therefore, is not theoretically guaranteed.

\begin{figure*}[ht]
     \centering
     \begin{subfigure}[t]{0.32\textwidth}
         \centering
         \includegraphics[width=\textwidth]{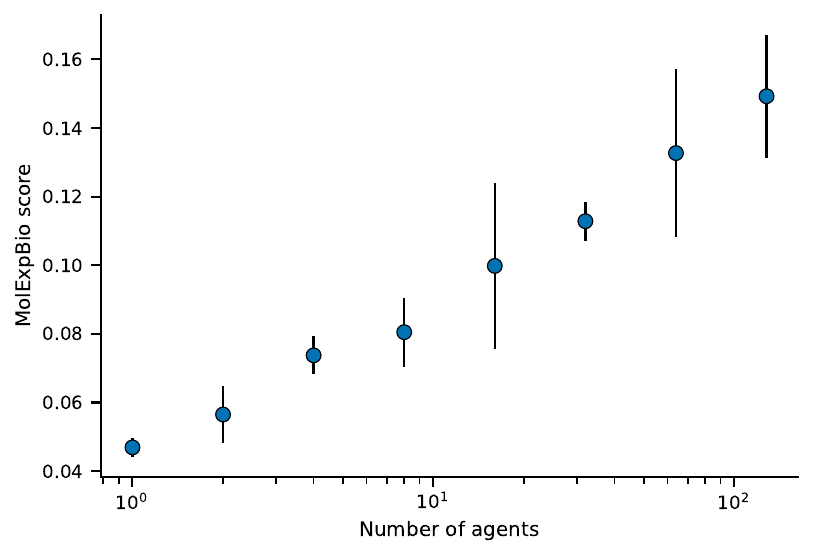}
         \caption{Scaling RL agents}
         \label{fig:MolExpBio_scaling_agents}
    \end{subfigure}
    \begin{subfigure}[t]{0.32\textwidth}
         \centering
         \includegraphics[width=\textwidth]{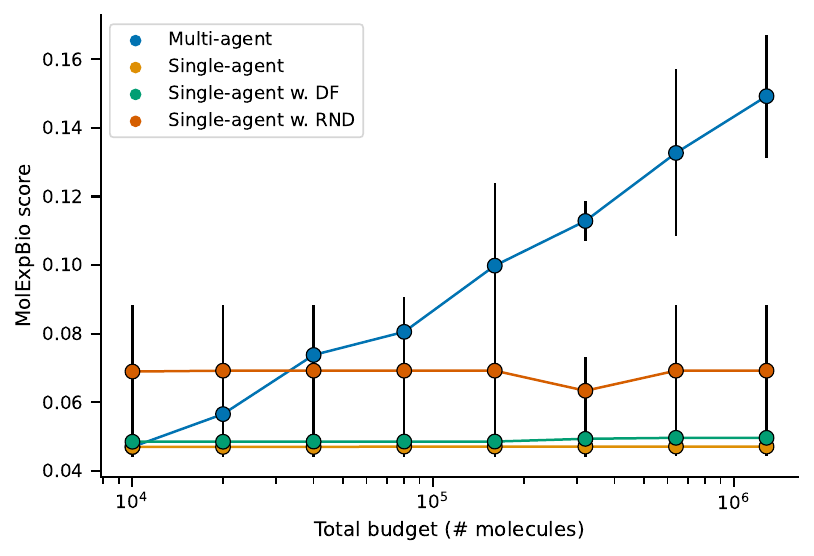}
         \caption{Scaling budget}
         \label{fig:MolExpBio_scaling_budget}
    \end{subfigure}
     \begin{subfigure}[t]{0.32\textwidth}
         \centering
         \includegraphics[width=\textwidth]{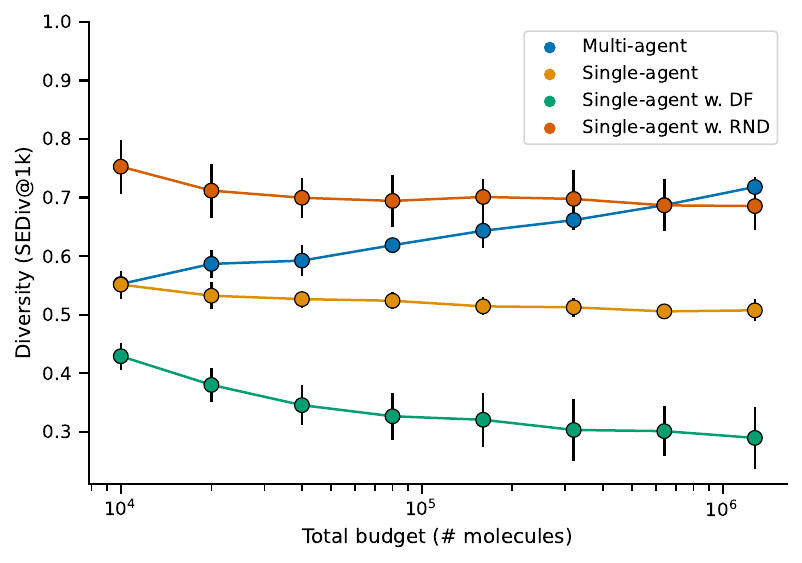}
         \caption{Molecular diversity}
         \label{fig:MolExpBio_scaling_diversity}
     \end{subfigure}
     \caption{Performance on the MolExpBio task with scaling. (a) Scaling the number of independent ACEGEN$_{MolOpt}$ and REINFORCE RL agents, each with a budget of 10,000. (b) Scaling the total budget allocated to a single agent, a single agent with a RND exploration bonus, and a single agent with a DF. (c) The diversity of sampled compounds as measured by sphere exclusion diversity.}
\end{figure*}

\autoref{fig:MolExpBio_scaling_agents} shows that scaling the number of independent RL agents increases the benchmark score, again more than scaling the budget of a single RL agent (\autoref{fig:MolExpBio_scaling_budget}). In other words, scaling RL agents increases the maximum similarity to the set of target molecules, and hence the chance of rediscovering the drug candidates. This represents increased chemical exploration of the reward landscape. This task highlights the translational benefit of TTT scaling with RL agents to real-world drug design objectives. 

%%%%%%%%%%%%%%%%%%%%%%%%%%%%%%%%%%%%%%%%%%%%%%%%%%
% Conclusion and impact statment
%%%%%%%%%%%%%%%%%%%%%%%%%%%%%%%%%%%%%%%%%%%%%%%%%%
\section{Conclusion}
% Summary of what was presented in the paper
In this paper, we probe the exploration of chemical space with a CLM to identify multiple candidate molecules in the context of drug discovery. In this effort, we introduce MolExp as a new benchmark, showing poor intrinsic exploration of current baseline algorithms. We showed that scaling a population of independent RL agents was the best approach to increasing benchmark performance, practically solving it with 128 agents. This included comparisons to scaling budgets for single RL agents and an extensive range of cooperative strategies. 

% Relation to TTT
Moreover, we found that performance improved log-linearly, similarly observed with test-time inference scaling of LLMs. In this regard, training the pre-trained CLM agent with RL on a specific test-time task (each of the four benchmark tasks) can be viewed as TTT. We than scale the whole TTT process by increasing the number of RL agents, resulting in similar scaling laws observed with test-time inference scaling in LLMs.

% Impact and outlook
Overall, this work takes an important step towards identifying \textit{all} possible molecules of interest in drug-like chemical space. Our MolExp benchmark is the first to explicitly ask this question, which we hope will guide algorithmic improvement until chemical space can be efficiently and thoroughly explored for a given context. TTT scaling through populations of RL agents are a promising solution towards this future.  

\subsection*{Limitations}
% Limitations
The similarity functions used for the tasks are unlikely to be used in real-world drug design. In practice, predictive models of bioactivity, binding affinity, and ADMET properties are more common. However, the set of \textit{all} the best possible solutions for such scoring functions (e.g., molecular docking) are currently unknowable. For this reason, MolExp(L) implements similarity functions as perfect scoring functions and tasks with known ground truths. We acknowledge this difference and strengthen the translational assumption with similar empirical results on other established benchmarks and a real-world objective in A2A bioactivity prediction. 

We did not investigate the effect of model architecture and training dataset in this work, which has been thoroughly investigated elsewhere \citep{skinnider2021chemical, ozccelik2024chemical}. Instead we focus on maximizing chemical exploration efficiency given the same or similar starting policy. These results are expected to translate to any starting policy.

The caveat of scaling TTT with more RL agents or increased budget is increased walltime requirements, as shown in \chreplaced{Section S9}{Appendix I}. This reiterates the importance of trying to use cooperative methods to improve exploration efficiency, as well as, the use of this benchmark and results as a tool for further research into targeted and efficient chemical space exploration.

\subsection*{Data and Software Availability}
All code is openly available under an MIT license. The CLM-based RL agents were implemented in ACEGEN, available on GitHub at \url{https://github.com/Acellera/acegen-open}, whilst the benchmark was implemented in MolScore available on GitHub at \url{https://github.com/MorganCThomas/MolScore} or in the Python Package Index \url{https://pypi.org/project/MolScore/}. Note the MolExp pretraining dataset is also available in the MolScore repository.

\subsection{Supporting Information}
The supporting information file includes the Appendices referred to throughout the text.

\subsection{Author Contributions}
\chadded{MT and AB experimented with, and implemented different cooperative RL strategies. MT wrote the manuscript and conducted the experiments detailed within, under the supervision of GDF.}

\subsection{Conflict of Interest}
\chadded{The authors declare no competing financial interests.}

\subsection{Acknowledgments}
This work was funded in part by the Flanders innovation \& entrepreneurship (VLAIO) project HBC.2021.1123

%%%%%%%%%%%%%%%%%%%%%%%%%%%%%%%%%%%%%%%%%%%%%%%%%%
% References
%%%%%%%%%%%%%%%%%%%%%%%%%%%%%%%%%%%%%%%%%%%%%%%%%%
\bibliography{refs}

\end{document}

\tableofcontents

%%%%%%%%%%%%%%%%%%%%%%%%%%%%%%%%%%%%%%%%%%%%%%%%%%%%%%%%%%%%%%%%%%%%%%%%%%%%%%%
%%%%%%%%%%%%%%%%%%%%%%%%%%%%%%%%%%%%%%%%%%%%%%%%%%%%%%%%%%%%%%%%%%%%%%%%%%%%%%%
% Supporting Information
%%%%%%%%%%%%%%%%%%%%%%%%%%%%%%%%%%%%%%%%%%%%%%%%%%%%%%%%%%%%%%%%%%%%%%%%%%%%%%%
%%%%%%%%%%%%%%%%%%%%%%%%%%%%%%%%%%%%%%%%%%%%%%%%%%%%%%%%%%%%%%%%%%%%%%%%%%%%%%%
\newpage
\onecolumn
%%%%%%%%%%%%%%%%%%%%%%%%%%%%%%%%%%%%%%%%%%%%%%%%%%%%%%%%%%%%%%%%%%%%%%%%%%%%%%%
\section{Model pre-training}\label{app:CLM}
\subsection{Datasets}

\textbf{MolExp benchmark} The pre-training dataset for the MolExp(L) benchmark was curated from the ChEMBL34 database \cite{gaulton2012chembl}. First, the database was filtered to ensure no molecules existed with more 10 rotatable bonds, a logP above 5.5, a molecular weight outside of 150-650 Da, or contained atoms not in a comment set $a \in {C, N, S, O, F, Cl, Br, H}$. All molecules were neutralized if possible, converted to non-isomeric SMILES (i.e., containing no stereoinformation) and de-duplicated. This resulted in a dataset of 1,711,022 unique molecules. Finally, the target molecules of the MolExp benchmark \autoref{app:MolExp} were purposefully confirmed to be present within the training dataset, to ensure that objectives were theoretically achievable.

\textbf{GuacaMol benchmark} The pre-training dataset for the GuacaMol \cite{brown2019guacamol} benchmark was taken directly from the original publication without any further processing. This dataset was downloaded from \url{https://figshare.com/projects/GuacaMol/56639}.

\subsection{ACEGEN}

\begin{figure}[ht]
%\vskip 0.2in
\begin{center}
\centerline{\includegraphics[width=\textwidth]{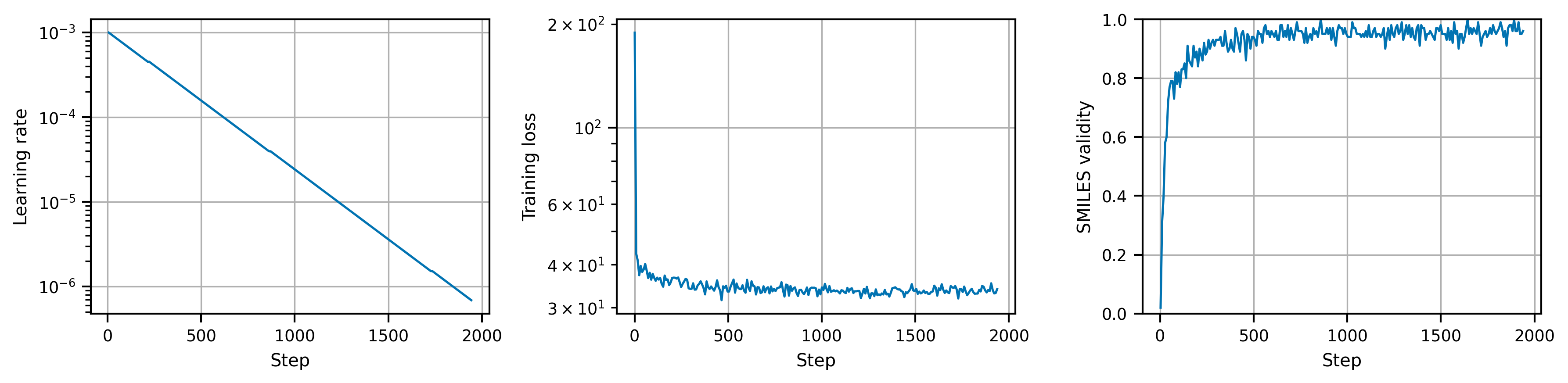}}
\caption{Pre-training of the GRU CLM on the MolExp benchmark pre-training dataset curated from ChEMBL34.}
\label{fig:clm_pretraining}
\end{center}
%\vskip -0.2in
\end{figure}

\textbf{MolExp benchmark} The CLM constituting the prior policy is a gated recurrent unit (GRU) network with an embedding of 256 and 3 layers of GRU cells with hidden dimension 512. This was trained with restricted SMILES randomization \cite{arus2019randomized} for 10 epochs with a batch size of 128. This was implemented such that a random SMILES string augmentation was applied each time a dataset molecule was sampled during training. An ADAM optimizer was employed with a learning rate of 0.001 and step scheduler that dropped the learning rate every 500 steps. The learning rate, training loss, and SMILES validity during training is shown in \autoref{fig:clm_pretraining}.

\textbf{GuacaMol benchmark} The CLM constituting the prior policy is an long short-term memory (LSTM) network with the same hyperparameters as the model used in GuacaMol benchmark. This includes an embedding of 1024 and 3 layers of LSTM cells with hidden dimension 1024, as well as a dropout rate of 0.2. The pre-trained weights were loaded from the GuacaMol benchmark resulting in the exact same prior and initial policy.

\subsection{MolRL-MGPT}
\begin{figure}[t]
%\vskip 0.2in
\begin{center}
\centerline{\includegraphics[width=\textwidth]{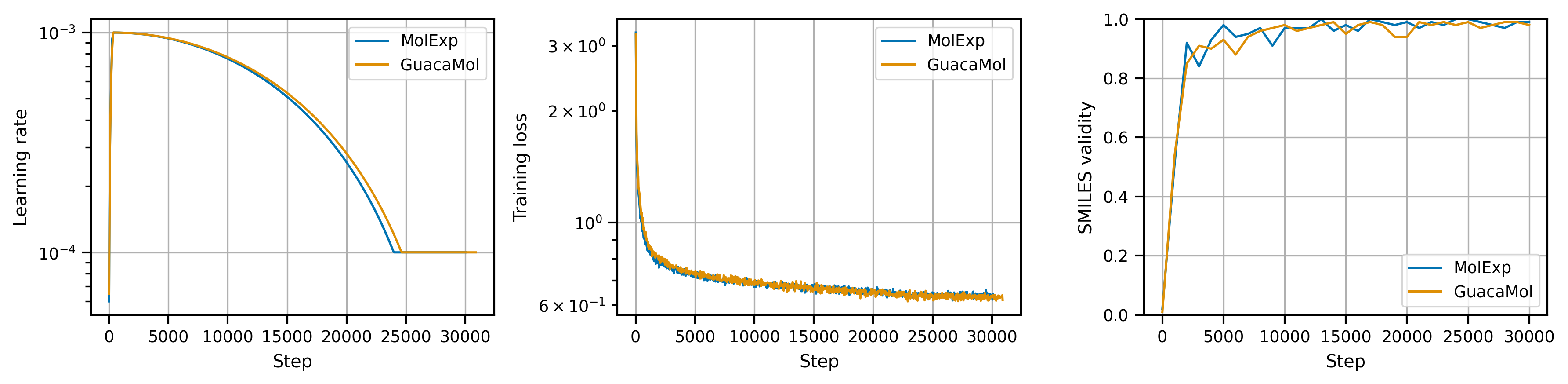}}
\caption{Pre-training of the MolRL-MGPT CLM on the MolExp benchmark pre-training dataset and GuacaMol benchmark pre-training dataset.}
\label{fig:molrl-mgpt_pretraining}
\end{center}
%\vskip -0.2in
\end{figure}

The MolRL-MGPT model was re-implemented following the code provided from the original publication \cite{hu2024novo}. This CLM constituting the prior policy is a GPT-style Transformer model with 8 layers each with 8 attention heads. For a standardized comparison, the model was pre-trained on the benchmark training datasets respectively. For hardware reasons the batch size was lowered from 2048 to 512, with all other hyperparameters kept equal including training for 10 epochs. The learning rate, training loss, and SMILES validity during training is shown in \autoref{fig:molrl-mgpt_pretraining}. 

%%%%%%%%%%%%%%%%%%%%%%%%%%%%%%%%%%%%%%%%%%%%%%%%%%%%%%%%%%%%%%%%%%%%%%%%%%%%%%%
\clearpage
\section{Molecular exploration benchmark}\label{app:MolExp}
\begin{figure}[ht]
%\vskip 0.2in
\begin{center}
\centerline{\includegraphics[width=0.5\textwidth]{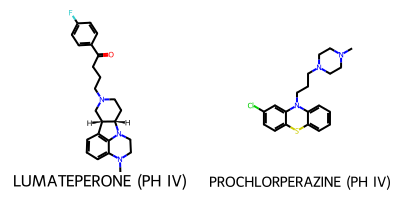}}
\caption{Antipsychotic (AP) drugs as molecular targets. Both are clinically approved antipsychotics, however, Lumateperone was approved in 1956, while Prochlorperazine was the most recently approved in 2019.}
\label{fig:AP_mols}
\end{center}
%\vskip -0.2in
\end{figure}

\begin{figure}[ht]
%\vskip 0.2in
\begin{center}
\centerline{\includegraphics[width=0.75\textwidth]{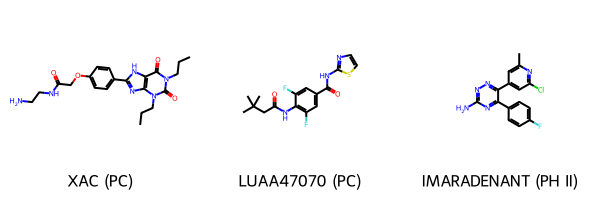}}
\caption{Adenosine A$_{2A}$ (A2A) receptor drug candidates as molecular targets. All the molecular candidates are A$_{2A}$ receptor antagonists that bind to the same orthosteric site but with alternative binding modes. XAC binds from N253 up towards the extracellular side with a polar tail that extends into solvent,  LUAA47070 binds via a water mediated interaction with N253 (note we use active metabolite, desphospho-structure), while Imaradenant binds from N253 down towards the intrahelical bundle occupying lipophilic hotspots.}
\label{fig:A2A_mols}
\end{center}
%\vskip -0.2in
\end{figure}

\begin{figure}[ht]
%\vskip 0.2in
\begin{center}
\centerline{\includegraphics[width=0.75\textwidth]{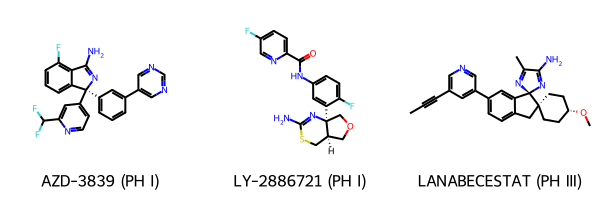}}
\caption{Beta-secretase I (BACE1) receptor drug candidates as molecular targets. All are high-affinity inhibitors of BACE1 receptor with a common functional amidine/hiourea group to bind to the dual Asparctic acids in the pocket but are otherwise topologically different.}
\label{fig:BACE1_mols}
\end{center}
%\vskip -0.2in
\end{figure}

\begin{figure}[ht]
%\vskip 0.2in
\begin{center}
\centerline{\includegraphics[width=\textwidth]{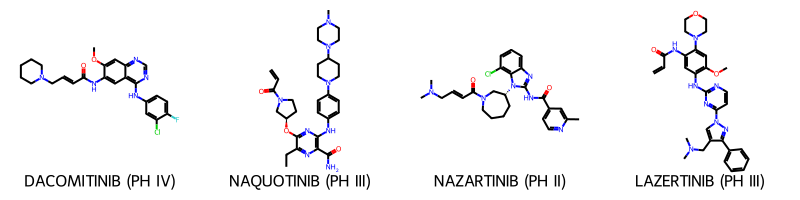}}
\caption{Epidermal growth factor (EGFR) drug candidates as molecular targets. All molecules are high-affinity covalent inhibitors of EGFR, sharing the reactive acrylamide functional group to bind to cysteine. Otherwise they are the four most topologically diverse molecules out of the 13 sub-100nM activity, at least phase I EGFR inhibitors found in ChEMBL34.}
\label{fig:EGFR_mols}
\end{center}
%\vskip -0.2in
\end{figure}

We selected four tasks, each with 2-4 objectives as molecular targets to be rediscovered based on real-world drug candidates ranging from pre-clinical (PC), clinical phases 1-3 (PH I-III) and approved (PH IV), shown in \autoref{fig:AP_mols} to \autoref{fig:EGFR_mols}. For each task, the goal is to maximize the reward $R(m)$ for a given molecule $m$, where the reward is similarity to the closest molecular target $m_t$:

\begin{equation}
R(m)=max(sim(m_{i},m_{t_{1}}), ..., sim(m_{i},m_{t_{N}}))
\end{equation}

In this work, the similarity function $sim$ is implemented as the Levenshtein similarity between the canonical SMILES strings (MolExpL). This was chosen because the CLM generates and acts in a SMILES environment. The Levenshtein similarity function was created by normalizing the Levenshtein distance by the length of the reference string, clipped to ensure the output value was bound between [0, 1]:

\begin{equation}
sim_\mathbb{L}(m, m_t) = clip(\frac{1-dis_\mathbb{L}(m,m_t)}{|m_t|})
\end{equation}

Note that the SMILES strings of both $m$ $m_t$ are canonicalized and both used non-isomeric SMILES. Moreover,we made an alternative similarity function available using the Tanimoto similarity of ECFP fingerprints for application with non-language models (MolExp), although the Levenshtein similarity will still work provided a SMILES representation is given.

Performance was evaluated based on the ability to generate molecules close to all molecular targets by taking the product:

\begin{equation}
\prod_{t \in T}\frac {\sum_{i \in N}sim(m_{i},m_{t})}{N}
\end{equation}

This formulation tests the generative models ability to maximize the reward and remain intrinsically curious and continue exploring - even when the expected reward may already be high.

%%%%%%%%%%%%%%%%%%%%%%%%%%%%%%%%%%%%%%%%%%%%%%%%%%%%%%%%%%%%%%%%%%%%%%%%%%%%%%%
\clearpage
\section{Baseline algorithms}\label{app:baseline_algorithm}

A description of baseline algorithms and in particular their update step are detailed below, followed by the hyperparameters for each algorithm, followed by a pseudo-algorithm describing the general RL rollout and update procedure for these single-agent baseline algorithms \ref{alg:single-agent}. Note this algorithm does not apply to MolRL-MGPT which is a multi-agent baseline.

\textbf{SCREEN:} This virtual screening baseline iteratively samples without replacement from the CLM training dataset \autoref{app:CLM} curated from ChEMBL34. Note that all the target molecules from MolExp benchmark are included in the dataset, therefore, it is possible to achieve a perfect score if they are sampled. However, this is unlikely given the budget of 10,000 molecules out of the possible 1.7M. 

\textbf{REINFORCE:} This baseline was implemented as the vanilla reinforce algorithm where the objective is to maximize the likelihood of sequences weighted by the cumulative reward $R(\tau)$. In this implementation we additionally applied experience replay to augment on-policy data with previously high rewarded molecules. All configuration parameters are shown in \autoref{table:reinforce_hp}. 

\begin{equation}
\nabla J(\theta) = \mathbb{E}_{\tau \sim \pi_{\theta}} \left[ \sum_{t=0}^{T} \nabla_\theta \log \pi_\theta(a_t|s_t) \cdot R(\tau) \right]
\end{equation}

\textbf{REINVENT:} This baseline was implemented following the REINVENT formulation \cite{olivecrona2017molecular} whereby a fixed version of the prior policy $\pi_{prior}$ (i.e., the pre-trained CLM) is used to regularize the updates of the agent policy $\pi_{agent}$ being trained. This is a form of reward shaping as shown in following \autoref{eq:reinvent}. This also introduces a hyperparameter $\sigma$ that determines the balance between $\pi_{prior} - \pi_{agent}$ difference or the reward $R(\tau)$. The hyperparameters used corresponding to the default frequently used in the literature are shown in \autoref{table:reinvent_hp}.

\begin{equation}\label{eq:reinvent}
R(\tau)_{reshaped} = \frac{(\pi_{prior} - \pi_{agent} + \sigma \cdot R(\tau))^2}{\pi_{agent}} 
\end{equation}

\textbf{REINVENT$_{MolOpt}$:} This baseline is a hyperparameter optimized configuration of  REINVENT for performance on the MolOpt benchmark. Notably this increases $\sigma$ to focus more on molecule reward higher than regularization to the prior and increases the amount of experience replay used. All configuration parameter are shown in \autoref{table:reinvent_molopt_hp}.

\textbf{AHC:} Augmented Hill-Climb \cite{thomas2022augmented} is a variation of REINVENT that conducts on-policy hill-climbing, i.e., on-policy molecules are ranked by reward $R(\tau)$ and only the top-$k$ molecules are retained. This shift focuses learning on high-reward molecules which improves efficiency, and helps to avoid too much regularization in low-reward scenarios. All configuration parameters are shown in \autoref{table:ahc_hp}.

\textbf{MolRL-MGPT:} MolRL-MGPT is a cooperative multi-agent RL baseline with 4 GPT-style CLM agents. This follows the REINVENT formulation with the addition of a second loss term for the $k$-th agent encouraging reward scaled deviation of the current agent to previous ones. This baseline was implemented using the code provided alongside the publication \cite{hu2024novo}. The default hyperparameters provided with the code were used, as there were insufficient details for full reproducibility of paper results.

\begin{equation}
\mathcal{L}^k_{DIFF_{\mathcal{N}}} = - \sigma_2 \cdot \sum_{i=1}^{k-1}R(\tau)\cdot \lvert \log\pi_k(\tau) - \log\pi_i(\tau) \rvert
\end{equation}

\clearpage

\begin{table}[!ht]
    \centering
    \caption{Hyperparameters for REINFORCE.}
    \vspace{0.2cm}
    \begin{tabular}{lc}
    \multicolumn{1}{c}{Hyperparameter} & \multicolumn{1}{c}{Value} \\ \toprule
    num\_envs & 128 \\ % Number of smiles to generate in parallel
    total\_smiles & 10,000 \\ % Total number of smiles to generate
    model & GRU (embedding of size 256 + 3 layer GRU of size 512 + MLP) \\
    lr & 0.0001 \\
    experience\_replay & True \\
    replay\_buffer\_size & 100 \\
    replay\_batch\_size & 10 \\
    \end{tabular}
    \label{table:reinforce_hp}
\end{table}

\begin{table}[!ht]
    \centering
    \caption{Hyperparameters for REINVENT.}
    \vspace{0.2cm}
    \begin{tabular}{lc}
    \multicolumn{1}{c}{Hyperparameter} & \multicolumn{1}{c}{Value} \\ \toprule
    num\_envs & 128 \\ % Number of smiles to generate in parallel
    total\_smiles & 10,000 \\ % Total number of smiles to generate
    model & GRU (embedding of size 256 + 3 layer GRU of size 512 + MLP) \\
    lr & 0.0001 \\
    experience\_replay & True \\
    replay\_buffer\_size & 100 \\
    replay\_batch\_size & 10 \\
    sigma & 120 \\
    \end{tabular}
    \label{table:reinvent_hp}
\end{table}

\begin{table}[!ht]
    \centering
    \caption{Hyperparameters for REINVENT$_{MolOpt}$.}
    \vspace{0.2cm}
    \begin{tabular}{lc}
    \multicolumn{1}{c}{Hyperparameter} & \multicolumn{1}{c}{Value} \\ \toprule
    num\_envs & 64 \\ % Number of smiles to generate in parallel
    total\_smiles & 10,000 \\ % Total number of smiles to generate
    model & GRU (embedding of size 256 + 3 layer GRU of size 512 + MLP) \\
    lr & 0.0005 \\
    experience\_replay & True \\
    replay\_buffer\_size & 100 \\
    replay\_batch\_size & 24 \\
    sigma & 500 \\
    \end{tabular}
    \label{table:reinvent_molopt_hp}
\end{table}

\begin{table}[!ht]
    \centering
    \caption{Hyperparameters for AHC.}
    \vspace{0.2cm}
    \begin{tabular}{lc}
    \multicolumn{1}{c}{Hyperparameter} & \multicolumn{1}{c}{Value} \\ \toprule
    num\_envs & 128 \\ % Number of smiles to generate in parallel
    total\_smiles & 10,000 \\ % Total number of smiles to generate
    model & GRU (embedding of size 256 + 3 layer GRU of size 512 + MLP) \\
    lr & 0.0001 \\
    experience\_replay & True \\
    replay\_buffer\_size & 100 \\
    replay\_batch\_size & 10 \\
    sigma & 60 \\
    topk & 0.5 \\
    \end{tabular}
    \label{table:ahc_hp}
\end{table}

\begin{table}[!ht]
    \centering
    \caption{Hyperparameters for ACEGEN$_{Practical}$.}
    \vspace{0.2cm}
    \begin{tabular}{lc}
    \multicolumn{1}{c}{Hyperparameter} & \multicolumn{1}{c}{Value} \\ \toprule
    num\_envs & 128 \\ % Number of smiles to generate in parallel
    total\_smiles & 10,000 \\ % Total number of smiles to generate
    model & GRU (embedding of size 256 + 3 layer GRU of size 512 + MLP) \\
    lr & 0.0001 \\
    experience\_replay & True \\
    replay\_batch\_size & 10 \\
    replay\_buffer\_size & 100 \\
    replay\_sampler & uniform \\
    sigma & 0.005 \\
    topk & 0.5 \\
    alpha & 5 \\
    baseline & mab \\
    \end{tabular}
    \label{table:acegen_hp_pract}
\end{table}

\begin{table}[!ht]
    \centering
    \caption{Hyperparameters for ACEGEN$_{MolOpt}$.}
    \vspace{0.2cm}
    \begin{tabular}{lc}
    \multicolumn{1}{c}{Hyperparameter} & \multicolumn{1}{c}{Value} \\ \toprule
    num\_envs & 32 \\ % Number of smiles to generate in parallel
    total\_smiles & 10,000 \\ % Total number of smiles to generate
    model & GRU (embedding of size 256 + 3 layer GRU of size 512 + MLP) \\
    lr & 0.0001 \\
    experience\_replay & True \\
    replay\_batch\_size & 50 \\
    replay\_buffer\_size & 100 \\
    replay\_sampler & prioritized \\
    sigma & 0.001 \\
    topk & 0.5 \\
    alpha & 3 \\
    baseline & False \\
    \end{tabular}
    \label{table:acegen_hp_molopt}
\end{table}

\begin{table}[!ht]
    \centering
    \caption{Hyperparameters for MolRL-MGPT. }
    \vspace{0.2cm}
    \begin{tabular}{lc}
    \multicolumn{1}{c}{Hyperparameter} & \multicolumn{1}{c}{Value} \\ \toprule
    num\_envs & 128 \\ % Number of smiles to generate in parallel
    total\_smiles & 10,000 \\ % Total number of smiles to generate
    model & GPT (embedding of size 256 + 8 layer with 8 attention heads + MLP) \\
    lr & 0.0001 \\
    experience\_replay & True \\
    replay\_batch\_size & 5 \\
    replay\_buffer\_size & 25 \\
    replay\_sampler & uniform \\
    sigma$_1$ & 100 \\
    sigma$_2$ & 0.5 \\
    \end{tabular}
    \label{table:molrl_mgpt}
\end{table}

\clearpage

\begin{algorithm}[H]
\begin{spacing}{1.2}
\DontPrintSemicolon
\SetAlgoLined
\label{alg:single-agent}
\caption{Single-Agent RL}
\KwIn{Stochastic policy $\pi_\theta$, horizon $T$, replay buffer $\mathcal{B}$, batch size $B$, (optional) sigma value $\sigma$}
%\KwOut{Trajectory batch $\tau$, updated policy $\pi_\theta$}

Initialize pretrained policy $\pi_\theta$ \;
Initialize replay buffer $\mathcal{B} \leftarrow \varnothing$ \;
Set counter $N \leftarrow 0$ \tcp*{total molecules sampled so far}
(Optional) Initialize frozen prior policy $\pi_{\text{prior}} \leftarrow \pi_\theta$ \;

\BlankLine
\While{$N < M$}{
    \tcp{(1) Rollout: generate on-policy batch}
    \For{$b=1$ \KwTo $B$}{
      $s_0 \leftarrow \texttt{BOS}$ \;
      \For{$t=0$ \KwTo $T-1$}{
        Sample action $a_t \sim \pi_\theta(\cdot \mid s_t)$ \;
        Update state $s_{t+1} \leftarrow \text{Append}(s_t,a_t)$ \;
        \If{$a_t = \texttt{EOS}$}{\textbf{break}}
      }
      Store trajectory $\tau^{(b)} = (s_0,a_0,\dots,s_T,a_T)$ \;
    }
    Denote $\tau_{\text{on}}^B = \{\tau^{(b)}\}_{b=1}^B$ \;
    Update counter $N \leftarrow N + |\tau_{\text{on}}^B|$ \;
    
    \BlankLine
    \tcp{(2) Augment with replay buffer samples}
    Sample $\tau_{\text{off}} \sim \mathcal{B}$ \;
    Form combined batch $\tau^B \leftarrow \tau_{\text{on}}^B \cup \tau_{\text{off}}$ \;
    
    \BlankLine
    \tcp{(3) Compute rewards via oracle}
    $R(\tau^B) \leftarrow f(\tau^B)$ \tcp*{May include DF penalty or RND bonus}
    
    \BlankLine
    \tcp{(4) Policy update using chosen method $\Psi$}
    $\pi_\theta \leftarrow \text{Update}_\Psi(\pi_\theta, \tau^B, R(\tau^B), [\pi_{\text{prior}}], [\sigma])$ \;
    
    \BlankLine
    \tcp{(5) Update replay buffer with top-performing trajectories}
    $\mathcal{B} \leftarrow \mathcal{B} \cup \text{Top-}k(\tau^B, R(\tau^B))$ \;
}
%\KwRet{$\pi_\theta$}
\end{spacing}
\end{algorithm}

%%%%%%%%%%%%%%%%%%%%%%%%%%%%%%%%%%%%%%%%%%%%%%%%%%%%%%%%%%%%%%%%%%%%%%%%%%%%%%%
\clearpage
\section{Baseline performance}\label{app:Baseline_performance}
\subsection{MolExpL}\label{app:MolExpL_baseline}

% ----- Moved to main
%\begin{figure}[ht!]
%\vskip 0.2in
%\begin{center}
%\centerline{\includegraphics[width=\columnwidth]{v2/images/MolExpL_baseline_similarity_seed0.pdf}}
%\caption{Baseline performance for each MolExpL task during RL training, single replicate. Each line color represents similarity to a target molecule. Note that ACEGEN and REINVENT$_{MolOpt}$ methods outperform due to their enhanced ability to optimize similarity to at-least one target molecule. Interestingly REINVENT$_{MolOpt}$ optimizes similarity to a different target molecule than ACEGEN for the EGFR task, despite identical initial policy parameterization.}
%\label{fig:MolExpL_baseline_sim}
%\end{center}
%\vskip -0.2in
%\end{figure}

\begin{figure}[ht!]
%\vskip 0.2in
\begin{center}
\centerline{\includegraphics[width=\columnwidth]{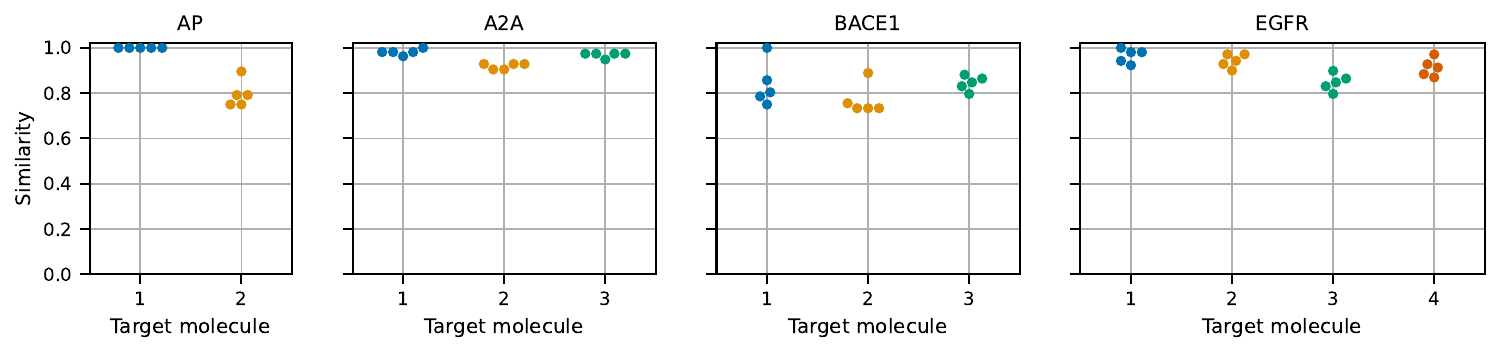}}
\caption{Estimated maximum performance achieved by ACEGEN$_{MolOpt}$. ACEGEN$_{MolOpt}$ was trained to maximize the similarity to each individual molecular target separately for a budget of 10,000 molecules. Based on this, the estimated maximum performance on the benchmark is 2.92.}
\label{fig:MolExpL_estimated_max}
\end{center}
%\vskip -0.2in
\end{figure}

%%%%%%%%%%%%%%%%%%%%%%%%%%%%%%%%%%%%%%%%%%%%%%%%%%%%%%%%%%%%%%%%%%%%%%%%%%%%%%%
\clearpage
\subsection{MolExp}\label{app:MolExp_baseline}

\begin{table}[ht!]
\centering
\small
\caption{Baseline performance on the MolExp benchmark.}
\label{tab:MolExp_baseline_performance}
\resizebox{\textwidth}{!}{%
\begin{tabular}{l|cccccccc}
\hline
 & SCREEN & MolRL-MGPT & REINFORCE & REINVENT & REINVENT$_{MolOpt}$ & AHC & ACEGEN$_{Practical}$ & ACEGEN$_{MolOpt}$ \\
\hline
AP & 0.32 ± 0.10 & 0.24 ± 0.05 & 0.37 ± 0.02 & 0.25 ± 0.04 & 0.45 ± 0.03 & \textbf{0.47 ± 0.04} & 0.40 ± 0.02 & 0.42 ± 0.04 \\
A2A & 0.08 ± 0.02 & 0.08 ± 0.01 & 0.09 ± 0.01 & 0.06 ± 0.01 & 0.12 ± 0.02 & 0.11 ± 0.03 & \textbf{0.13 ± 0.03} & \textbf{0.13 ± 0.05} \\
BACE1 & \textbf{0.16 ± 0.07} & 0.07 ± 0.01 & 0.05 ± 0.01 & 0.05 ± 0.01 & 0.06 ± 0.01 & 0.06 ± 0.01 & 0.06 ± 0.01 & 0.06 ± 0.01 \\
EGFR & 0.04 ± 0.01 & 0.03 ± 0.01 & 0.04 ± 0.01 & 0.03 ± 0.00 & 0.05 ± 0.02 & 0.04 ± 0.01 & 0.05 ± 0.02 & \textbf{0.06 ± 0.01} \\
\hline
Sum & 0.60 ± 0.12 & 0.42 ± 0.05 & 0.55 ± 0.02 & 0.39 ± 0.04 & 0.68 ± 0.05 & 0.68 ± 0.05 & 0.64 ± 0.04 & \textbf{1.07 ± 0.08} \\
\hline
\end{tabular}%
}
\end{table}

\begin{figure}[ht!]
%\vskip 0.2in
\begin{center}
\centerline{\includegraphics[width=\columnwidth]{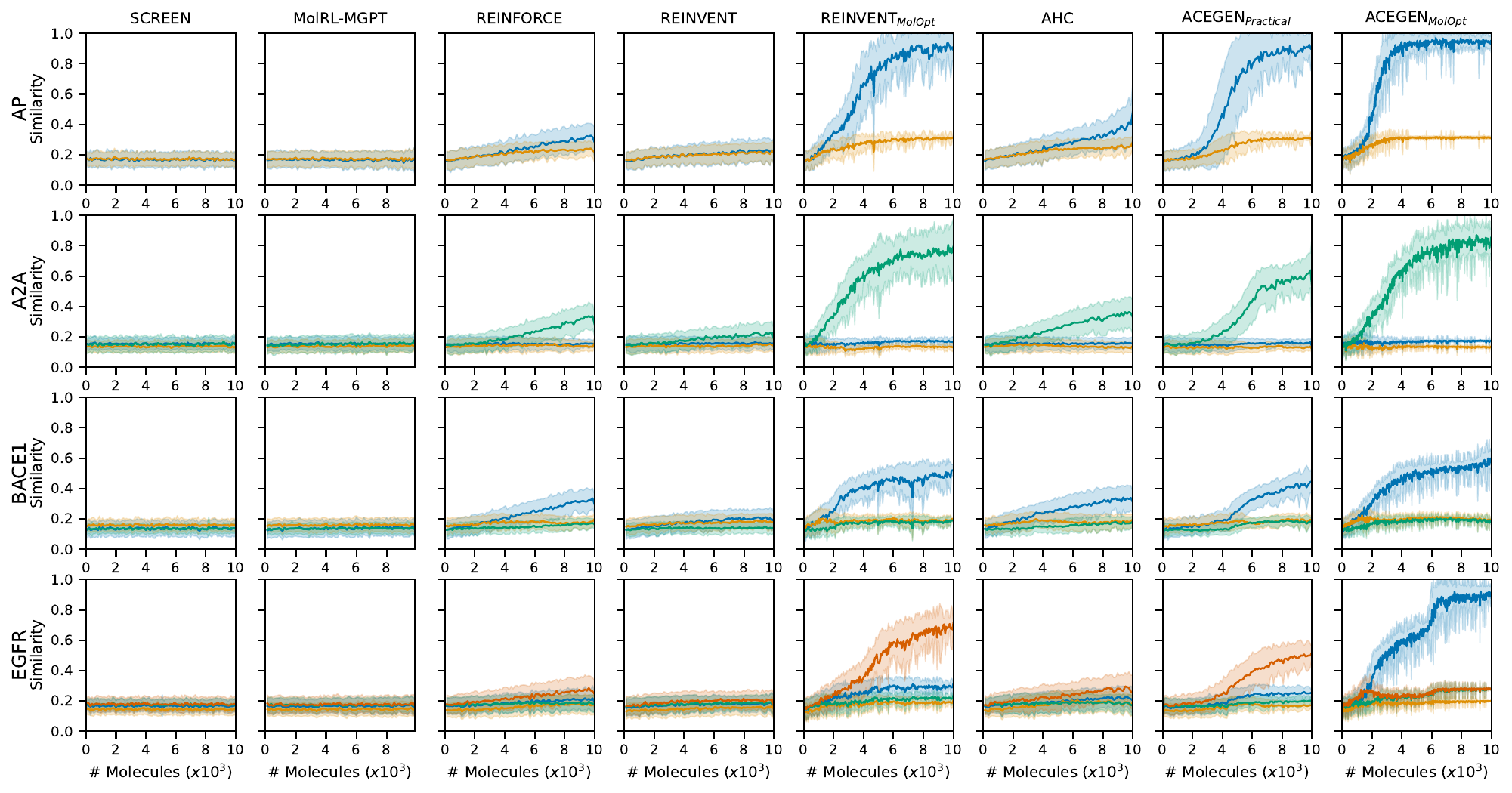}}
\caption{Baseline performance for each MolExp task during RL training, single replicate. Each line color represents similarity to a target molecule. Note that ACEGEN and REINVENT$_{MolOpt}$ methods outperform due to their enhanced ability to optimize similarity to at-least one target molecule. }
\label{fig:MolExp_baseline_sim}
\end{center}
%\vskip -0.2in
\end{figure}

\begin{figure}[ht!]
%\vskip 0.2in
\begin{center}
\centerline{\includegraphics[width=\columnwidth]{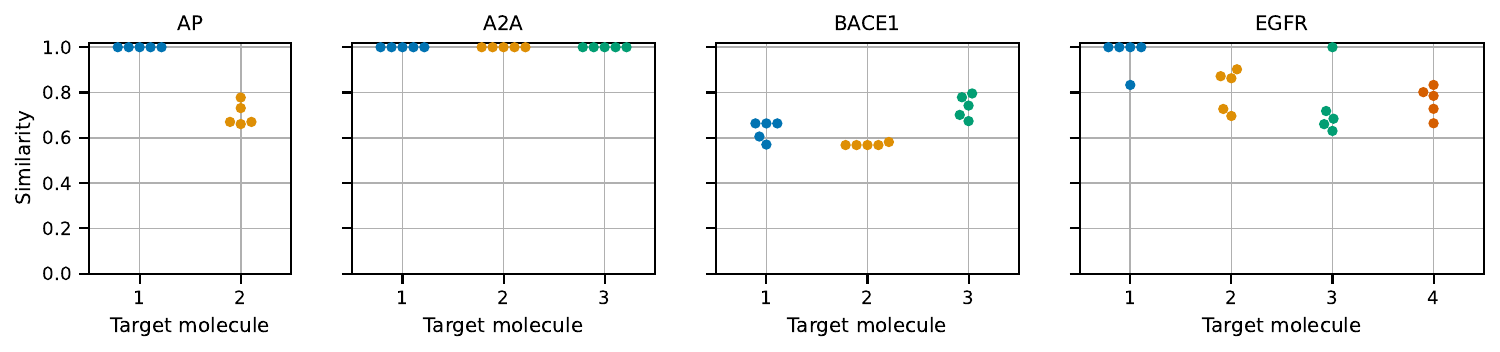}}
\caption{Estimated maximum performance achieved by ACEGEN$_{MolOpt}$. ACEGEN$_{MolOpt}$ was trained to maximize the similarity to each individual molecular target separately for a budget of 10,000 molecules. Based on this, the estimated maximum performance on the benchmark is 2.41.}
\label{fig:MolExp_estimated_max}
\end{center}
%\vskip -0.2in
\end{figure}

%%%%%%%%%%%%%%%%%%%%%%%%%%%%%%%%%%%%%%%%%%%%%%%%%%%%%%%%%%%%%%%%%%%%%%%%%%%%%%%
\clearpage
\subsection{GuacaMol}\label{app:GuacaMol_baseline}

Baseline algorithms were run on the GuacaMol benchmark also with 5 replicates, and for a standardized comparison, a budget constraint of 10,000 molecules. Previously published benchmark values have no restriction on budget, therefore, varying budgets have been used per algorithm. For this reason, values will differ from those previously published.

\begin{table}[ht!]
\centering
\caption{Baseline performance on the GuacaMol benchmark, part I.}
\label{tab:GuacaMol_baseline_performance_I}
\resizebox{\textwidth}{!}{%
\begin{tabular}{l|cccccc}
\hline
Task & SMILES LSTM & SMILES GA & Graph GA & Graph MCTS & Frag GT & SCREEN \\
\hline
Albuterol similarity & 0.92 ± 0.02 & 0.45 ± 0.05 & 0.91 ± 0.05 & 0.59 ± 0.03 & 0.99 ± 0.00 & 0.53 ± 0.01 \\
Amlodipine MPO & 0.65 ± 0.02 & 0.52 ± 0.05 & 0.65 ± 0.04 & 0.44 ± 0.01 & 0.69 ± 0.04 & 0.56 ± 0.03 \\
Aripiprazole similarity & 0.81 ± 0.03 & 0.37 ± 0.04 & 0.83 ± 0.10 & 0.25 ± 0.01 & 0.87 ± 0.06 & 0.50 ± 0.00 \\
C11H24 & 0.59 ± 0.05 & 0.01 ± 0.01 & 0.63 ± 0.16 & 0.06 ± 0.01 & 0.81 ± 0.02 & 0.05 ± 0.01 \\
C9H10N2O2PF2Cl & 0.60 ± 0.01 & 0.14 ± 0.06 & 0.67 ± 0.10 & 0.56 ± 0.02 & 0.77 ± 0.03 & 0.39 ± 0.01 \\
Celecoxib rediscovery & \textbf{1.00 ± 0.00} & 0.30 ± 0.01 & 0.70 ± 0.12 & 0.29 ± 0.03 & 0.70 ± 0.07 & 0.44 ± 0.03 \\
Deco hop & 0.66 ± 0.01 & 0.59 ± 0.01 & 0.65 ± 0.02 & 0.55 ± 0.00 & \textbf{0.73 ± 0.01} & 0.62 ± 0.01 \\
Fexofenadine MPO & 0.76 ± 0.01 & 0.67 ± 0.02 & 0.79 ± 0.01 & 0.58 ± 0.00 & 0.83 ± 0.02 & 0.72 ± 0.02 \\
Median molecules 1 & 0.34 ± 0.00 & 0.14 ± 0.02 & 0.28 ± 0.01 & 0.18 ± 0.01 & 0.28 ± 0.02 & 0.20 ± 0.01 \\
Median molecules 2 & 0.30 ± 0.01 & 0.18 ± 0.01 & 0.28 ± 0.02 & 0.13 ± 0.00 & 0.33 ± 0.06 & 0.23 ± 0.01 \\
Mestranol similarity & 0.68 ± 0.05 & 0.32 ± 0.04 & 0.64 ± 0.11 & 0.27 ± 0.01 & 0.57 ± 0.02 & 0.48 ± 0.03 \\
Osimertinib MPO & 0.83 ± 0.01 & 0.77 ± 0.02 & 0.83 ± 0.01 & 0.71 ± 0.01 & \textbf{0.87 ± 0.01} & 0.79 ± 0.00 \\
Perindopril MPO & 0.51 ± 0.01 & 0.41 ± 0.03 & 0.53 ± 0.03 & 0.28 ± 0.01 & 0.58 ± 0.02 & 0.47 ± 0.02 \\
Ranolazine MPO & 0.77 ± 0.01 & 0.55 ± 0.02 & 0.72 ± 0.02 & 0.24 ± 0.04 & \textbf{0.91 ± 0.01} & 0.71 ± 0.01 \\
Scaffold hop & 0.58 ± 0.01 & 0.46 ± 0.01 & 0.58 ± 0.02 & 0.42 ± 0.00 & 0.77 ± 0.12 & 0.51 ± 0.02 \\
Sitagliptin MPO & 0.34 ± 0.03 & 0.22 ± 0.03 & \textbf{0.49 ± 0.08} & 0.21 ± 0.02 & 0.45 ± 0.03 & 0.30 ± 0.02 \\
Thiothixene rediscovery & 0.64 ± 0.04 & 0.30 ± 0.02 & 0.59 ± 0.07 & 0.25 ± 0.01 & 0.58 ± 0.05 & 0.41 ± 0.02 \\
Troglitazone rediscovery & 0.48 ± 0.03 & 0.24 ± 0.02 & 0.44 ± 0.05 & 0.24 ± 0.01 & 0.45 ± 0.11 & 0.36 ± 0.03 \\
Valsartan smarts & \textbf{0.03 ± 0.00} & 0.02 ± 0.00 & \textbf{0.03 ± 0.00} & 0.01 ± 0.00 & \textbf{0.03 ± 0.00} & 0.02 ± 0.00 \\
Zaleplon MPO & 0.49 ± 0.01 & 0.36 ± 0.03 & 0.47 ± 0.01 & 0.32 ± 0.02 & 0.53 ± 0.02 & 0.45 ± 0.01 \\
\hline
Total Score & 11.96 ± 0.10 & 7.01 ± 0.13 & 11.70 ± 0.30 & 6.59 ± 0.07 & 12.75 ± 0.22 & 8.76 ± 0.08 \\
\hline
Total Quality & \textbf{16.03 ± 0.25} & 5.84 ± 0.56 & 9.96 ± 0.82 & 3.29 ± 0.29 & 13.40 ± 0.99 & 15.12 ± 0.19 \\
\hline
\end{tabular}%
}
\end{table}

\begin{table}[ht!]
\centering
\caption{Baseline performance on the GuacaMol benchmark, part II.}
\label{tab:GuacaMol_baseline_performance_II}
\resizebox{\textwidth}{!}{%
\begin{tabular}{l|ccccccc}
\hline
Task & MolRL-MGPT & REINFORCE & REINVENT & REINVENT$_{MolOpt}$ & AHC & ACEGEN$_{Practical}$ & ACEGEN$_{MolOpt}$ \\
\hline
Albuterol similarity & 0.60 ± 0.02 & 0.64 ± 0.03 & 0.60 ± 0.02 & 0.60 ± 0.02 & 0.71 ± 0.04 & 0.82 ± 0.05 & \textbf{1.00 ± 0.00} \\
Amlodipine MPO & 0.58 ± 0.02 & 0.61 ± 0.01 & 0.59 ± 0.01 & 0.59 ± 0.01 & 0.63 ± 0.01 & 0.67 ± 0.02 & \textbf{0.84 ± 0.06} \\
Aripiprazole similarity & 0.59 ± 0.06 & 0.58 ± 0.04 & 0.53 ± 0.02 & 0.53 ± 0.02 & 0.64 ± 0.03 & 0.74 ± 0.07 & \textbf{1.00 ± 0.00} \\
C11H24 & 0.04 ± 0.01 & 0.58 ± 0.04 & 0.06 ± 0.02 & 0.06 ± 0.02 & 0.06 ± 0.01 & 0.18 ± 0.08 & \textbf{0.88 ± 0.05} \\
C9H10N2O2PF2Cl & 0.38 ± 0.02 & 0.54 ± 0.01 & 0.48 ± 0.01 & 0.48 ± 0.01 & 0.51 ± 0.01 & 0.48 ± 0.01 & \textbf{0.81 ± 0.02} \\
Celecoxib rediscovery & 0.65 ± 0.13 & 0.56 ± 0.03 & 0.54 ± 0.07 & 0.54 ± 0.07 & 0.59 ± 0.03 & 0.87 ± 0.14 & \textbf{1.00 ± 0.00} \\
Deco hop & 0.62 ± 0.01 & 0.62 ± 0.01 & 0.61 ± 0.01 & 0.61 ± 0.01 & 0.63 ± 0.00 & 0.64 ± 0.01 & 0.72 ± 0.03 \\
Fexofenadine MPO & 0.73 ± 0.01 & 0.74 ± 0.02 & 0.74 ± 0.01 & 0.74 ± 0.01 & 0.73 ± 0.01 & 0.74 ± 0.01 & \textbf{0.87 ± 0.03} \\
Median molecules 1 & 0.20 ± 0.01 & 0.30 ± 0.02 & 0.25 ± 0.02 & 0.25 ± 0.02 & 0.28 ± 0.02 & 0.28 ± 0.04 & \textbf{0.39 ± 0.01} \\
Median molecules 2 & 0.21 ± 0.01 & 0.23 ± 0.01 & 0.23 ± 0.01 & 0.23 ± 0.01 & 0.25 ± 0.02 & 0.25 ± 0.02 & \textbf{0.36 ± 0.03} \\
Mestranol similarity & 0.54 ± 0.05 & 0.58 ± 0.01 & 0.53 ± 0.01 & 0.53 ± 0.01 & 0.57 ± 0.02 & 0.76 ± 0.05 & \textbf{1.00 ± 0.00} \\
Osimertinib MPO & 0.80 ± 0.01 & 0.81 ± 0.00 & 0.80 ± 0.00 & 0.80 ± 0.00 & 0.81 ± 0.00 & 0.80 ± 0.00 & \textbf{0.87 ± 0.01} \\
Perindopril MPO & 0.48 ± 0.01 & 0.50 ± 0.00 & 0.48 ± 0.01 & 0.48 ± 0.01 & 0.49 ± 0.01 & 0.50 ± 0.01 & \textbf{0.60 ± 0.02} \\
Ranolazine MPO & 0.75 ± 0.01 & 0.76 ± 0.01 & 0.75 ± 0.01 & 0.75 ± 0.01 & 0.76 ± 0.01 & 0.76 ± 0.00 & 0.85 ± 0.01 \\
Scaffold hop & 0.52 ± 0.01 & 0.52 ± 0.01 & 0.52 ± 0.00 & 0.52 ± 0.00 & 0.53 ± 0.01 & 0.57 ± 0.03 & \textbf{0.78 ± 0.18} \\
Sitagliptin MPO & 0.28 ± 0.02 & 0.33 ± 0.03 & 0.31 ± 0.02 & 0.31 ± 0.02 & 0.29 ± 0.02 & 0.29 ± 0.02 & 0.48 ± 0.03 \\
Thiothixene rediscovery & 0.38 ± 0.02 & 0.46 ± 0.01 & 0.42 ± 0.01 & 0.42 ± 0.01 & 0.46 ± 0.03 & 0.61 ± 0.08 & \textbf{0.98 ± 0.06} \\
Troglitazone rediscovery & 0.36 ± 0.03 & 0.39 ± 0.05 & 0.37 ± 0.03 & 0.37 ± 0.03 & 0.38 ± 0.03 & 0.44 ± 0.05 & \textbf{0.86 ± 0.20} \\
Valsartan smarts & 0.02 ± 0.00 & \textbf{0.03 ± 0.00} & 0.02 ± 0.00 & 0.02 ± 0.00 & 0.02 ± 0.00 & 0.02 ± 0.00 & \textbf{0.03 ± 0.00} \\
Zaleplon MPO & 0.45 ± 0.01 & 0.47 ± 0.01 & 0.47 ± 0.01 & 0.47 ± 0.01 & 0.47 ± 0.01 & 0.47 ± 0.01 & \textbf{0.55 ± 0.04} \\
\hline
Total Score & 9.19 ± 0.16 & 10.23 ± 0.10 & 9.29 ± 0.09 & 9.29 ± 0.09 & 9.82 ± 0.09 & 10.88 ± 0.22 & \textbf{14.85 ± 0.30} \\
\hline
Total Quality & 15.00 ± 0.17 & 15.69 ± 0.19 & 15.22 ± 0.16 & 15.22 ± 0.16 & 15.38 ± 0.20 & 15.67 ± 0.25 & 15.72 ± 0.85 \\
\hline
\end{tabular}%
}
\end{table}

%%%%%%%%%%%%%%%%%%%%%%%%%%%%%%%%%%%%%%%%%%%%%%%%%%%%%%%%%%%%%%%%%%%%%%%%%%%%%%%
\clearpage
\section{TTT scaling}

This section refers to test-time training scaling, in other words, multiple independent agents trained on the same task, which is generally described in Algorithm \ref{alg:multiple-agent}, further per-step details can be seen in \ref{alg:single-agent}.

\begin{algorithm}[H]
\begin{spacing}{1.2}
\DontPrintSemicolon
\SetAlgoLined
\label{alg:multiple-agent}
\caption{Independent Multi-Agent RL}
\KwIn{Agent index set $\mathcal{N}=\{1,\dots,n\}$; policies $\{\pi_{\theta_i}\}_{i\in\mathcal{N}}$; horizon $T$; batch size $B$; per-agent budgets $\{M_i\}$; pre-agent replay buffers $\{\mathcal{B}_i\}$, (optional) sigma value $\sigma$}
%\KwOut{Updated policies $\{\pi_{\theta_i}\}_{i\in\mathcal{N}}$}

Initialize agent policies $\pi_{\theta_i}$ \tcp*{Identical policies at initialization}
Initialize counters $N_i \leftarrow 0$ for all $i \in \mathcal{N}$ \tcp*{molecules sampled so far by agent $i$}
(Optional) Initialize frozen prior $\pi_{\text{prior}}$ \;

\BlankLine
\While{$\exists\, i \in \mathcal{N} \text{ s.t. } N_i < M_i$}{
  \ForEach{$i \in \mathcal{N}$ \textbf{with} $N_i < M_i$}{
    %\tcp{(1) Rollout (Alg.~Single-Agent step (1))}
    $\tau^{B}_{\text{on},i} \leftarrow \text{Rollout}\!\left(\pi_{\theta_i}, T, B\right)$ \tcp*{Do rollout}
    
    $N_i \leftarrow N_i + |\tau^{B_i}_{\text{on},i}|$ \tcp*{Update counter}
    
    $\tau_{i}^{B} \leftarrow \tau_{on, i}^{B} \cup \tau_{off,i}, \text{where } \tau_{off,i} \sim \mathcal{B}_{i}$ \tcp*{Sample from replay buffer}
    
    $R_i(\tau_{i}^{B}) \leftarrow f(\tau_{i}^{B})$ \tcp*{Compute rewards}
    
    $\pi_{\theta_i} \leftarrow \text{Update}_{\Psi}\!\left(\pi_{\theta_i}, \tau^{B}_{i}, R_i\!\left(\tau^{B}_{i}\right), [\pi_{\text{prior}}], [\sigma] \right)$ \tcp*{Update policy}
    
    $\mathcal{B}_i \leftarrow \mathcal{B}_i \cup \text{Top-}k\!\left(\tau^{B}_{i}, R_i\!\left(\tau^{B}_{i}\right)\right)$ \tcp*{Update replay buffer}
  }
}
%\KwRet{$\{\pi_{\theta_i}\}_{i\in\mathcal{N}}$}
\end{spacing}
\end{algorithm}

\subsection{MolExpL}\label{app:MolExpL_scaling}

\begin{table}[ht!]
\caption{Performance on MolExpL benchmark with increasing number of independent REINFORCE agents, each with a budget of 10,000.}
\label{tab:MolExpL_REINFORCE_scaling}
\resizebox{\textwidth}{!}{%
\begin{tabular}{l|cccccccc}
\hline
Task & 1 & 2 & 4 & 8 & 16 & 32 & 64 & 128 \\
\hline
AP & 0.56 ± 0.05 & 0.59 ± 0.03 & 0.67 ± 0.05 & 0.64 ± 0.02 & 0.67 ± 0.03 & 0.72 ± 0.02 & 0.72 ± 0.02 & \textbf{0.75 ± 0.03} \\
A2A & 0.41 ± 0.06 & 0.42 ± 0.04 & 0.46 ± 0.05 & 0.50 ± 0.04 & 0.55 ± 0.04 & 0.54 ± 0.04 & 0.62 ± 0.05 & \textbf{0.63 ± 0.03} \\
BACE1 & 0.22 ± 0.02 & 0.25 ± 0.02 & 0.26 ± 0.02 & 0.29 ± 0.03 & 0.29 ± 0.02 & 0.32 ± 0.04 & 0.34 ± 0.02 & \textbf{0.37 ± 0.03} \\
EGFR & 0.15 ± 0.02 & 0.17 ± 0.03 & 0.17 ± 0.02 & 0.21 ± 0.02 & 0.24 ± 0.01 & 0.24 ± 0.01 & 0.29 ± 0.02 & \textbf{0.32 ± 0.02} \\
\hline
Sum & 1.34 ± 0.08 & 1.42 ± 0.06 & 1.57 ± 0.08 & 1.64 ± 0.05 & 1.75 ± 0.06 & 1.82 ± 0.06 & 1.97 ± 0.06 & \textbf{2.08 ± 0.06} \\
\hline
\end{tabular}%
}
\end{table}

\begin{table}[ht!]
\caption{Performance on MolExpL benchmark with increasing number of independent ACEGEN$_{MolOpt}$ agents, each with a budget of 10,000.}
\label{tab:MolExpL_agent_scaling}
\resizebox{\textwidth}{!}{%
\begin{tabular}{l|cccccccc}
\hline
Task & 1 & 2 & 4 & 8 & 16 & 32 & 64 & 128 \\
\hline
AP & 0.62 ± 0.03 & 0.64 ± 0.02 & 0.70 ± 0.06 & 0.72 ± 0.03 & 0.70 ± 0.02 & 0.72 ± 0.01 & \textbf{0.75 ± 0.05} & \textbf{0.75 ± 0.02} \\
A2A & 0.41 ± 0.07 & 0.51 ± 0.08 & 0.65 ± 0.08 & 0.70 ± 0.15 & 0.76 ± 0.10 & 0.87 ± 0.13 & 0.86 ± 0.11 & \textbf{0.99 ± 0.01} \\
BACE1 & 0.31 ± 0.07 & 0.40 ± 0.03 & 0.49 ± 0.11 & 0.53 ± 0.09 & 0.64 ± 0.07 & 0.77 ± 0.03 & 0.82 ± 0.02 & \textbf{0.90 ± 0.05} \\
EGFR & 0.27 ± 0.06 & 0.34 ± 0.05 & 0.46 ± 0.11 & 0.57 ± 0.07 & 0.65 ± 0.10 & 0.77 ± 0.09 & 0.80 ± 0.05 & \textbf{0.89 ± 0.04} \\
\hline
Sum & 1.62 ± 0.12 & 1.89 ± 0.10 & 2.31 ± 0.18 & 2.51 ± 0.19 & 2.74 ± 0.15 & 3.13 ± 0.16 & 3.24 ± 0.13 & \textbf{3.52 ± 0.07} \\
\hline
\end{tabular}%
}
\end{table}

\begin{table}[ht!]
\caption{Performance on MolExpL benchmark with increasing budget for a single ACEGEN$_{MolOpt}$ agent.}
\label{tab:MolExpL_single_scaling}
\resizebox{\textwidth}{!}{%
\begin{tabular}{l|cccccccc}
\hline
Task & 10k & 20k & 40k & 80k & 160k & 320k & 640k & 1280k \\
\hline
AP & 0.62 ± 0.03 & 0.64 ± 0.02 & \textbf{0.68 ± 0.05} & 0.64 ± 0.03 & 0.65 ± 0.03 & 0.65 ± 0.06 & 0.65 ± 0.04 & 0.65 ± 0.05 \\
A2A & 0.41 ± 0.07 & 0.45 ± 0.06 & 0.45 ± 0.06 & 0.45 ± 0.06 & 0.45 ± 0.06 & 0.45 ± 0.06 & \textbf{0.46 ± 0.06} & \textbf{0.46 ± 0.06} \\
BACE1 & 0.31 ± 0.07 & 0.30 ± 0.07 & \textbf{0.35 ± 0.03} & 0.32 ± 0.02 & 0.32 ± 0.06 & 0.32 ± 0.07 & 0.33 ± 0.03 & 0.34 ± 0.06 \\
EGFR & 0.27 ± 0.06 & 0.25 ± 0.04 & \textbf{0.28 ± 0.05} & 0.27 ± 0.03 & 0.26 ± 0.03 & 0.25 ± 0.02 & 0.24 ± 0.04 & 0.25 ± 0.06 \\
\hline
Sum & 1.62 ± 0.12 & 1.64 ± 0.11 & \textbf{1.76 ± 0.10} & 1.68 ± 0.08 & 1.68 ± 0.09 & 1.67 ± 0.11 & 1.67 ± 0.09 & 1.70 ± 0.12 \\
\hline
\end{tabular}%
}
\end{table}

\begin{table}[ht!]
\caption{Performance on MolExpL benchmark with increasing budget for a single ACEGEN$_{MolOpt}$ agent with a RND exploration bonus.}
\label{tab:MolExpL_singleRND_scaling}
\resizebox{\textwidth}{!}{%
\begin{tabular}{l|cccccccc}
\hline
Task & 10k & 20k & 40k & 80k & 160k & 320k & 640k & 1280k \\
\hline
AP & 0.65 ± 0.03 & 0.66 ± 0.03 & 0.65 ± 0.04 & 0.65 ± 0.01 & 0.65 ± 0.02 & 0.66 ± 0.04 & 0.65 ± 0.01 & \textbf{0.67 ± 0.03} \\
A2A & 0.43 ± 0.03 & 0.45 ± 0.03 & \textbf{0.47 ± 0.06} & \textbf{0.47 ± 0.06} & \textbf{0.47 ± 0.06} & \textbf{0.47 ± 0.06} & \textbf{0.47 ± 0.06} & \textbf{0.47 ± 0.06} \\
BACE1 & 0.23 ± 0.01 & \textbf{0.38 ± 0.05} & 0.35 ± 0.06 & 0.32 ± 0.03 & 0.34 ± 0.03 & 0.37 ± 0.04 & \textbf{0.38 ± 0.05} & 0.37 ± 0.02 \\
EGFR & 0.18 ± 0.02 & 0.26 ± 0.04 & \textbf{0.32 ± 0.04} & \textbf{0.32 ± 0.07} & 0.31 ± 0.04 & 0.30 ± 0.03 & 0.30 ± 0.03 & 0.29 ± 0.06 \\
\hline
Sum & 1.50 ± 0.05 & 1.75 ± 0.08 & 1.79 ± 0.10 & 1.76 ± 0.09 & 1.76 ± 0.08 & 1.80 ± 0.09 & \textbf{1.81 ± 0.09} & 1.80 ± 0.09 \\
\hline
\end{tabular}%
}
\end{table}

\begin{table}[ht!]
\caption{Performance on MolExpL benchmark with increasing budget for a single ACEGEN$_{MolOpt}$ agent with DF penalization.}
\label{tab:MolExpL_singleDF_scaling}
\resizebox{\textwidth}{!}{%
\begin{tabular}{l|cccccccc}
\hline
Task & 10k & 20k & 40k & 80k & 160k & 320k & 640k & 1280k \\
\hline
AP & \textbf{0.66 ± 0.02} & \textbf{0.66 ± 0.04} & 0.65 ± 0.03 & 0.64 ± 0.02 & 0.65 ± 0.05 & 0.64 ± 0.02 & 0.65 ± 0.05 & 0.65 ± 0.02 \\
A2A & \textbf{0.50 ± 0.04} & 0.45 ± 0.04 & \textbf{0.50 ± 0.04} & \textbf{0.50 ± 0.04} & \textbf{0.50 ± 0.04} & \textbf{0.50 ± 0.04} & \textbf{0.50 ± 0.04} & 0.49 ± 0.02 \\
BACE1 & 0.29 ± 0.05 & 0.38 ± 0.08 & 0.34 ± 0.09 & 0.35 ± 0.04 & 0.33 ± 0.03 & 0.32 ± 0.08 & \textbf{0.40 ± 0.08} & 0.37 ± 0.06 \\
EGFR & 0.23 ± 0.06 & 0.27 ± 0.03 & 0.27 ± 0.06 & 0.24 ± 0.03 & 0.25 ± 0.04 & \textbf{0.32 ± 0.04} & 0.29 ± 0.07 & 0.28 ± 0.05 \\
\hline
Sum & 1.67 ± 0.09 & 1.77 ± 0.10 & 1.76 ± 0.12 & 1.72 ± 0.06 & 1.72 ± 0.08 & 1.78 ± 0.10 & \textbf{1.84 ± 0.12} & 1.79 ± 0.08 \\
\hline
\end{tabular}%
}
\end{table}

\begin{figure}[ht!]
%\vskip 0.2in
\begin{center}
\centerline{\includegraphics[width=\columnwidth]{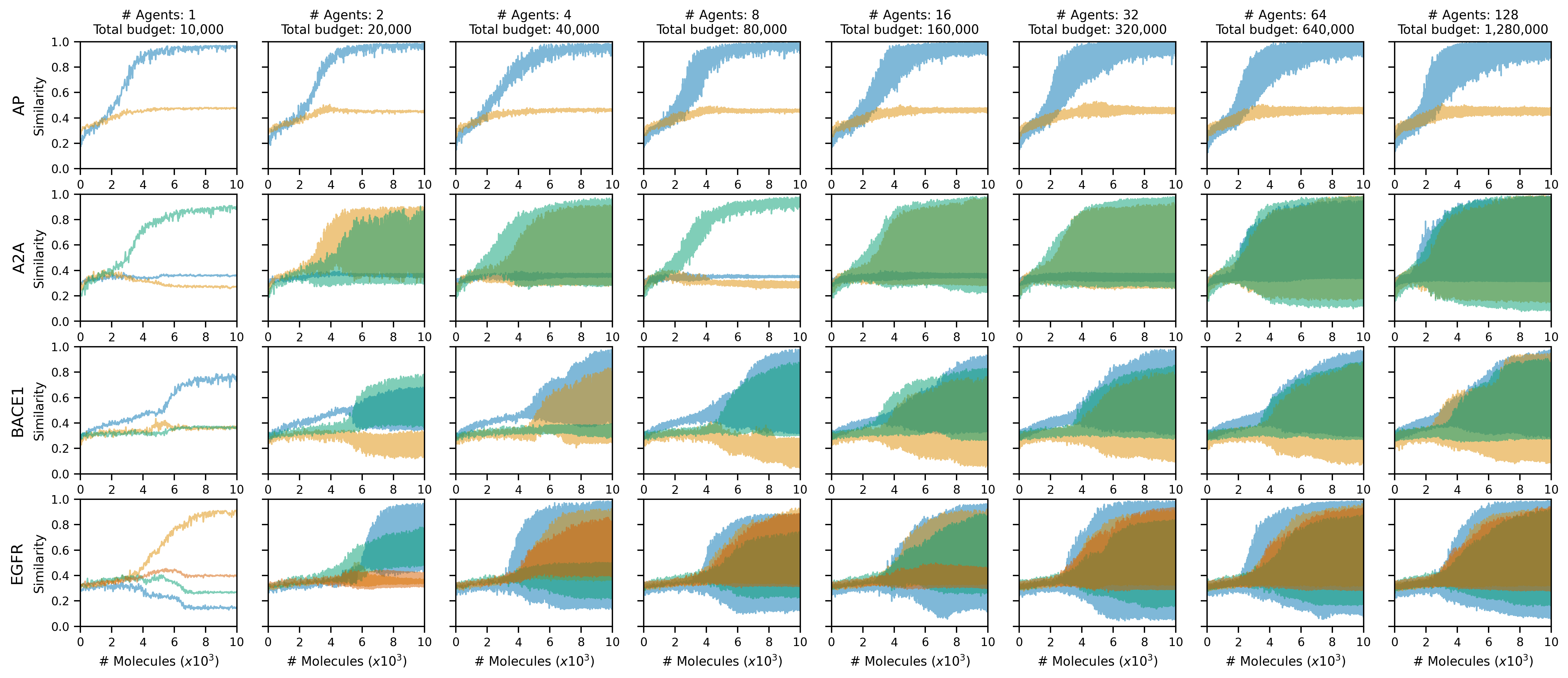}}
\caption{Multiple independent ACEGEN$_{MolOpt}$ agents on the MolExpL benchmark tasks, single replicate. Each line represents average similarity to one of the tasks target molecules, and for clarity the standard deviation is not plotted due to already high variance across agents. As the number of agents increases (increasing the total budget) stochasticity increases, resulting in divergence in agent behavior. Interestingly, even with 128 independent agents all seem to choose maximization towards one AP target molecule over the other.}
\label{fig:MolExpL_agent_scaling_sim}
\end{center}
%\vskip -0.2in
\end{figure}

\begin{figure}[ht!]
%\vskip 0.2in
\begin{center}
\centerline{\includegraphics[width=\columnwidth]{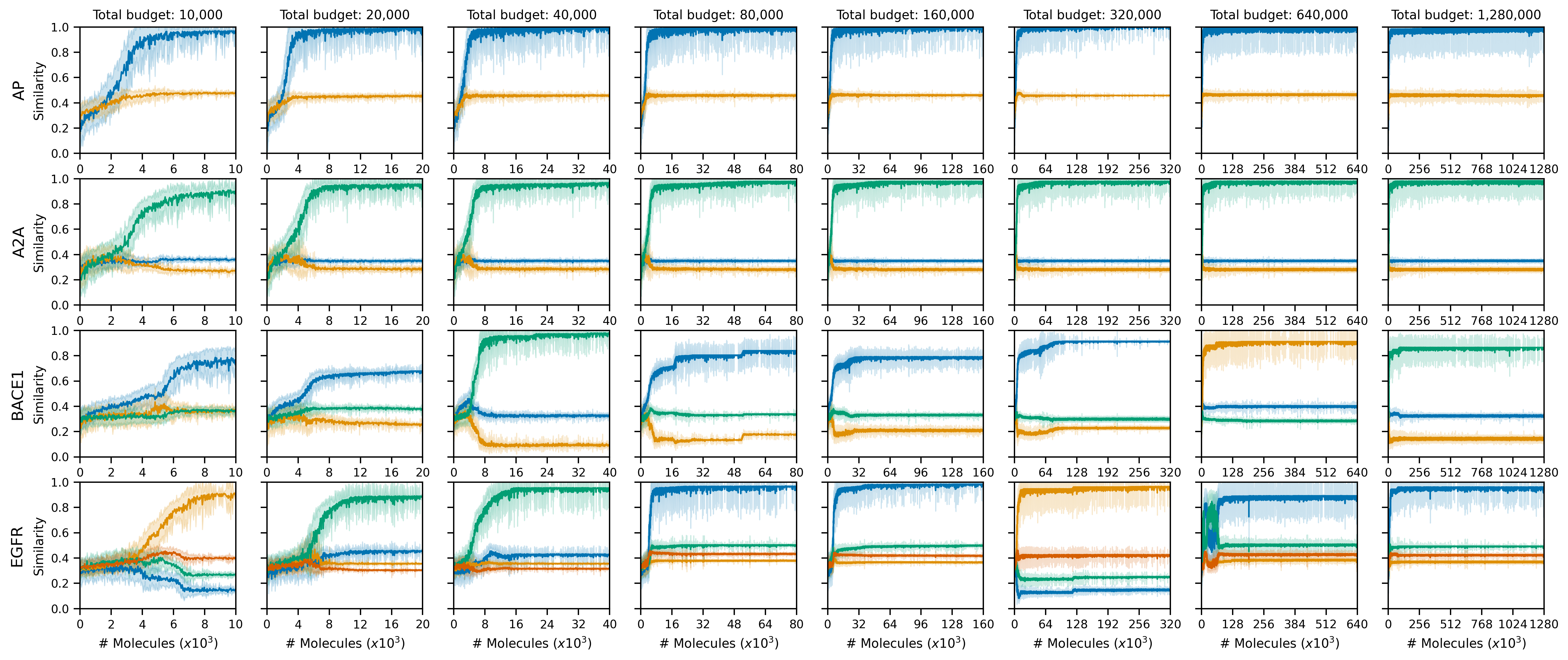}}
\caption{Single ACEGEN$_{MolOpt}$ agent on the MolExpL benchmark tasks, single replicate. Each line represents average similarity to one of the tasks target molecules. In general, a single agent optimizes for similarity to one target molecule and doesn't change throughout training. One notable exception exists for EGFR with a budget of 640,000, where the agent seems to switch molecular target (in this case from green to blue).}
\label{fig:MolExpL_single_scaling_sim}
\end{center}
%\vskip -0.2in
\end{figure}

\begin{figure}[ht!]
%\vskip 0.2in
\begin{center}
\centerline{\includegraphics[width=\columnwidth]{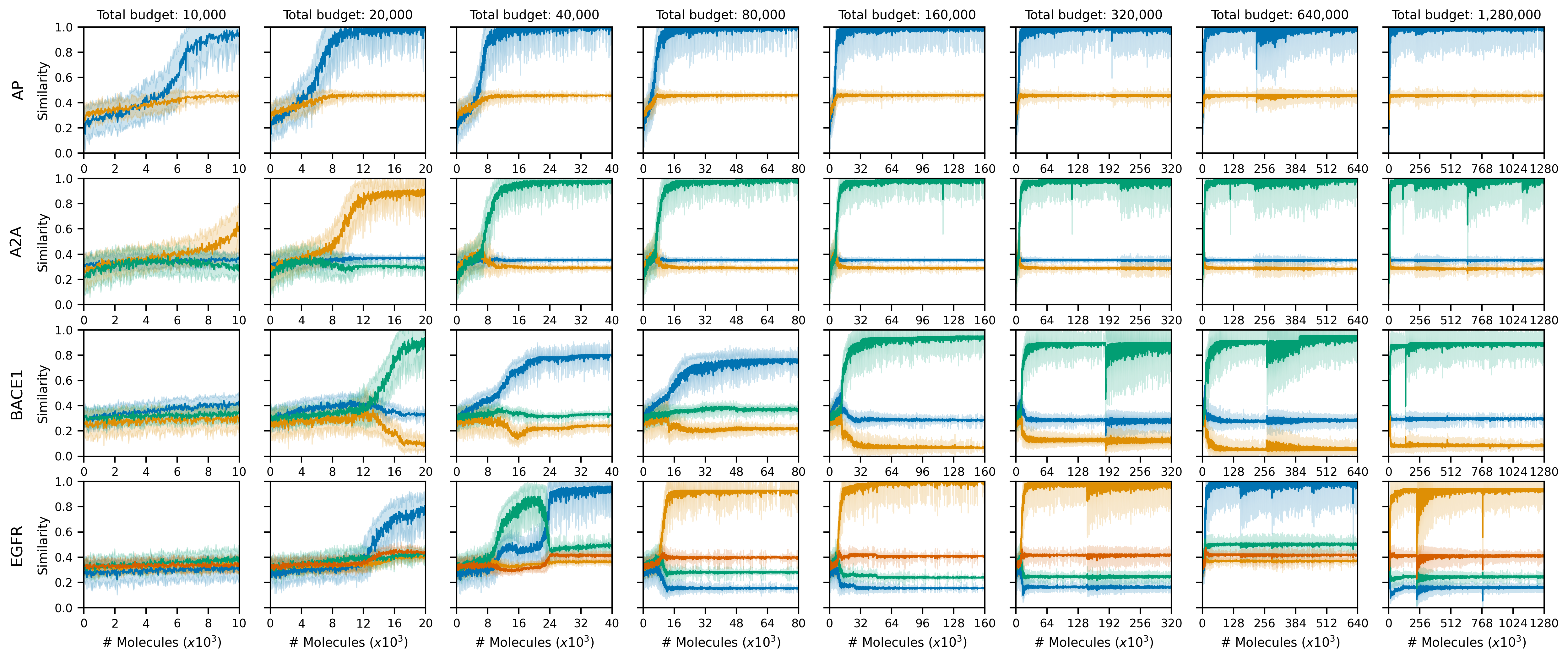}}
\caption{Single ACEGEN$_{MolOpt}$ agent with RND exploration bonus on the MolExpL benchmark tasks, single replicate. Each line represents average similarity to one of the tasks target molecules. In general, a single agent optimizes for similarity to one target molecule and doesn't change throughout training, even with exploration bonuses. Note that average score can drop from a budget of 320,000 onward, but doesn't resulted in a switching of target molecule.}
\label{fig:MolExpL_singleRND_scaling_sim}
\end{center}
%\vskip -0.2in
\end{figure}

\begin{figure}[ht!]
%\vskip 0.2in
\begin{center}
\centerline{\includegraphics[width=\columnwidth]{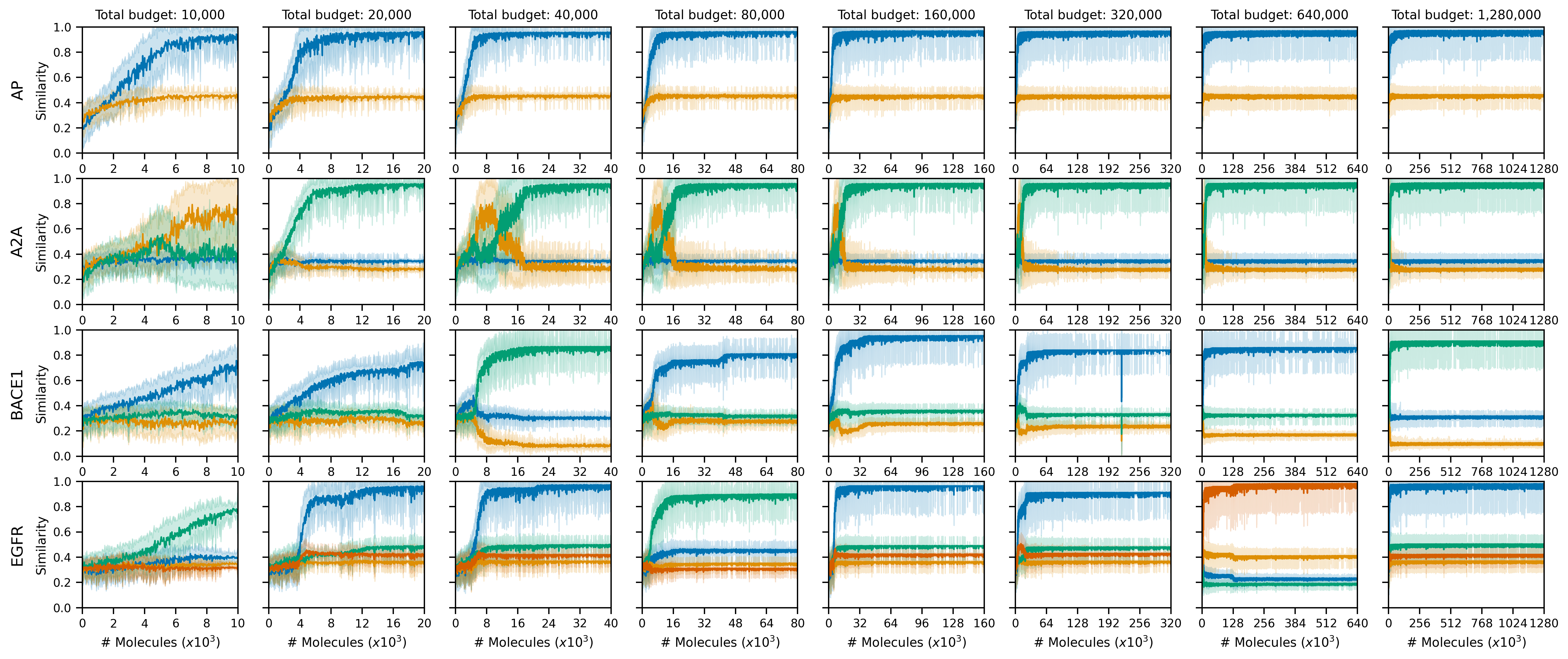}}
\caption{Single ACEGEN$_{MolOpt}$ agent with DF penalization on the MolExpL benchmark tasks, single replicate. Each line represents average similarity to one of the tasks target molecules.}
\label{fig:MolExpL_singleDF_scaling_sim}
\end{center}
%\vskip -0.2in
\end{figure}

%%%%%%%%%%%%%%%%%%%%%%%%%%%%%%%%%%%%%%%%%%%%%%%%%%%%%%%%%%%%%%%%%%%%%%%%%%%%%%%
\clearpage
\subsection{MolExp}\label{app:MolExp_scaling}

\begin{table}[ht!]
\caption{Performance on MolExp benchmark with increasing number of independent ACEGEN$_{MolOpt}$ agents, each with a budget of 10,000.}
\label{tab:MolExp_scaling_agents}
\resizebox{\textwidth}{!}{%
\begin{tabular}{l|cccccccc}
\hline
Task & 1 & 2 & 4 & 8 & 16 & 32 & 64 & 128 \\
\hline
AP & 0.42 ± 0.04 & 0.43 ± 0.03 & 0.50 ± 0.04 & 0.50 ± 0.02 & 0.48 ± 0.03 & 0.51 ± 0.02 & 0.54 ± 0.01 & \textbf{0.60 ± 0.09} \\
A2A & 0.13 ± 0.05 & 0.13 ± 0.02 & 0.15 ± 0.03 & 0.16 ± 0.02 & 0.28 ± 0.15 & 0.30 ± 0.12 & 0.44 ± 0.12 & \textbf{0.61 ± 0.23} \\
BACE1 & 0.06 ± 0.01 & 0.07 ± 0.01 & 0.09 ± 0.01 & 0.10 ± 0.02 & 0.11 ± 0.02 & 0.12 ± 0.02 & 0.15 ± 0.04 & \textbf{0.18 ± 0.04} \\
EGFR & 0.06 ± 0.01 & 0.08 ± 0.04 & 0.12 ± 0.03 & 0.18 ± 0.02 & 0.21 ± 0.06 & 0.24 ± 0.04 & 0.31 ± 0.05 & \textbf{0.36 ± 0.03} \\
\hline
Sum & 0.66 ± 0.07 & 0.71 ± 0.05 & 0.86 ± 0.06 & 0.94 ± 0.04 & 1.08 ± 0.16 & 1.17 ± 0.13 & 1.44 ± 0.14 & \textbf{1.75 ± 0.25} \\
\hline
\end{tabular}%
}
\end{table}

\begin{figure*}[ht]
     \centering
     \begin{subfigure}[t]{0.45\textwidth}
         \centering
         \includegraphics[width=\textwidth]{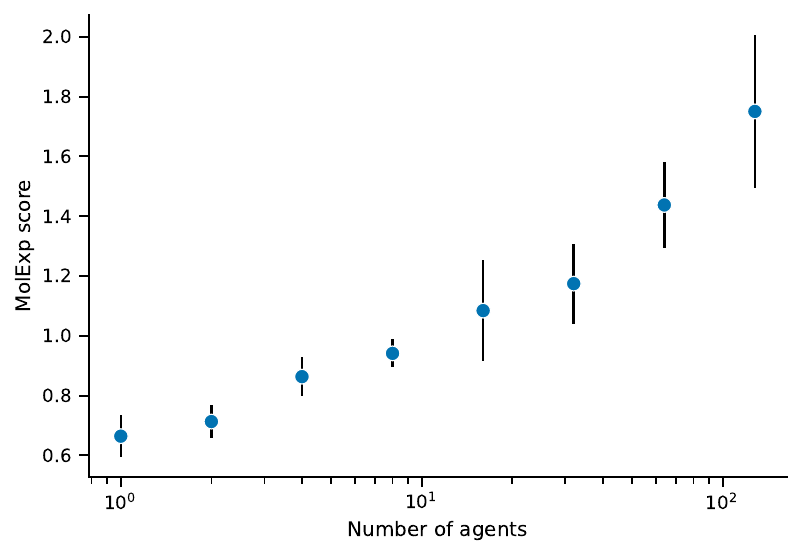}
         \caption{Scaling RL agents}
         \label{fig:MolExp_scaling_agents}
    \end{subfigure}
     \begin{subfigure}[t]{0.45\textwidth}
         \centering
         \includegraphics[width=\textwidth]{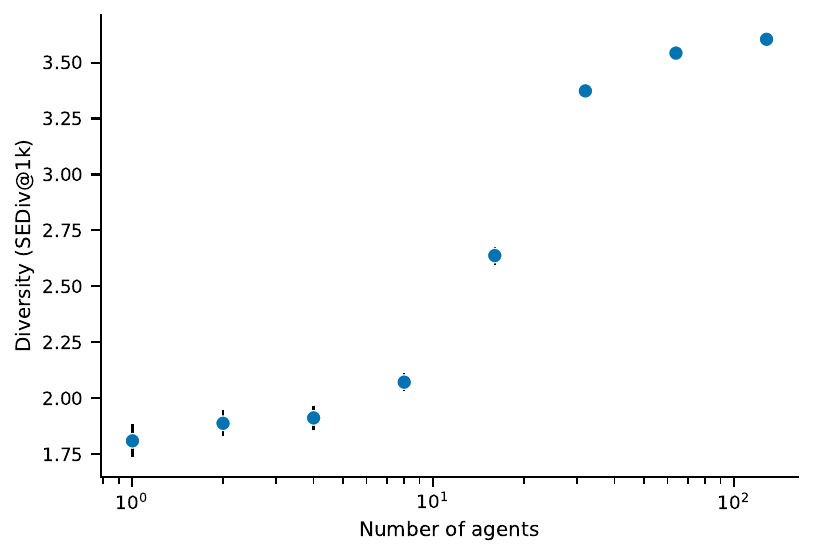}
         \caption{Molecular diversity}
         \label{fig:MolExp_scaling_diversity}
     \end{subfigure}
     \caption{Performance on the MolExp benchmark with scaling. (a) Scaling the number of independent ACEGEN$_{MolOpt}$, each with a budget of 10,000. (b) The diversity of sampled compounds as measured by sphere exclusion diversity.}
\end{figure*}

\begin{figure}[ht]
%\vskip 0.2in
\begin{center}
\centerline{\includegraphics[width=\columnwidth]{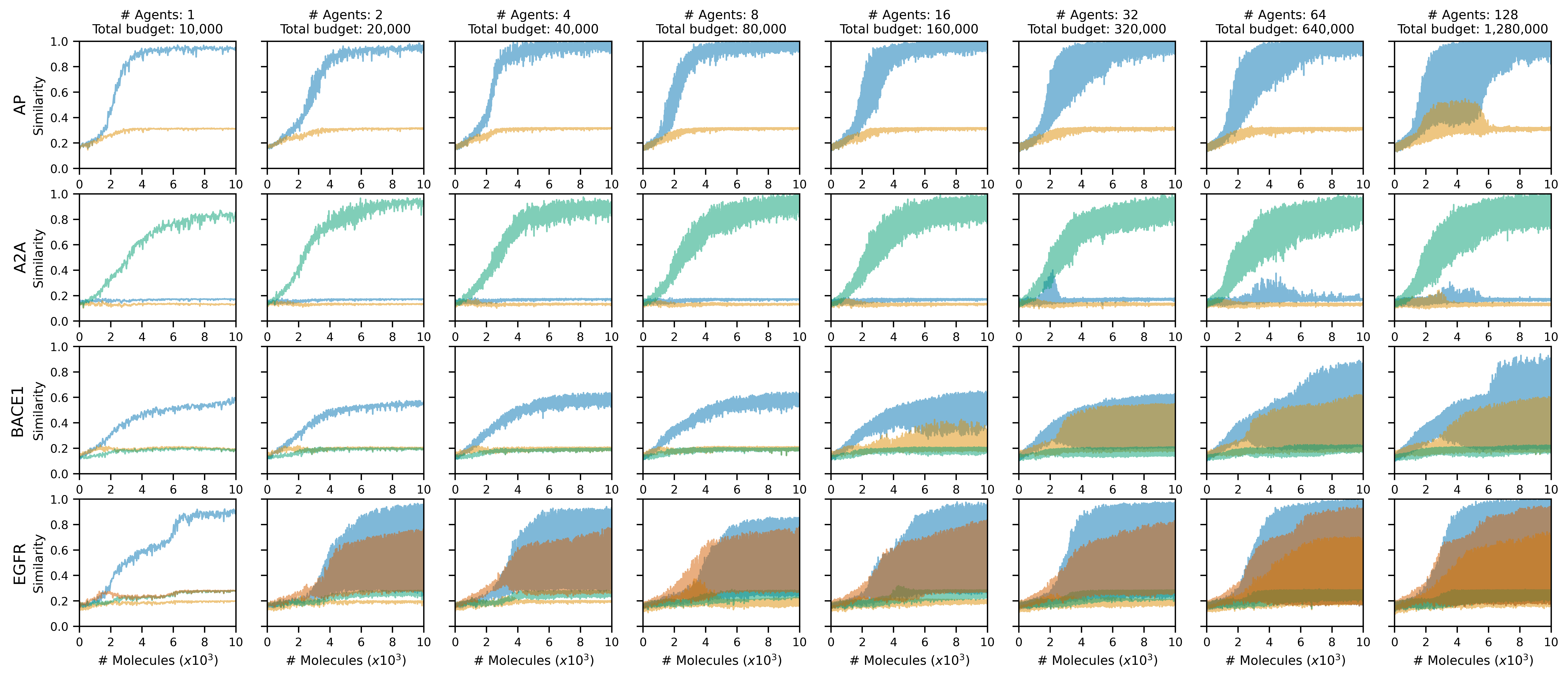}}
\caption{Multiple independent ACEGEN$_{MolOpt}$ agents on the MolExp benchmark tasks, single replicate. Each line represents average similarity to one of the tasks target molecules, and for clarity the standard deviation is not plotted due to already high variance across agents.}
\label{fig:MolExp_agent_scaling_sim}
\end{center}
%\vskip -0.2in
\end{figure}

%%%%%%%%%%%%%%%%%%%%%%%%%%%%%%%%%%%%%%%%%%%%%%%%%%%%%%%%%%%%%%%%%%%%%%%%%%%%%%%
\clearpage
\subsection{GuacaMol}\label{app:GuacaMol_scaling}

% TODO
\begin{table}[ht!]
\caption{Performance on GuacaMol benchmark with increasing number of independent ACEGEN$_{MolOpt}$ agents, each with a budget of 10,000.}
\label{tab:GuacaMol_scaling}
\resizebox{\textwidth}{!}{%
\begin{tabular}{l|cccccc}
\hline
Task & 1 & 2 & 4 & 8 & 16 & 32 \\
\hline
Albuterol similarity & \textbf{1.00 ± 0.00} & \textbf{1.00 ± 0.00} & \textbf{1.00 ± 0.00} & \textbf{1.00 ± 0.00} & \textbf{1.00 ± 0.00} & \textbf{1.00 ± 0.00} \\
Amlodipine MPO & 0.84 ± 0.06 & 0.86 ± 0.03 & 0.89 ± 0.00 & 0.89 ± 0.00 & \textbf{0.90 ± 0.00} & \textbf{0.90 ± 0.00} \\
Aripiprazole similarity & \textbf{1.00 ± 0.00} & \textbf{1.00 ± 0.00} & \textbf{1.00 ± 0.00} & \textbf{1.00 ± 0.00} & \textbf{1.00 ± 0.00} & \textbf{1.00 ± 0.00} \\
C11H24 & 0.88 ± 0.05 & 0.92 ± 0.04 & 0.97 ± 0.02 & 0.99 ± 0.00 & 0.99 ± 0.00 & \textbf{1.00 ± 0.00} \\
C9H10N2O2PF2Cl & 0.81 ± 0.02 & 0.83 ± 0.03 & 0.85 ± 0.01 & 0.88 ± 0.00 & 0.88 ± 0.00 & \textbf{0.89 ± 0.01} \\
Celecoxib rediscovery & \textbf{1.00 ± 0.00} & \textbf{1.00 ± 0.00} & \textbf{1.00 ± 0.00} & \textbf{1.00 ± 0.00} & \textbf{1.00 ± 0.00} & \textbf{1.00 ± 0.00} \\
Deco hop & 0.72 ± 0.03 & 0.73 ± 0.02 & 0.73 ± 0.02 & 0.73 ± 0.01 & \textbf{0.75 ± 0.00} & \textbf{0.75 ± 0.00} \\
Fexofenadine MPO & 0.87 ± 0.03 & 0.90 ± 0.02 & 0.89 ± 0.01 & 0.91 ± 0.01 & 0.92 ± 0.01 & \textbf{0.93 ± 0.01} \\
Median molecules 1 & 0.39 ± 0.01 & 0.41 ± 0.03 & 0.42 ± 0.01 & \textbf{0.44 ± 0.01} & \textbf{0.44 ± 0.01} & \textbf{0.44 ± 0.00} \\
Median molecules 2 & 0.36 ± 0.03 & 0.37 ± 0.02 & 0.39 ± 0.02 & 0.40 ± 0.01 & \textbf{0.41 ± 0.01} & 0.40 ± 0.00 \\
Mestranol similarity & \textbf{1.00 ± 0.00} & \textbf{1.00 ± 0.01} & \textbf{1.00 ± 0.00} & \textbf{1.00 ± 0.00} & \textbf{1.00 ± 0.00} & \textbf{1.00 ± 0.00} \\
Osimertinib MPO & 0.87 ± 0.01 & 0.88 ± 0.01 & 0.88 ± 0.01 & 0.89 ± 0.01 & 0.89 ± 0.00 & \textbf{0.90 ± 0.01} \\
Perindopril MPO & 0.60 ± 0.02 & 0.65 ± 0.03 & 0.65 ± 0.05 & 0.71 ± 0.05 & 0.72 ± 0.04 & \textbf{0.76 ± 0.01} \\
Ranolazine MPO & 0.85 ± 0.01 & 0.85 ± 0.01 & 0.85 ± 0.00 & 0.85 ± 0.01 & \textbf{0.87 ± 0.01} & \textbf{0.87 ± 0.01} \\
Scaffold hop & 0.78 ± 0.18 & 0.86 ± 0.13 & 0.79 ± 0.18 & 0.94 ± 0.06 & \textbf{0.99 ± 0.00} & \textbf{0.99 ± 0.01} \\
Sitagliptin MPO & 0.48 ± 0.03 & 0.47 ± 0.04 & 0.52 ± 0.07 & 0.56 ± 0.04 & 0.58 ± 0.03 & \textbf{0.61 ± 0.06} \\
Thiothixene rediscovery & 0.98 ± 0.06 & \textbf{1.00 ± 0.00} & \textbf{1.00 ± 0.00} & \textbf{1.00 ± 0.00} & \textbf{1.00 ± 0.00} & \textbf{1.00 ± 0.00} \\
Troglitazone rediscovery & 0.86 ± 0.20 & 0.87 ± 0.17 & \textbf{1.00 ± 0.00} & \textbf{1.00 ± 0.00} & \textbf{1.00 ± 0.00} & \textbf{1.00 ± 0.00} \\
Valsartan smarts & \textbf{0.03 ± 0.00} & \textbf{0.03 ± 0.00} & \textbf{0.03 ± 0.00} & \textbf{0.03 ± 0.00} & \textbf{0.03 ± 0.00} & \textbf{0.03 ± 0.00} \\
Zaleplon MPO & 0.55 ± 0.04 & 0.55 ± 0.00 & 0.58 ± 0.01 & 0.58 ± 0.02 & 0.61 ± 0.03 & \textbf{0.62 ± 0.03} \\
\hline
GuacaMol Score & 14.85 ± 0.30 & 15.17 ± 0.23 & 15.47 ± 0.20 & 15.79 ± 0.10 & 15.98 ± 0.06 & \textbf{16.10 ± 0.07} \\
\hline
GuacaMol Quality & \textbf{15.72 ± 0.85} & 15.05 ± 0.52 & 15.27 ± 0.53 & 14.53 ± 0.50 & 14.25 ± 0.44 & 13.99 ± 0.45 \\
\hline
\end{tabular}%
}
\end{table}

\begin{figure*}[ht]
     \centering
     \begin{subfigure}[t]{0.45\textwidth}
         \centering
         \includegraphics[width=\textwidth]{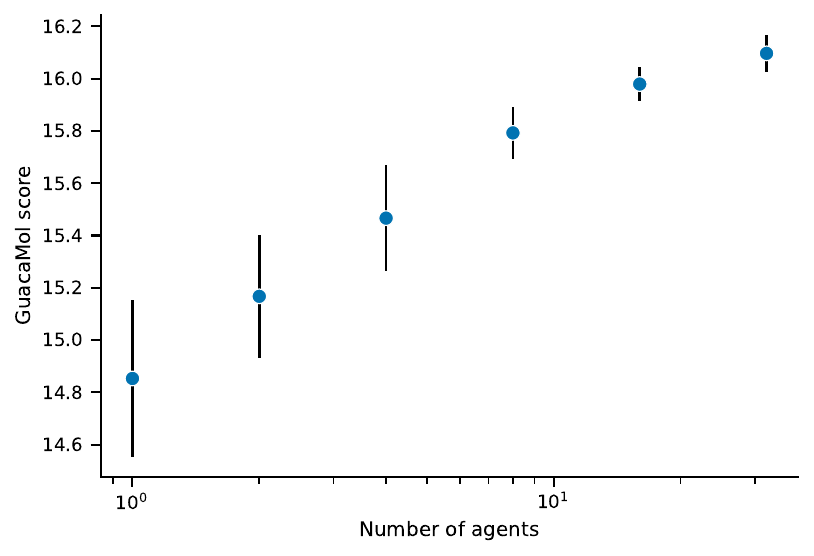}
         \caption{Scaling RL agents}
         \label{fig:GuacaMol_scaling_agents}
    \end{subfigure}
     \begin{subfigure}[t]{0.45\textwidth}
         \centering
         \includegraphics[width=\textwidth]{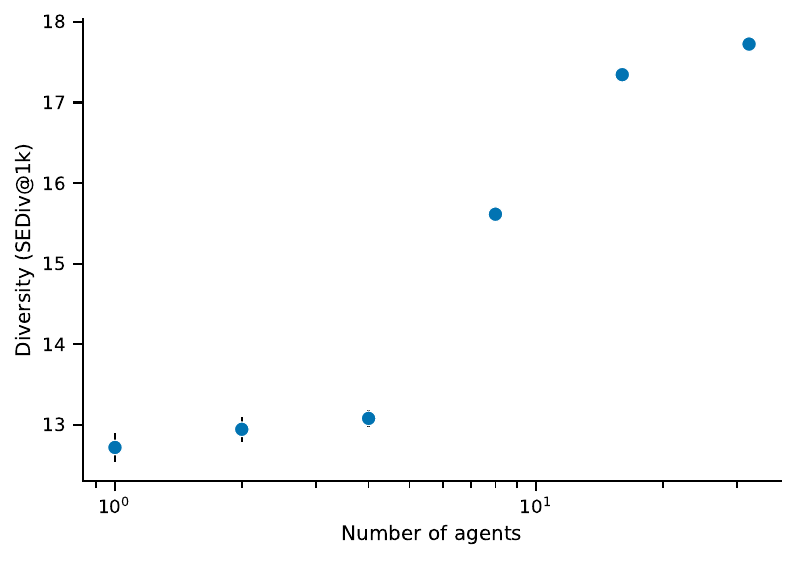}
         \caption{Molecular diversity}
         \label{fig:GuacaMol_scaling_diversity}
     \end{subfigure}
     \caption{Performance on the GuacaMol benchmark with scaling. (a) Scaling the number of independent ACEGEN$_{MolOpt}$, each with a budget of 10,000. (b) The diversity of sampled compounds as measured by sphere exclusion diversity.}
\end{figure*}

%%%%%%%%%%%%%%%%%%%%%%%%%%%%%%%%%%%%%%%%%%%%%%%%%%%%%%%%%%%%%%%%%%%%%%%%%%%%%%%
\clearpage
\section{Cooperative strategies}\label{app:CoopStrats}

We employed the following cooperative strategies using $\langle \mathcal{S}, \mathcal{A}^i, R, P^i \rangle$ for $i \in \mathcal{N}$ where the states $S$ and $R$ may be shared across agents in a cooperative manner. For brevity, we denote the probability of a full sequence generated as $\pi_{i}(\tau)$. These cooperative strategies are adaptations to the general algorithm outlined in Algorithm \ref{alg:multiple-agent}.

\begin{figure}[ht]
%\vskip 0.2in
\begin{center}
\centerline{\includegraphics[width=\columnwidth]{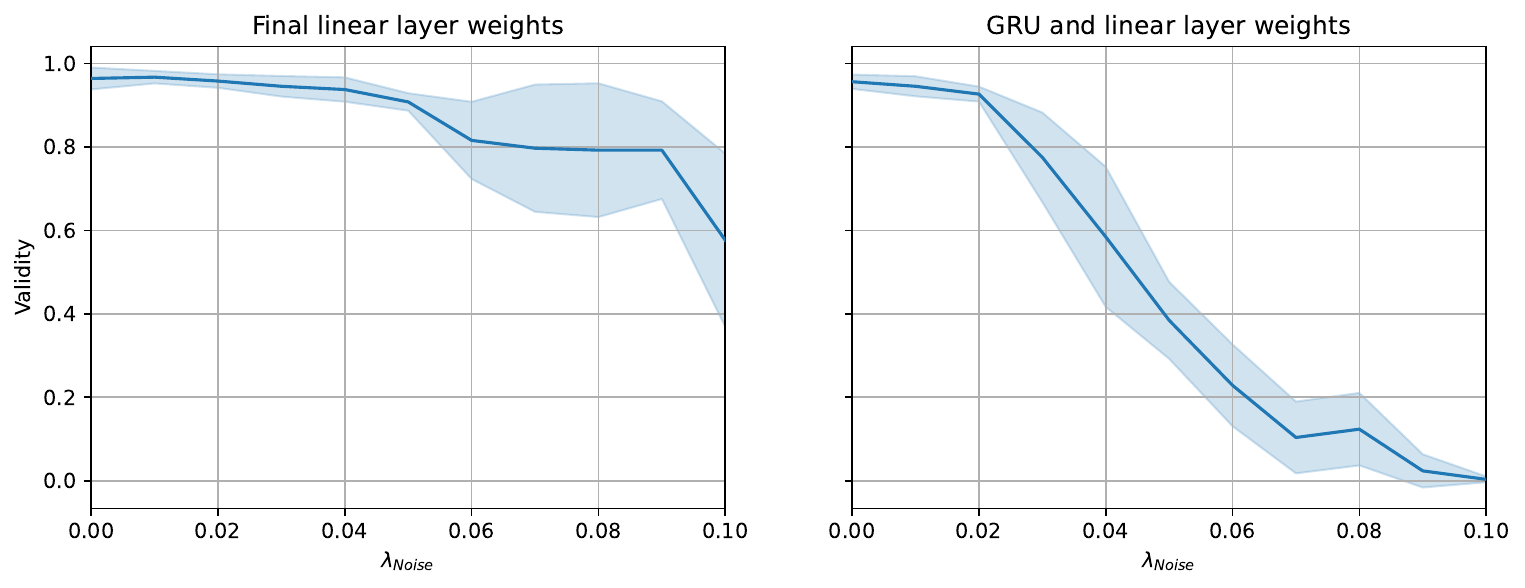}}
\caption{Application of noise with varying coefficients $\lambda_{Noise}$ to either the final linear layer weights or all GRU hidden layers and the final linear layer weights. Increasing amounts of noise decrease the validity of molecules generated.}
\label{fig:noise_validity}
\end{center}
%\vskip -0.2in
\end{figure}

\begin{description}[topsep=0.01in, itemsep=0.01in]

    \item[Purge] Purging each agent's replay buffer $\mathcal{B}_i$ to ensure no molecule exists in more than one replay buffer, resulting in a unique replay buffer for each agent.

\begin{equation}
\forall \tau \in \mathcal{B}_i, \quad \tau \notin \mathcal{B}_j \quad \text{for all } j \neq i .
\end{equation}
    
    \item[Shared] One shared replay buffer $\mathcal{B}$ is used by all agents as a method of communicating high-rewarding areas of chemical space by on-policy augmentation. We anticipate this to result in convergence of agents to the same areas of chemical space and hence act as a negative control.

    \item[Shared with bonus] The same as \textbf{Shared}, however, an additional intrinsic reward $R(\tau)_{Nov}$ was applied as a novelty bonus for molecules not already contained within the replay buffer $\mathcal{B}$. This reward shaping was applied before the ACEGEN reward shaping mechanisms outlined in the main body of text.

\begin{equation}
R(\tau)_{reshaped} = \frac{R(\tau) + R(\tau)_{Nov}}{2}
\end{equation}
    
    \item[Noise] Gaussian noise was applied to each agent's parameters once at initialization. Such as to perturb each agent's starting policy to encourage learning divergence. The effect of $\lambda$ was tested on the validity of generated SMILES strings \autoref{fig:noise_validity}. In cooperative experiments, we applied $\lambda_{Noise}=0.01$ noise to all GRU hidden layers and the final linear layer.
    
\begin{equation}
\theta'_{step=0} = \theta_{step=0} + \lambda \cdot \mathcal{N}(0, 1)
\end{equation}

    \item[RND] Random network distillation (RND) was applied to provide an exploration bonus to each agent to universally novel molecules. This bonus is calculated as the squared error between a target network $f(x)$ and predictor network $\hat{f}(x)$ mapping of $x$. The predictor network is trained iteratively on all visited states $s\in\mathcal{S}$.
    
\begin{equation}
R(\tau)' = R(\tau) + \lambda \cdot ||\hat{f}(x;\theta)-f(x)||^{2}
\end{equation}

    The networks used in our implementation employ the same architecture as the CLM agents. The predictor network is trained to minimize the log-likelihood of SMILES sequences. The bonus is computed as the difference in policies between SMILES sequences measured by the summed log-likelihoods $\sum_{t=1}^T \log P_\theta(x_t | x_{<t})$.

\item[ENT$_\mathcal{S}$] An additional loss term for the $k$-th agent was added to minimize the entropy across on-policy final states of agents $\mathcal{S}^{\{1, \dots, K\} \subseteq \mathcal{N}}$, to encourage the preferential sampling of states sampled by the $k$-th agent and discourage states sampled by other agents. To measure the likelihood of an agent sampling a state we used perplexity ($\mathcal{P}$).

\begin{equation}
\begin{split}
\mathcal{P}(\tau) &= e^{-\frac{1}{T} \sum_{t=1}^T \log \pi(a_t \mid s_t)} \\
\end{split}
\end{equation}

The perplexities given the $k$-th agent policy $\pi_k$ were summed over on-policy states of agents $\mathcal{S}^{\{1, \dots, K\} \subseteq \mathcal{N}}$ resulting in a state-conditioned policy uncertainty value for each set of agent $i$ states $\mathcal{P}_{\text{sum}}(\pi_k, \mathcal{S}^i)$. The negative of this vector (so that smaller numbers represent more uncertainty) was softmax normalized resulting in a pseudo-probability $\tilde{P}(\mathcal{S}^i|\pi_k)$ of agent $k$ sampling states derived from different agents.

\begin{equation}
\begin{split}
% Compute the Entropy
\mathcal{L}^{k}_{ENT_\mathcal{S}} &= \lambda \cdot \sum_{i=1}^{k} \tilde{P}(\mathcal{S}^i|\pi_k)\log \tilde{P}(\mathcal{S}^i|\pi_k)
\end{split}
\end{equation}

\item[CE$_{\mathcal{S}}$] A natural extension from \textbf{ENT$_\mathcal{S}$} is the addition of a cross-entropy loss term. The same steps from \textbf{ENT$_\mathcal{S}$} were followed to result in a pseudo-probability $\tilde{P}(\mathcal{S}^i|\pi_k)$ of agent $k$ sampling states derived from different agents. Then cross-entropy was applied to explicitly indicate that the $k$-th agent should assign a higher pseudo-probability to sampling states it has collected itself $\mathcal{S}^k$ than states collected by other agents $\mathcal{S}^i$. 

\begin{equation}
\begin{split}
% Compute the Entropy
\mathcal{L}^{k}_{CE_\mathcal{S}} &= -\lambda \cdot \log \tilde{P}(\mathcal{S}^k|\pi_k)
\end{split}
\end{equation}
 
\item[DIFF$_\mathcal{S}$] An additional loss term for the $k$-th agent to explicitly maximize the difference between uncertainty of states collected by previous agents relative to those collected by itself. Similar to \textbf{ENT$_\mathcal{S}$} the summed perplexity $\mathcal{P}_{\text{sum}}(\pi_k, \mathcal{S}^i)$ for collected states given the $k$-th agent policy was calculated. 
    
\begin{equation}
\mathcal{L}^k_{DIFF_{\mathcal{S}}} = - \lambda \cdot \lvert \frac{\sum_{i = 1}^{k-1} \mathcal{P}_{\text{sum}}(\pi_k, \mathcal{S}^i)}{k-1} - \mathcal{P}_{\text{sum}}(\pi_k, \mathcal{S}^k) \rvert
\end{equation}

\item[DIFF$_\mathcal{N}$] As initially proposed by \citet{hu2024novo}, an additional loss penalty was added which seeks to maximize the difference in log-probability for on-policy states $\mathcal{S}^k$ between the $k$-th agent and previous agents, scaled by the reward $R(\tau)$.

\begin{equation}
\mathcal{L}^k_{DIFF_{\mathcal{N}}} = - \lambda \cdot \sum_{i=1}^{k-1}R(\tau)\cdot \lvert \log\pi_k(\tau) - \log\pi_i(\tau) \rvert
\end{equation}

\item[DvD] Taking inspiration from \citet{parker2020effective}, we implemented Diversity via Determinant to enforce divergent behavior of agent policies. A random selection of 100 on-policy states were sampled and used to form an embedding that encoded policy behavior. This embedding was formed by concatenating the transition probabilities for possible actions $P(s^i_{t+1}|a^i_t,s_t)$ for a given agent policy $\pi_i$ for each sampled state $s$. Let this form the agent embedding $\mathbf{e_i}$ for each agent $i$ which were then used to compute pairwise similarities via a radial basis function (RBF) kernel:

\begin{equation}
K_{ij} = \exp\left(-\frac{\|\mathbf{e}_i - \mathbf{e}_j\|^2}{2\sigma^2}\right),
\end{equation}
where \(K_{ij}\) is the kernel value between agents \(i\) and \(j\), \(\|\mathbf{e}_i - \mathbf{e}_j\|^2\) is the squared Euclidean distance, and \(\sigma\) is a kernel bandwidth parameter.

The determinant of the kernel matrix \(K\), denoted as \(\det(K)\), was used as a measure of policy divergence which quantifies diversity, with higher values indicating greater divergence in agent policies and hence, was used as a penalty term:
\begin{equation}
\mathcal{L}^i_\text{DvD} = - \det(K),
\end{equation}

\item[POPNORM] Population normalization was used as a strategy to encourage divergent behavior from previous agents by adding a penalty term of the average population return up to the current $k$-th agent. 

\begin{equation}
\mathcal{L}^k_\text{POPNORM} = - \frac{1}{k - 1}\sum^{k-1}_{i = 1} \pi_{i}(\tau) \cdot R(\tau)
\end{equation}
    
\end{description}

%%%%%%%%%%%%%%%%%%%%%%%%%%%%%%%%%%%%%%%%%%%%%%%%%%%%%%%%%%%%%%%%%%%%%%%%%%%%%%%
\clearpage
\section{Cooperative RL}
\subsection{MolExpL}\label{app:MolExpL_coop}

\begin{table}[!ht]
\caption{Performance on MolExpL benchmark of different 4-agent cooperative strategies, part I.}
\label{tab:MolExpL_coop_ptI}
\resizebox{\textwidth}{!}{%
\begin{tabular}{l|ccccccccc}
\hline
Task & Independent & Noise & Shared & Shared
w. bonus & Purge & RND & DvD & POPNORM & MolRL-MGPT \\
\hline
AP & \textbf{0.70 ± 0.06} & 0.66 ± 0.02 & 0.68 ± 0.04 & 0.64 ± 0.01 & 0.65 ± 0.02 & 0.68 ± 0.02 & 0.69 ± 0.04 & 0.67 ± 0.03 & 0.54 ± 0.02 \\
A2A & 0.65 ± 0.08 & 0.64 ± 0.18 & 0.51 ± 0.12 & 0.55 ± 0.11 & 0.63 ± 0.05 & 0.66 ± 0.07 & 0.68 ± 0.10 & 0.68 ± 0.08 & 0.52 ± 0.05 \\
BACE1 & 0.49 ± 0.11 & 0.53 ± 0.10 & 0.39 ± 0.03 & 0.37 ± 0.03 & 0.42 ± 0.13 & 0.35 ± 0.06 & 0.40 ± 0.05 & 0.53 ± 0.08 & 0.31 ± 0.03 \\
EGFR & 0.46 ± 0.11 & 0.44 ± 0.06 & 0.29 ± 0.06 & 0.29 ± 0.03 & 0.45 ± 0.06 & 0.41 ± 0.06 & 0.43 ± 0.05 & 0.48 ± 0.05 & 0.23 ± 0.04 \\
\hline
Sum & 2.31 ± 0.18 & 2.27 ± 0.21 & 1.86 ± 0.14 & 1.86 ± 0.12 & 2.16 ± 0.15 & 2.10 ± 0.11 & 2.19 ± 0.13 & 2.35 ± 0.13 & 1.59 ± 0.07 \\
\hline
\end{tabular}%
}
\end{table}

\begin{table}[!ht]
\caption{Performance on MolExpL benchmark of different 4-agent cooperative strategies, part II.}
\label{tab:MolExpL_coop_ptII}
\resizebox{\textwidth}{!}{%
\begin{tabular}{l|cccccccc}
\hline
Task & ENT$_S$-0.001 & ENT$_S$-0.01 & ENT$_S$-0.1 & ENT$_S$-1 & CE$_S$-0.001 & CE$_S$-0.01 & CE$_S$-0.1 & CE$_S$-1 \\
\hline
AP & 0.66 ± 0.03 & 0.67 ± 0.03 & 0.65 ± 0.02 & 0.68 ± 0.02 & 0.69 ± 0.03 & 0.68 ± 0.01 & 0.68 ± 0.04 & 0.68 ± 0.04 \\
A2A & 0.67 ± 0.20 & 0.72 ± 0.12 & 0.65 ± 0.12 & 0.60 ± 0.08 & 0.65 ± 0.18 & 0.62 ± 0.07 & 0.73 ± 0.13 & 0.70 ± 0.09 \\
BACE1 & \textbf{0.59 ± 0.08} & 0.51 ± 0.17 & 0.50 ± 0.09 & 0.41 ± 0.05 & 0.46 ± 0.15 & 0.53 ± 0.08 & 0.43 ± 0.05 & 0.36 ± 0.07 \\
EGFR & 0.47 ± 0.05 & 0.44 ± 0.10 & 0.47 ± 0.09 & 0.38 ± 0.07 & 0.46 ± 0.15 & 0.44 ± 0.10 & 0.47 ± 0.08 & 0.39 ± 0.11 \\
\hline
Sum & \textbf{2.39 ± 0.22} & 2.33 ± 0.23 & 2.28 ± 0.18 & 2.07 ± 0.12 & 2.25 ± 0.28 & 2.27 ± 0.14 & 2.31 ± 0.16 & 2.13 ± 0.16 \\
\hline
\end{tabular}%
}
\end{table}

\begin{table}[!ht]
\caption{Performance on MolExpL benchmark of different 4-agent cooperative strategies, part III.}
\label{tab:MolExpL_coop_ptIII}
\resizebox{\textwidth}{!}{%
\begin{tabular}{l|cccccccc}
\hline
Task & DIFF$_S$-0.001 & DIFF$_S$-0.01 & DIFF$_S$-0.1 & DIFF$_S$-1 & DIFF$_N$-0.001 & DIFF$_N$-0.01 & DIFF$_N$-0.1 & DIFF$_N$-1 \\
\hline
AP & 0.68 ± 0.02 & 0.69 ± 0.02 & 0.66 ± 0.03 & 0.63 ± 0.03 & 0.67 ± 0.03 & 0.68 ± 0.03 & 0.66 ± 0.02 & 0.68 ± 0.04 \\
A2A & 0.57 ± 0.07 & 0.61 ± 0.08 & 0.54 ± 0.06 & 0.46 ± 0.04 & 0.68 ± 0.14 & 0.64 ± 0.04 & 0.63 ± 0.12 & \textbf{0.74 ± 0.07} \\
BACE1 & 0.45 ± 0.06 & 0.46 ± 0.06 & 0.35 ± 0.04 & 0.31 ± 0.06 & 0.48 ± 0.11 & 0.43 ± 0.13 & 0.47 ± 0.07 & 0.47 ± 0.13 \\
EGFR & 0.45 ± 0.10 & 0.44 ± 0.05 & 0.36 ± 0.10 & 0.26 ± 0.05 & 0.50 ± 0.04 & \textbf{0.53 ± 0.03} & 0.45 ± 0.05 & 0.50 ± 0.10 \\
\hline
Sum & 2.14 ± 0.13 & 2.20 ± 0.12 & 1.92 ± 0.13 & 1.66 ± 0.10 & 2.33 ± 0.19 & 2.27 ± 0.14 & 2.21 ± 0.15 & \textbf{2.39 ± 0.18} \\
\hline
\end{tabular}%
}
\end{table}

\begin{figure}[ht]
%\vskip 0.2in
\begin{center}
\centerline{\includegraphics[width=\columnwidth]{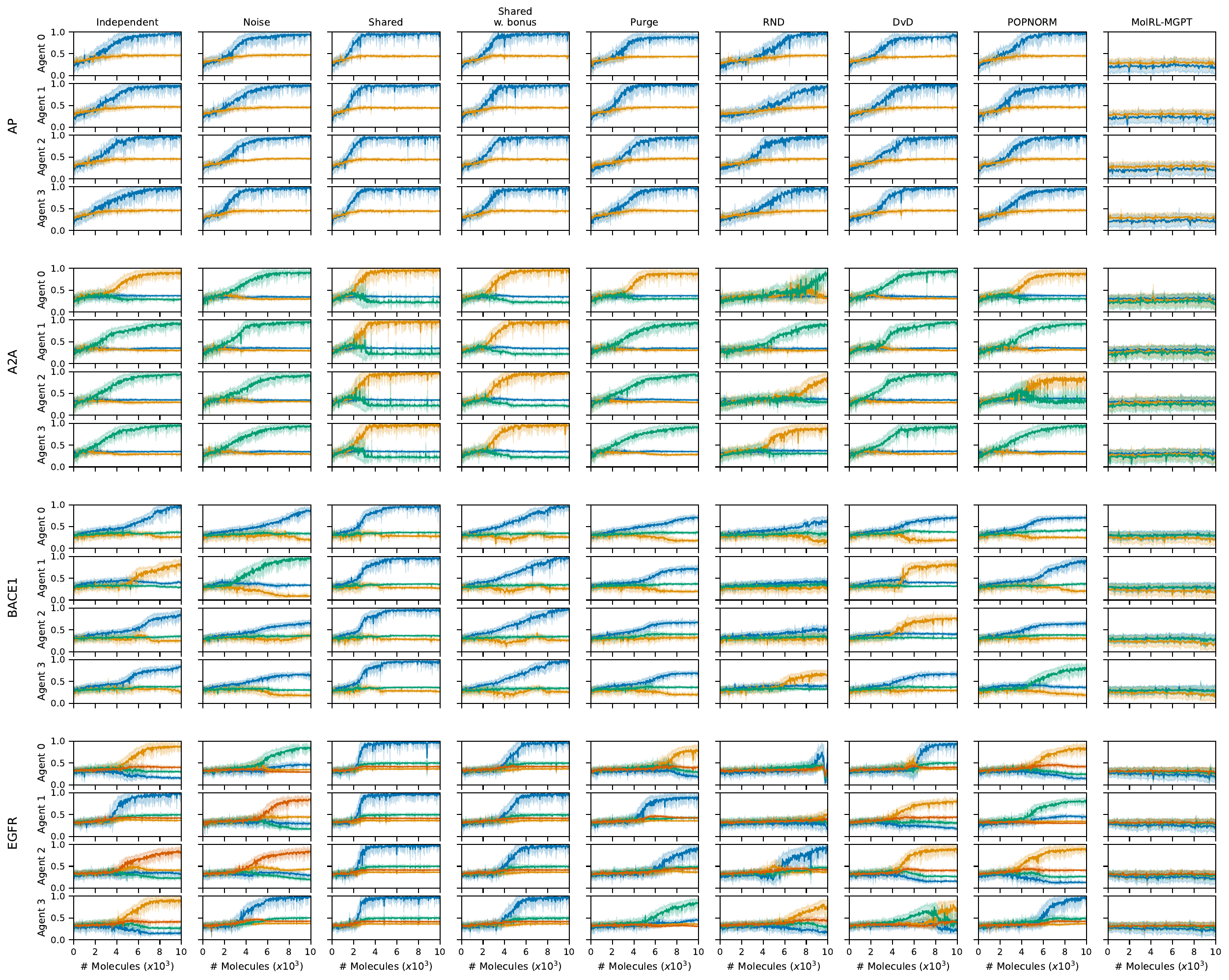}}
\caption{Cooperative strategies using 4-agents on the MolExpL benchmark tasks, single replicate, part I. Each line represents average similarity to one of the tasks target molecules. Training of each agent is plotted. Perfect cooperation is each agent learning to rediscover a different target molecule. Most cooperative strategies result in no additional divergent behavior or in some cases, slower learning.}
\label{fig:coop_curves_molexp_I}
\end{center}
%\vskip -0.2in
\end{figure}

\begin{figure}[ht]
%\vskip 0.2in
\begin{center}
\centerline{\includegraphics[width=\columnwidth]{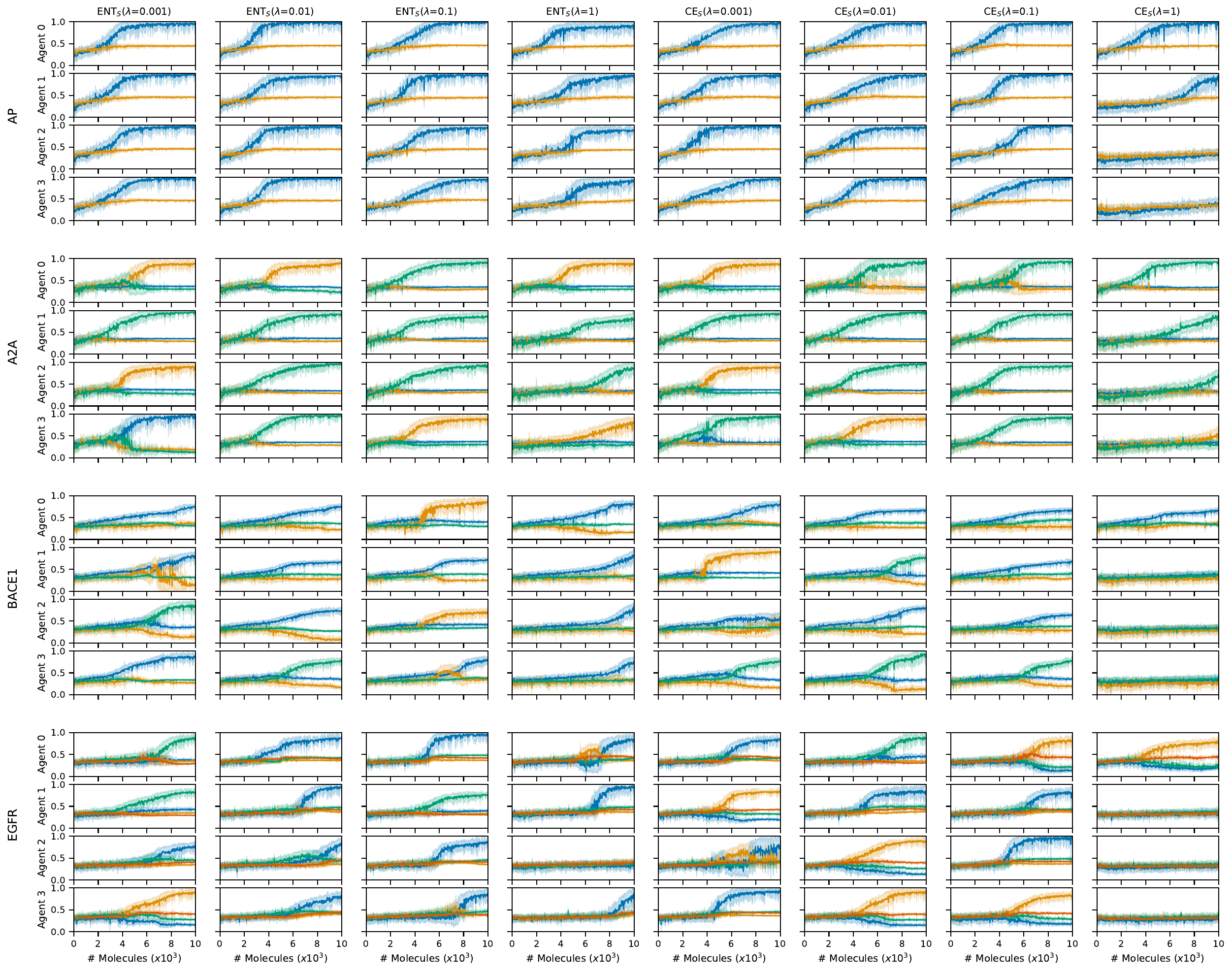}}
\caption{Cooperative strategies using 4-agents on the MolExpL benchmark tasks, single replicate, part II. Each line represents average similarity to one of the tasks target molecules. Training of each agent is plotted. Perfect cooperation is each agent learning to rediscover a different target molecule. Most cooperative strategies result in no additional divergent behavior or in some cases, slower learning.}
\label{fig:coop_curves_molexp_II}
\end{center}
%\vskip -0.2in
\end{figure}

\begin{figure}[ht]
%\vskip 0.2in
\begin{center}
\centerline{\includegraphics[width=\columnwidth]{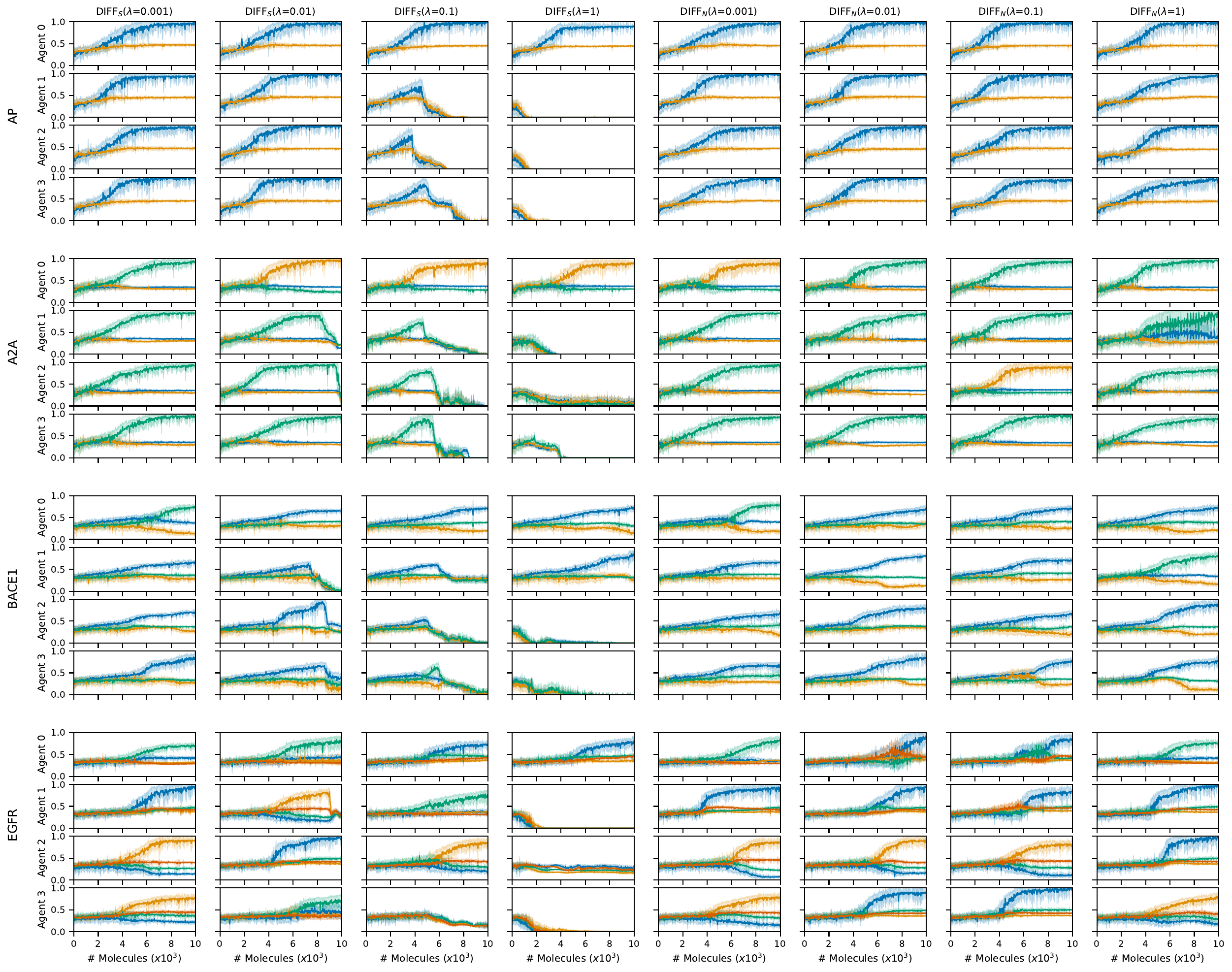}}
\caption{Cooperative strategies using 4-agents on the MolExpL benchmark tasks, single replicate, part III. Each line represents average similarity to one of the tasks target molecules. Training of each agent is plotted. Perfect cooperation is each agent learning to rediscover a different target molecule. Most cooperative strategies result in no additional divergent behavior or in some cases, slower learning.}
\label{fig:coop_curves_molexp_III}
\end{center}
%\vskip -0.2in
\end{figure}

%%%%%%%%%%%%%%%%%%%%%%%%%%%%%%%%%%%%%%%%%%%%%%%%%%%%%%%%%%%%%%%%%%%%%%%%%%%%%%%
\clearpage
\subsection{MolExp}\label{app:MolExp_coop}

\begin{table}[!ht]
\caption{Performance on MolExp benchmark of different 4-agent cooperative strategies, part I.}
\label{tab:MolExp_coop_ptI}
\resizebox{\textwidth}{!}{%
\begin{tabular}{l|ccccccccc}
\hline
Task & Independent & Noise & Shared & Shared
w. bonus & Purge & RND & DvD & POPNORM & MolRL-MGPT \\
\hline
AP & 0.50 ± 0.04 & 0.47 ± 0.04 & 0.43 ± 0.02 & 0.44 ± 0.02 & 0.44 ± 0.01 & 0.47 ± 0.04 & 0.48 ± 0.06 & 0.46 ± 0.03 & 0.31 ± 0.06 \\
A2A & 0.15 ± 0.03 & 0.25 ± 0.13 & 0.15 ± 0.04 & 0.17 ± 0.03 & 0.17 ± 0.05 & 0.17 ± 0.04 & 0.20 ± 0.12 & 0.17 ± 0.06 & 0.12 ± 0.02 \\
BACE1 & 0.09 ± 0.01 & 0.09 ± 0.01 & 0.07 ± 0.01 & 0.07 ± 0.01 & 0.08 ± 0.01 & 0.08 ± 0.01 & 0.07 ± 0.00 & 0.08 ± 0.01 & \textbf{0.15 ± 0.04} \\
EGFR & 0.12 ± 0.03 & 0.11 ± 0.04 & 0.07 ± 0.02 & 0.06 ± 0.01 & 0.12 ± 0.04 & 0.08 ± 0.01 & 0.11 ± 0.02 & 0.13 ± 0.10 & 0.05 ± 0.01 \\
\hline
Sum & \textbf{1.30 ± 0.07} & 0.91 ± 0.14 & 0.73 ± 0.05 & 0.74 ± 0.04 & 0.81 ± 0.07 & 0.80 ± 0.06 & 0.86 ± 0.13 & 0.85 ± 0.12 & 0.62 ± 0.07 \\
\hline
\end{tabular}%
}
\end{table}

\begin{table}[!ht]
\caption{Performance on MolExp benchmark of different 4-agent cooperative strategies, part II.}
\label{tab:MolExp_coop_ptII}
\resizebox{\textwidth}{!}{%
\begin{tabular}{l|cccccccc}
\hline
Task & ENT$_S$-0.001 & ENT$_S$-0.01 & ENT$_S$-0.1 & ENT$_S$-1 & CE$_S$-0.001 & CE$_S$-0.01 & CE$_S$-0.1 & CE$_S$-1 \\
\hline
AP & 0.45 ± 0.04 & 0.45 ± 0.03 & \textbf{0.54 ± 0.07} & 0.47 ± 0.04 & 0.51 ± 0.04 & 0.49 ± 0.06 & \textbf{0.54 ± 0.09} & 0.48 ± 0.05 \\
A2A & 0.15 ± 0.02 & 0.25 ± 0.12 & 0.23 ± 0.16 & 0.25 ± 0.13 & 0.22 ± 0.11 & 0.15 ± 0.03 & \textbf{0.42 ± 0.07} & 0.27 ± 0.12 \\
BACE1 & 0.11 ± 0.06 & 0.08 ± 0.00 & 0.09 ± 0.02 & 0.09 ± 0.01 & 0.08 ± 0.02 & 0.08 ± 0.01 & 0.11 ± 0.03 & 0.09 ± 0.01 \\
EGFR & 0.13 ± 0.05 & 0.13 ± 0.04 & 0.12 ± 0.05 & 0.10 ± 0.03 & 0.10 ± 0.03 & 0.12 ± 0.04 & 0.14 ± 0.04 & 0.08 ± 0.04 \\
\hline
Sum & 0.83 ± 0.08 & 0.91 ± 0.12 & 0.98 ± 0.18 & 0.91 ± 0.14 & 0.91 ± 0.13 & 0.84 ± 0.08 & 1.21 ± 0.12 & 0.92 ± 0.14 \\
\hline
\end{tabular}%
}
\end{table}

\begin{table}[!ht]
\caption{Performance on MolExp benchmark of different 4-agent cooperative strategies, part III.}
\label{tab:MolExp_coop_ptIII}
\resizebox{\textwidth}{!}{%
\begin{tabular}{l|cccccccc}
\hline
Task & DIFF$_S$-0.001 & DIFF$_S$-0.01 & DIFF$_S$-0.1 & DIFF$_S$-1 & DIFF$_N$-0.001 & DIFF$_N$-0.01 & DIFF$_N$-0.1 & DIFF$_N$-1 \\
\hline
AP & 0.47 ± 0.04 & 0.45 ± 0.05 & 0.45 ± 0.03 & 0.42 ± 0.03 & 0.46 ± 0.03 & 0.47 ± 0.02 & 0.45 ± 0.01 & 0.53 ± 0.12 \\
A2A & 0.16 ± 0.04 & 0.17 ± 0.04 & 0.14 ± 0.01 & 0.14 ± 0.02 & 0.16 ± 0.03 & 0.17 ± 0.04 & 0.19 ± 0.10 & 0.41 ± 0.19 \\
BACE1 & 0.09 ± 0.01 & 0.11 ± 0.03 & 0.07 ± 0.02 & 0.06 ± 0.01 & 0.07 ± 0.01 & 0.09 ± 0.02 & 0.11 ± 0.03 & 0.10 ± 0.03 \\
EGFR & 0.12 ± 0.02 & 0.12 ± 0.01 & 0.09 ± 0.03 & 0.06 ± 0.02 & 0.10 ± 0.05 & \textbf{0.16 ± 0.03} & 0.11 ± 0.04 & 0.12 ± 0.02 \\
\hline
Sum & 0.84 ± 0.06 & 0.84 ± 0.07 & 0.74 ± 0.05 & 0.69 ± 0.04 & 0.79 ± 0.06 & 0.89 ± 0.06 & 0.86 ± 0.12 & 1.15 ± 0.23 \\
\hline
\end{tabular}%
}
\end{table}

\begin{figure*}[ht]
     \centering
     \begin{subfigure}[l]{\textwidth}
         \centering
         \includegraphics[width=\textwidth]{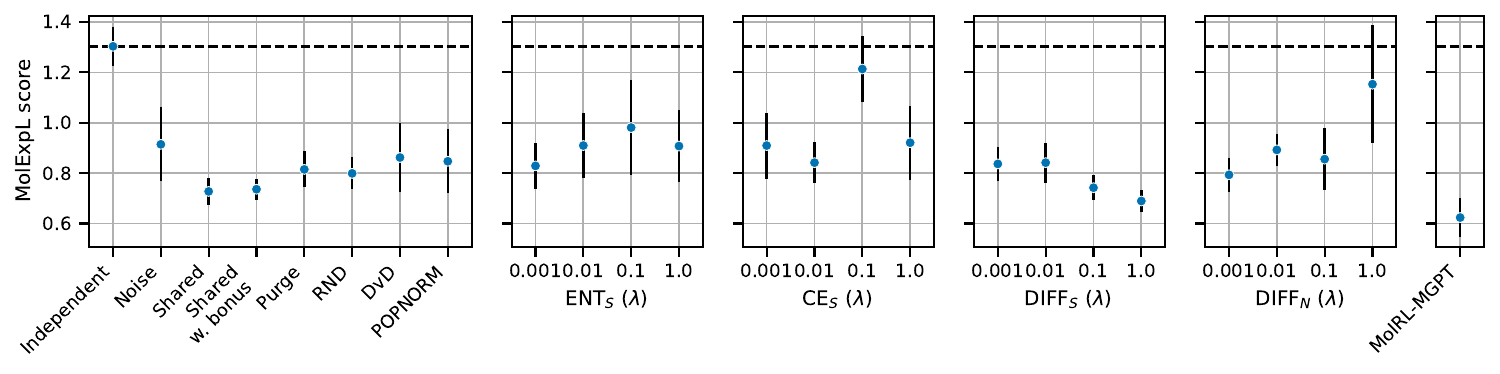}
         \caption{Cooperative RL MolExp benchmark performance}
         \label{fig:MolExp_coop_score}
     \end{subfigure}
     \begin{subfigure}[l]{\textwidth}
         \centering
         \includegraphics[width=\textwidth]{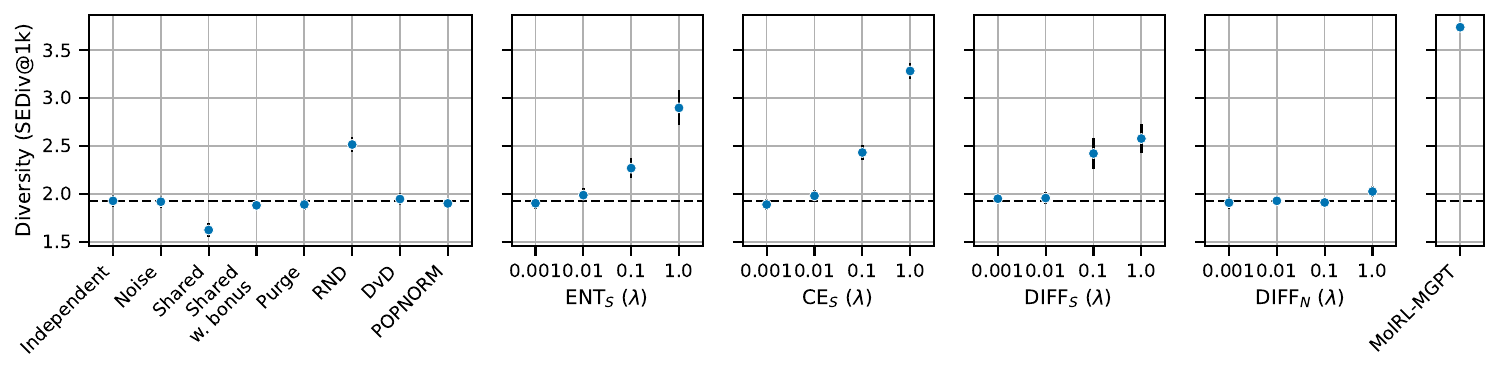}
         \caption{Cooperative RL molecular diversity}
         \label{fig:MolExp_coop_diversity}
     \end{subfigure}
    \caption{Performance comparison of different 4-agent cooperative strategies on the MolExp benchmark, each with a budget of 10k. The dashed line represents the average of 4 independent agents as baseline.}
\end{figure*}

\begin{figure}[ht]
%\vskip 0.2in
\begin{center}
\centerline{\includegraphics[width=\columnwidth]{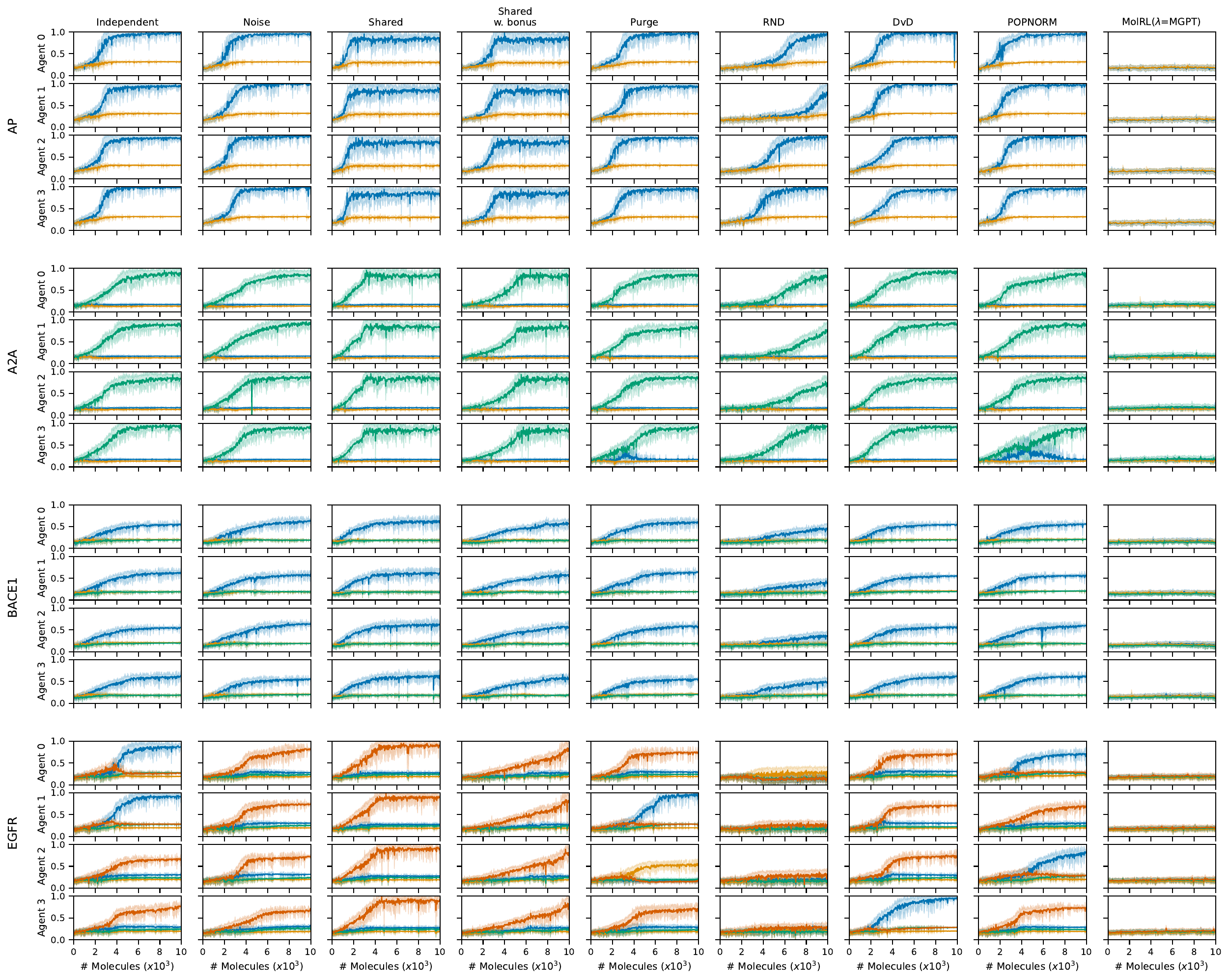}}
\caption{Cooperative strategies using 4-agents on the MolExp benchmark tasks, single replicate, part I. Each line represents average similarity to one of the tasks target molecules. Training of each agent is plotted. Perfect cooperation is each agent learning to rediscover a different target molecule. Most cooperative strategies result in no additional divergent behavior or in some cases, slower learning.}
\label{fig:MolExp_coop_sim_I}
\end{center}
%\vskip -0.2in
\end{figure}

\begin{figure}[ht]
%\vskip 0.2in
\begin{center}
\centerline{\includegraphics[width=\columnwidth]{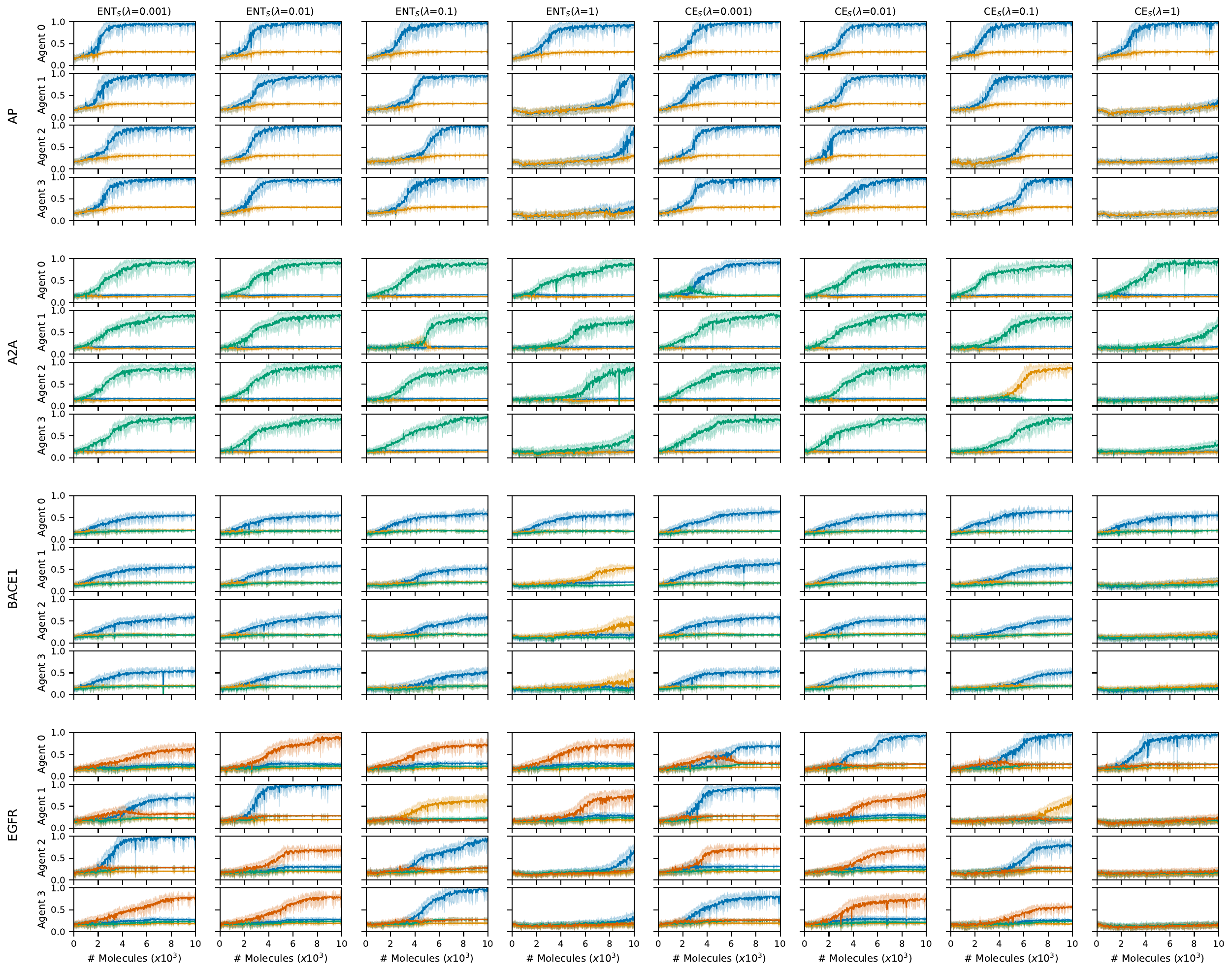}}
\caption{Cooperative strategies using 4-agents on the MolExp benchmark tasks, single replicate, part II. Each line represents average similarity to one of the tasks target molecules. Training of each agent is plotted. Perfect cooperation is each agent learning to rediscover a different target molecule. Most cooperative strategies result in no additional divergent behavior or in some cases, slower learning.}
\label{fig:MolExp_coop_sim_II}
\end{center}
%\vskip -0.2in
\end{figure}

\begin{figure}[ht]
%\vskip 0.2in
\begin{center}
\centerline{\includegraphics[width=\columnwidth]{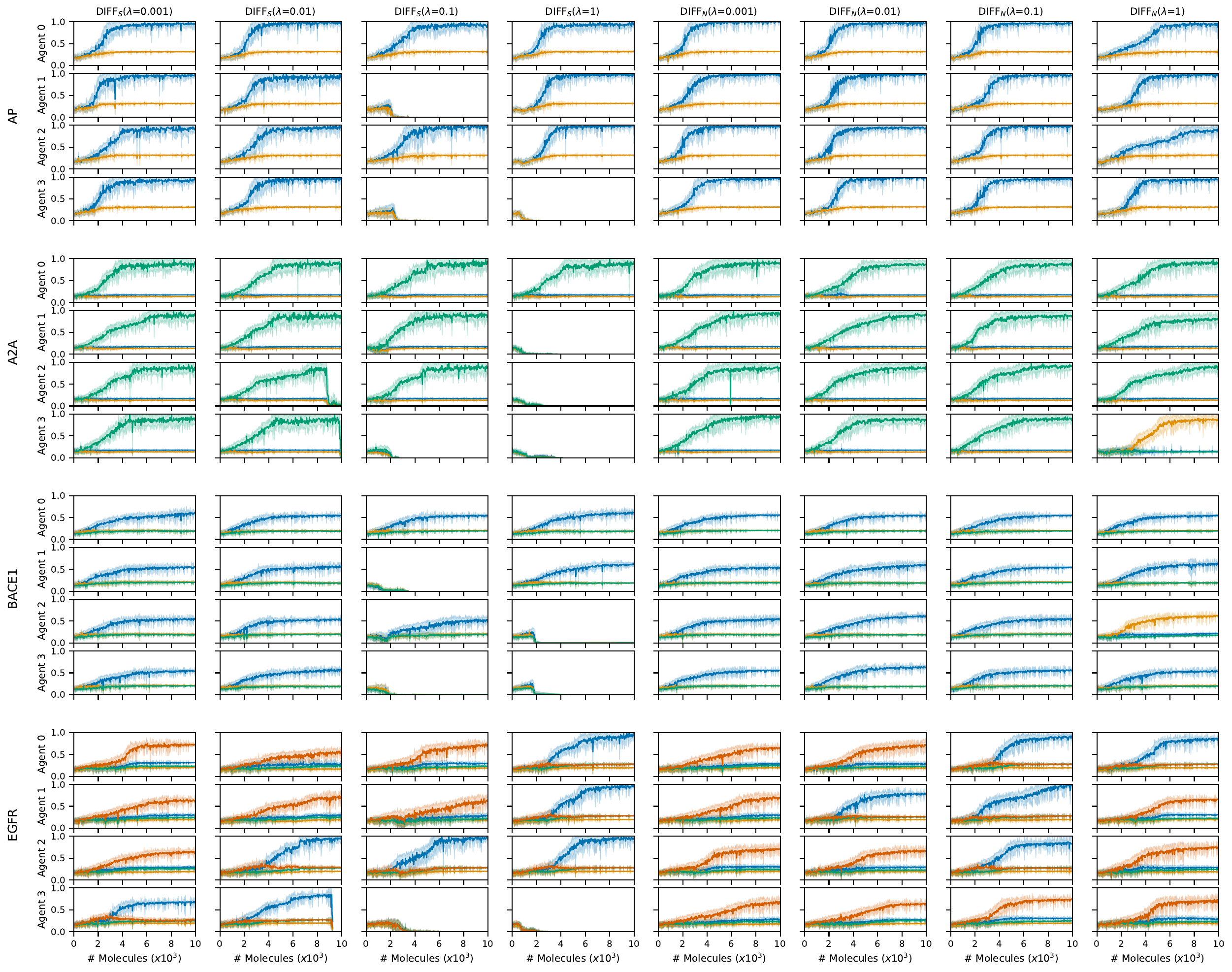}}
\caption{Cooperative strategies using 4-agents on the MolExp benchmark tasks, single replicate, part III. Each line represents average similarity to one of the tasks target molecules. Training of each agent is plotted. Perfect cooperation is each agent learning to rediscover a different target molecule. Most cooperative strategies result in no additional divergent behavior or in some cases, slower learning.}
\label{fig:MolExp_coop_sim_III}
\end{center}
%\vskip -0.2in
\end{figure}

%%%%%%%%%%%%%%%%%%%%%%%%%%%%%%%%%%%%%%%%%%%%%%%%%%%%%%%%%%%%%%%%%%%%%%%%%%%%%%%
\clearpage
\section{A2A bioactivity maximization}\label{app:MolExpBio_scaling}

\begin{table}[ht!]
\caption{Performance on MolExpBio task with increasing number of independent ACEGEN$_{MolOpt}$ agents, each with a budget of 10,000.}
\label{tab:MolExpBio_scaling_agents}
\resizebox{\textwidth}{!}{%
\begin{tabular}{l|cccccccc}
\hline
 & 1 & 2 & 4 & 8 & 16 & 32 & 64 & 128 \\
\hline
Target 1 max sim & 0.31 ± 0.02 & 0.32 ± 0.03 & 0.37 ± 0.03 & 0.36 ± 0.01 & 0.43 ± 0.07 & 0.44 ± 0.04 & 0.49 ± 0.07 & \textbf{0.54 ± 0.07} \\
Target 2 max sim & 0.31 ± 0.03 & 0.34 ± 0.03 & 0.36 ± 0.01 & 0.38 ± 0.03 & 0.39 ± 0.02 & 0.41 ± 0.02 & 0.43 ± 0.05 & \textbf{0.44 ± 0.04} \\
Target 3 max sim & 0.50 ± 0.08 & 0.53 ± 0.06 & 0.55 ± 0.02 & 0.60 ± 0.03 & 0.60 ± 0.03 & 0.62 ± 0.03 & \textbf{0.63 ± 0.03} & 0.63 ± 0.01 \\
\hline
MolExpBio score & 0.05 ± 0.00 & 0.06 ± 0.01 & 0.07 ± 0.01 & 0.08 ± 0.01 & 0.10 ± 0.02 & 0.11 ± 0.01 & 0.13 ± 0.02 & \textbf{0.15 ± 0.02} \\
\hline
\end{tabular}%
}
\end{table}

\begin{table}[ht!]
\caption{Performance on MolExpBio task with increasing budget for a single ACEGEN$_{MolOpt}$ agent.}
\label{tab:MolExpBio_scaling_single}
\resizebox{\textwidth}{!}{%
\begin{tabular}{l|cccccccc}
\hline
 & 10k & 20k & 40k & 80k & 160k & 320k & 640k & 1280k \\
\hline
Target 1 max sim & \textbf{0.31 ± 0.02} & 0.31 ± 0.02 & 0.31 ± 0.02 & 0.31 ± 0.02 & 0.31 ± 0.02 & 0.31 ± 0.02 & 0.31 ± 0.02 & 0.31 ± 0.02 \\
Target 2 max sim & \textbf{0.31 ± 0.03} & 0.31 ± 0.03 & 0.31 ± 0.03 & 0.31 ± 0.03 & 0.31 ± 0.03 & 0.31 ± 0.03 & 0.31 ± 0.03 & 0.31 ± 0.03 \\
Target 3 max sim & \textbf{0.50 ± 0.08} & 0.50 ± 0.08 & 0.50 ± 0.08 & 0.50 ± 0.08 & 0.50 ± 0.08 & 0.50 ± 0.08 & 0.50 ± 0.08 & 0.50 ± 0.08 \\
\hline
MolExpBio score & \textbf{0.05 ± 0.00} & 0.05 ± 0.00 & 0.05 ± 0.00 & 0.05 ± 0.00 & 0.05 ± 0.00 & 0.05 ± 0.00 & 0.05 ± 0.00 & 0.05 ± 0.00 \\
\hline
\end{tabular}%
}
\end{table}

\begin{table}[ht!]
\caption{Performance on MolExpBio task with increasing budget for a single ACEGEN$_{MolOpt}$ agent with RND.}
\label{tab:MolExpBio_scaling_single_RND}
\resizebox{\textwidth}{!}{%
\begin{tabular}{l|cccccccc}
\hline
 & 10k & 20k & 40k & 80k & 160k & 320k & 640k & 1280k \\
\hline
Target 1 max sim & \textbf{0.37 ± 0.07} & 0.37 ± 0.07 & 0.37 ± 0.07 & 0.37 ± 0.07 & 0.37 ± 0.07 & 0.34 ± 0.02 & 0.37 ± 0.07 & 0.37 ± 0.07 \\
Target 2 max sim & 0.33 ± 0.02 & 0.33 ± 0.02 & 0.33 ± 0.02 & 0.33 ± 0.02 & 0.33 ± 0.02 & \textbf{0.34 ± 0.03} & 0.33 ± 0.02 & 0.33 ± 0.02 \\
Target 3 max sim & 0.55 ± 0.06 & \textbf{0.56 ± 0.06} & 0.56 ± 0.06 & 0.56 ± 0.06 & 0.56 ± 0.06 & 0.54 ± 0.04 & 0.56 ± 0.06 & 0.56 ± 0.06 \\
\hline
MolExpBio score & \textbf{0.07 ± 0.02} & 0.07 ± 0.02 & 0.07 ± 0.02 & 0.07 ± 0.02 & 0.07 ± 0.02 & 0.06 ± 0.01 & 0.07 ± 0.02 & 0.07 ± 0.02 \\
\hline
\end{tabular}%
}
\end{table}

\begin{table}[ht!]
\caption{Performance on MolExpBio task with increasing budget for a single ACEGEN$_{MolOpt}$ agent with DF penalization.}
\label{tab:MolExpBio_scaling_single_DF}
\resizebox{\textwidth}{!}{%
\begin{tabular}{l|cccccccc}
\hline
 & 10k & 20k & 40k & 80k & 160k & 320k & 640k & 1280k \\
\hline
Target 1 max sim & \textbf{0.31 ± 0.02} & 0.31 ± 0.02 & 0.31 ± 0.02 & 0.31 ± 0.02 & 0.31 ± 0.02 & 0.31 ± 0.02 & 0.31 ± 0.02 & 0.31 ± 0.02 \\
Target 2 max sim & 0.31 ± 0.03 & 0.31 ± 0.03 & 0.31 ± 0.03 & 0.31 ± 0.03 & 0.31 ± 0.03 & \textbf{0.32 ± 0.02} & 0.32 ± 0.02 & 0.32 ± 0.02 \\
Target 3 max sim & \textbf{0.50 ± 0.05} & 0.50 ± 0.05 & 0.50 ± 0.05 & 0.50 ± 0.05 & 0.50 ± 0.05 & 0.50 ± 0.05 & 0.50 ± 0.05 & 0.50 ± 0.05 \\
\hline
MolExpBio score & \textbf{0.05 ± 0.00} & 0.05 ± 0.00 & 0.05 ± 0.00 & 0.05 ± 0.00 & 0.05 ± 0.00 & 0.05 ± 0.00 & 0.05 ± 0.00 & 0.05 ± 0.00 \\
\hline
\end{tabular}%
}
\end{table}

\begin{figure}[ht!]
%\vskip 0.2in
\begin{center}
\centerline{\includegraphics[width=\columnwidth]{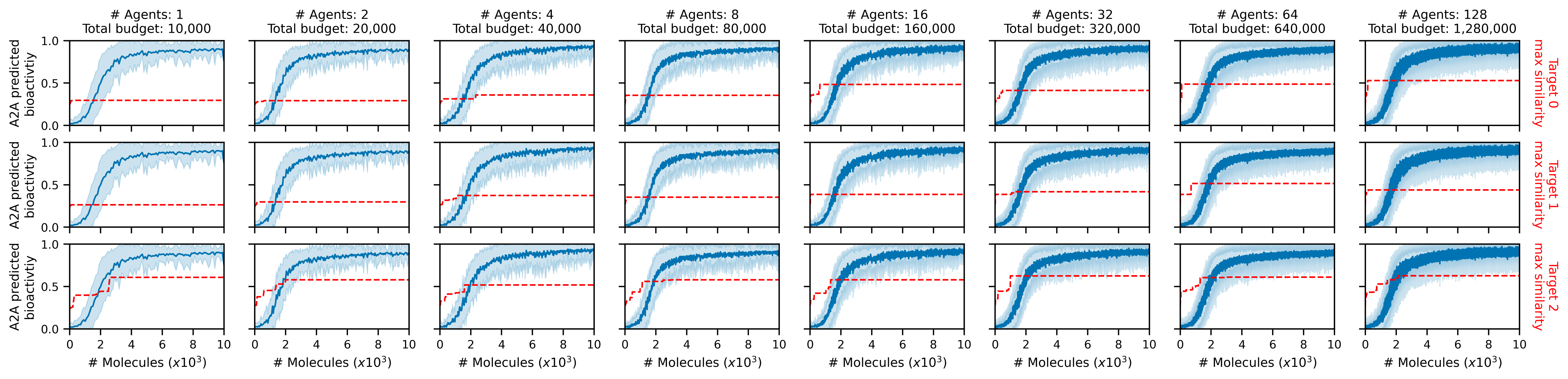}}
\caption{Multiple independent ACEGEN$_{MolOpt}$ agents on the MolExpBio task of maximizing A2A predicted bioactivity, single replicate. Maximization of predicted A2A bioactivity is shown, as well as the maximum similarity to each A2A target molecule in the set from the MolExp A2A task in red.}
\label{fig:MolExpBio_agent_scaling_sim}
\end{center}
%\vskip -0.2in
\end{figure}

\begin{figure}[ht!]
%\vskip 0.2in
\begin{center}
\centerline{\includegraphics[width=\columnwidth]{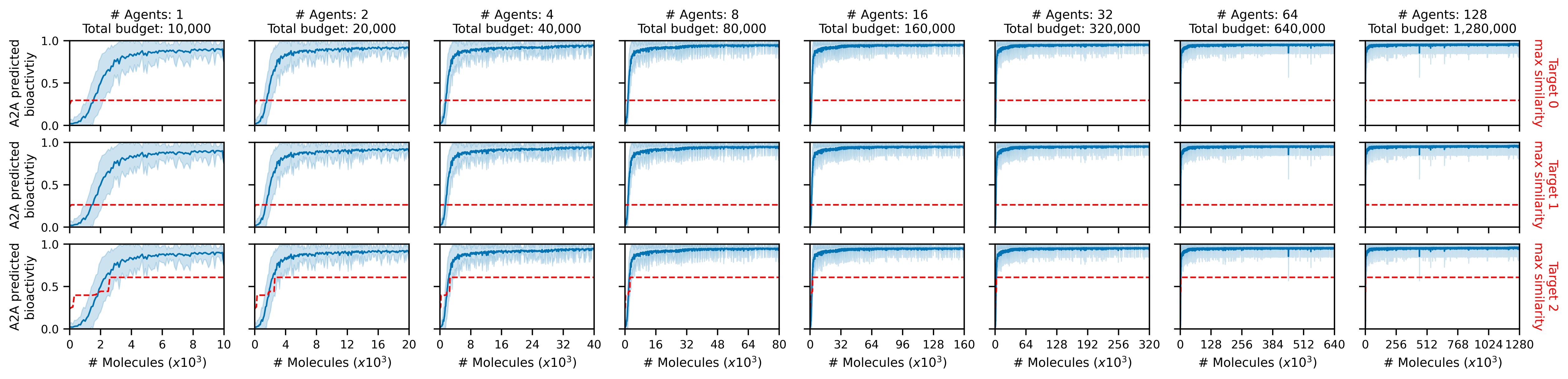}}
\caption{Single ACEGEN$_{MolOpt}$ agent on the MolExpBio task of maximizing A2A predicted bioactivity, single replicate. Maximization of predicted A2A bioactivity is shown, as well as the maximum similarity to each A2A target molecule in the set from the MolExp A2A task in red.}
\label{fig:MolExpBio_single_scaling_sim}
\end{center}
%\vskip -0.2in
\end{figure}

\begin{figure}[ht!]
%\vskip 0.2in
\begin{center}
\centerline{\includegraphics[width=\columnwidth]{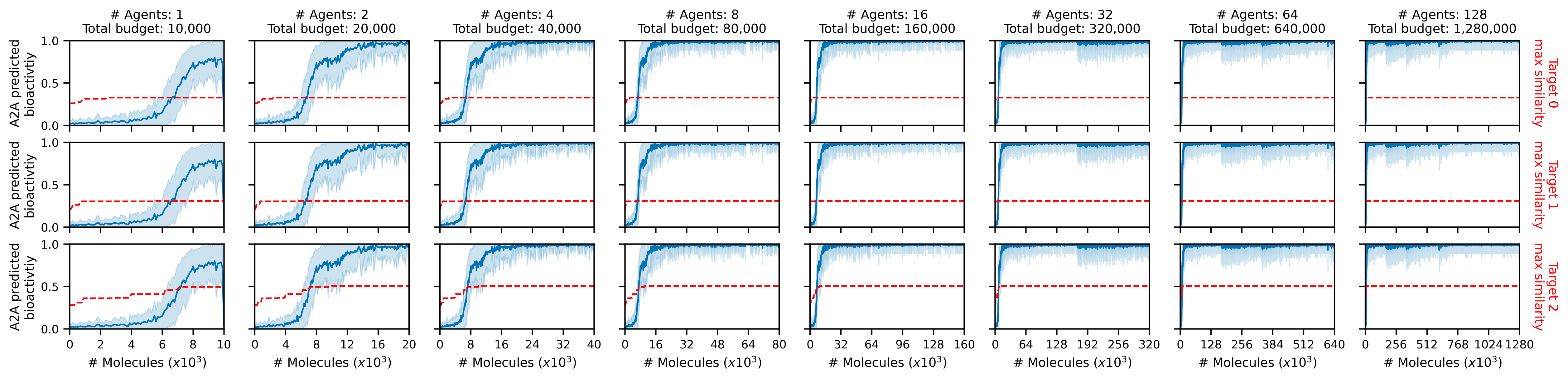}}
\caption{Single ACEGEN$_{MolOpt}$ agent with RND exploration bonus on the MolExpBio task of maximizing A2A predicted bioactivity, single replicate. Maximization of predicted A2A bioactivity is shown, as well as the maximum similarity to each A2A target molecule in the set from the MolExp A2A task in red.}
\label{fig:MolExpBio_singleRND_scaling_sim}
\end{center}
%\vskip -0.2in
\end{figure}

\begin{figure}[ht!]
%\vskip 0.2in
\begin{center}
\centerline{\includegraphics[width=\columnwidth]{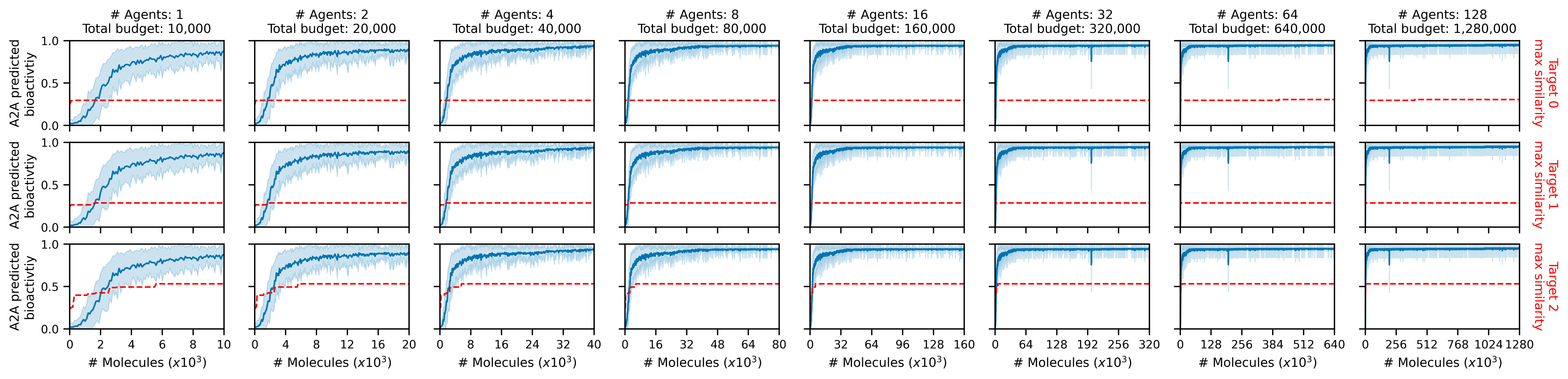}}
\caption{Single ACEGEN$_{MolOpt}$ agent with DF penalization on the MolExpBio task of maximizing A2A predicted bioactivity, single replicate. Maximization of predicted A2A bioactivity is shown, as well as the maximum similarity to each A2A target molecule in the set from the MolExp A2A task in red.}
\label{fig:MolExpBio_singleDF_scaling_sim}
\end{center}
%\vskip -0.2in
\end{figure}

%%%%%%%%%%%%%%%%%%%%%%%%%%%%%%%%%%%%%%%%%%%%%%%%%%%%%%%%%%%%%%%%%%%%%%%%%%%%%%%
\clearpage
\section{Computational resource}\label{app:comp_requirements}
All experiments detailed in this work were conducted on a single consumer grade GPU, more specifically an NVIDIA RTX 3090 with 1 CPU core.

\begin{figure}[ht]
     \centering
     \begin{subfigure}[l]{0.32\textwidth}
         \centering
         \includegraphics[width=\textwidth]{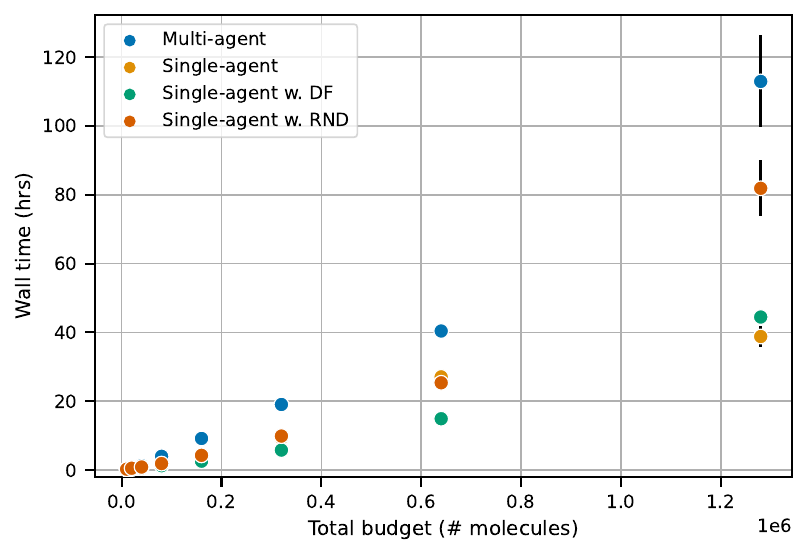}
         \caption{MolExpL benchmark}
         \label{fig:MolExpL_walltime}
     \end{subfigure}
      \begin{subfigure}[l]{0.32\textwidth}
         \centering
         \includegraphics[width=\textwidth]{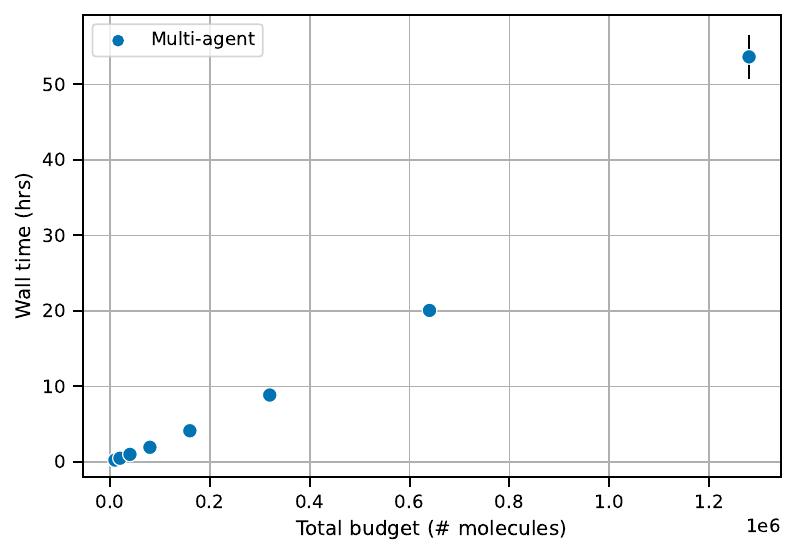}
         \caption{MolExp benchmark}
         \label{fig:MolExp_walltime}
     \end{subfigure}
      \begin{subfigure}[l]{0.32\textwidth}
         \centering
         \includegraphics[width=\textwidth]{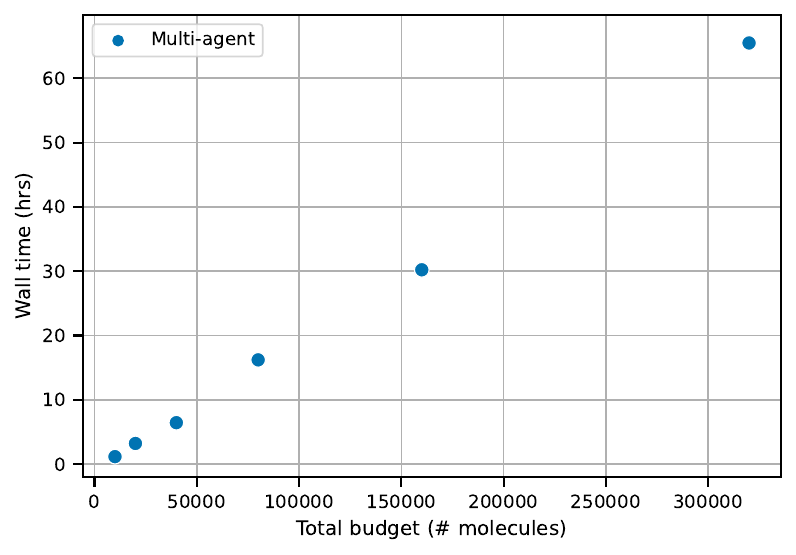}
         \caption{GuacaMol benchmark}
         \label{fig:GuacaMol_walltime}
     \end{subfigure}
     \caption{Wall time required to run benchmarks detailed in this work. Not that the DF was implemented following some code optimizations.}
\end{figure}

%%%%%%%%%%%%%%%%%%%%%%%%%%%%%%%%%%%%%%%%%%%%%%%%%%
% References
%%%%%%%%%%%%%%%%%%%%%%%%%%%%%%%%%%%%%%%%%%%%%%%%%%
\clearpage
\bibliography{refs}